%% file: robust_mfrl.tex
\documentclass{article}

\usepackage[final,nonatbib]{neurips_2020}

\usepackage[utf8]{inputenc} 
\usepackage[T1]{fontenc}    
\usepackage{hyperref}       
\usepackage{url}            
\usepackage{booktabs}       
\usepackage{amsfonts}       
\usepackage{nicefrac}       
\usepackage{microtype}      

\RequirePackage{algorithm}
\RequirePackage{algorithmic}
\usepackage{mathtools}

\usepackage{amsmath}
\usepackage{amsthm}
\usepackage{stmaryrd}
\usepackage{multirow}
\usepackage{amsfonts} 
\usepackage{xcolor}
\usepackage{subcaption}
\usepackage{comment}

\newcommand{\bc}[1]{\left\{{#1}\right\}}
\newcommand{\br}[1]{\left({#1}\right)}
\newcommand{\bs}[1]{\left[{#1}\right]}
\newcommand{\abs}[1]{\left| {#1} \right|}

\newcommand{\Ex}[1]{\mathbb{E}\bs{{#1}}}
\newcommand{\Ee}[2]{\underset{#1}{\mathbb{E}}\bs{{#2}}}

\newtheorem{theorem}{Theorem}

\newtheorem{remark}{Remark}

\definecolor{dark-green}{rgb}{0.0, 0.5, 0.0}

\DeclarePairedDelimiter\floor{\lfloor}{\rfloor}

\usepackage{xcolor}

\allowdisplaybreaks

\title{Robust Reinforcement Learning via\\ Adversarial training with Langevin Dynamics}

\author{%
  Parameswaran Kamalaruban\thanks{Work done while Parameswaran Kamalaruban and Cheng Shi were working at LIONS, EPFL.} \\
  The Alan Turing Institute \\ 
  \texttt{kparameswaran@turing.ac.uk} \\
  \And
  Yu-Ting Huang \\
  EPFL \\
  \texttt{yu.huang@epfl.ch} \\ 
  \And
 Ya-Ping Hsieh \\
 LIONS, EPFL \\
  \texttt{ya-ping.hsieh@epfl.ch} \\ 
  \And
  Paul Rolland \\
  LIONS, EPFL \\
  \texttt{paul.rolland@epfl.ch} \\ 
  \And
  Cheng Shi \\
  University of Basel \\
  \texttt{cheng.shi@unibas.ch} \\ 
  \And
  Volkan Cevher \\
  LIONS, EPFL \\
  \texttt{volkan.cevher@epfl.ch} \\ 
  \And
}

\begin{document}

\maketitle

\input{0_abstract}

\input{1_introduction}
\input{2_background}

\input{3_two_player}

\input{3.5_case_study}

\input{4_experiments}

\input{5_conclusions}

\newpage
\section*{Broader Impact}
Our work would be beneficial in control/robotics applications where the training happens in a simulation environment which is only a rough estimation of the real domain where the agent will be deployed after training. Thus our adversarial training would account for this mismatch and would lead to a stable performance. Even though our adversarial training method improves robustness, being overly conservative might result in lower performance. Thus one should carefully tune the robustness related hyperparameters, in our case $\delta$. We could not imagine any immediate negative ethical/societal impact of our work. 

\begin{ack}
This work has received funding from the European Research Council (ERC) under the European Union's Horizon 2020 research and innovation program (grant agreement n 725594 - time-data), the Swiss National Science Foundation (SNSF) under grant number 407540\_167319, and the Army Research Office under grant number W911NF-19-1-0404. 
\end{ack}

{\small
\bibliography{robust_mfrl}
\bibliographystyle{unsrt}
}

\newpage
\appendix
\input{7_appendix}

\end{document}

%% file: 0_abstract.tex
\begin{abstract}
%
We introduce a \emph{sampling} perspective to tackle the challenging task of training robust Reinforcement Learning (RL) agents. Leveraging the powerful Stochastic Gradient Langevin Dynamics, we present a novel, scalable two-player RL algorithm, which is a sampling variant of the two-player policy gradient method. Our algorithm consistently outperforms existing baselines, in terms of generalization across different training and testing conditions, on several MuJoCo environments. Our experiments also show that, even for objective functions that entirely ignore potential environmental shifts, our sampling approach remains highly robust in comparison to standard RL algorithms.
\end{abstract}

%% file: 1_introduction.tex
\section{Introduction}
\label{sec:introduction}

Reinforcement learning (RL) promise automated solutions to many real-world tasks with beyond-human performance. Indeed,
recent advances in policy gradient methods \cite{sutton2000policy,silver2014deterministic,schulman2015trust,schulman2017proximal} and deep reinforcement learning have demonstrated impressive performance in games \cite{mnih2015human,silver2017mastering}, continuous control \cite{lillicrap2015continuous}, and robotics \cite{levine2016end}.

Despite the success of deep RL, the progress is still upset by the fragility in real-life deployments. In particular, the majority of these methods fail to perform well when there is some difference between training and testing scenarios, thereby posting serious safety and security concerns. To this end, learning policies that are \emph{robust} to environmental shifts, mismatched configurations, and even mismatched control actions are becoming increasingly more important. 

A powerful framework to learning robust policies is to interpret the changing of the environment as an adversarial perturbation. This notion naturally lends itself to a two-player max-min problem involving a pair of agents, a protagonist and an adversary, where the protagonist learns to fulfill the original task goals while being robust to the disruptions generated by its adversary. Two prominent examples along this research vein, differing in how they model the adversary, are the Robust Adversarial Reinforcement Learning (RARL) ~\cite{pinto2017robust} and Noisy Robust Markov Decision Process (NR-MDP) ~\cite{tessler2019action}. 

Despite the impressive empirical progress, {the training of the robust RL objectives} remains an open and critical challenge. In particular, \cite{tessler2019action} prove that it is in fact strictly suboptimal to directly apply (deterministic) policy gradient steps to their NR-MDP max-min objectives. Owing to the lack of a better algorithm, the policy gradient is nonetheless still employed in their experiments; similar comments also apply to \cite{pinto2017robust}.

The main difficulty originates from the highly non-convex-concave nature of the robust RL objectives, posing significant burdens to all optimization methods. In game-theoretical terms, these methods search for pure Nash Equilibria (pure NE) which might not even exist \cite{dasgupta1986existence}. Worse, even when pure NE are well-defined, we show that optimization methods can still get stuck at non-equilibrium stationary points on certain extremely simple non-convex-concave objectives; \textit{cf.}\ Section~\ref{sec:case.study}.

In this paper, we contend that, instead of viewing robust RL as a max-min optimization problem, the \emph{sampling} perspective \cite{hsieh2018finding} from the so-called \emph{mixed Nash Equilibrium} (mixed NE) presents a potential solution to the grand challenge of training robust RL agents. We substantiate our claim by demonstrating the advantages of sampling algorithms over-optimization methods on three fronts:

\begin{enumerate}
\item We show in Section~\ref{sec:case.study} that, even in stylized examples that trap the common optimization methods, the sampling algorithms can still make progress towards the optimum in expectation, even tracking the NE points. 
\vspace{-1mm}
\item We conduct extensive experiments to show that sampling algorithms consistently outperform state-of-the-arts in training robust RL agents. Moreover, our experiments on the MuJoCo dataset reveal that sampling algorithms are able to handle previous failure cases of optimization methods, such as the inverted pendulum.
\vspace{-1mm}
\item Finally, we provide strong empirical evidence that sampling algorithms are inherently more robust than optimization methods for RL. Specifically, we apply sampling algorithms to train an RL agent with \emph{non-robust} objective (\textit{i.e.}, the standard expected cumulative reward maximizing objective in RL), and we compare against the policy learned by \emph{optimizing the robust objective} (\textit{i.e.}, the max-min formulation). Despite the disadvantage, our results show that the sampling algorithms still achieve comparable or better performance than optimization methods (\emph{cf.} Appendix~\ref{sec:one-player-ddpg-app}).
\end{enumerate}

%% file: 2_background.tex
\section{Background}
\label{sec:background}

\paragraph{Stochastic Gradient Langevin Dynamics.}
For any probability distribution $p\br{z} \propto \exp\br{-g\br{z}}$, the Stochastic Gradient Langevin Dynamics (SGLD) \cite{welling2011bayesian} iterates as
\begin{equation}
\label{eq:sgld-sampling}
z_{k+1} ~\gets~ z_{k} - \eta \bs{\widehat{\nabla_z g \br{z}}}_{z = z_k} + \sqrt{2 \eta} \epsilon \xi_k , 
\end{equation}
where $\eta$ is the step-size, $\widehat{\nabla_z g \br{z}}$ is an unbiased estimator of $\nabla_z g \br{z}$, $\epsilon > 0$ is a temperature parameter, and $\xi_k \sim \mathcal{N}\br{0,I}$ is a standard normal vector, independently drawn across different iterations. In some cases, the convergence rate of SGLD can be improved by scaling the noise using a positive-definite symmetric matrix $C$. We thus define a preconditioned variant of the above update \eqref{eq:sgld-sampling} as follows:
\begin{equation}
\label{eq:sgld-sampling-pre}
z_{k+1} ~\gets~ z_{k} - \eta C^{-1} \bs{\widehat{\nabla_z g \br{z}}}_{z = z_k} + \sqrt{2 \eta} \epsilon C^{-\frac{1}{2}} \xi_k .
\end{equation}
In the experiments, we use a RMSProp-preconditioned version of the SGLD \cite{li2016preconditioned}.  

\paragraph{Saddle Point Problems and Pure NE.}
Consider the following Saddle Point Problem (SPP):
\begin{equation}
\label{eq:spp}
\max_{\theta \in \mathbb{R}^n} \min_{\omega \in \mathbb{R}^m} f(\theta , \omega).
\end{equation}
Solving \eqref{eq:spp} equals finding a point $(\theta^\star,\omega^\star)$ such that
\begin{equation}
\label{eq:ne}
f(\theta, \omega^\star) ~\leq~  f(\theta^\star, {\omega^\star} ) ~\leq~  f(\theta^\star, \omega) , \quad \forall \theta \in \mathbb{R}^n , \omega \in \mathbb{R}^m.
\end{equation}
In the language of game theory, we say that $(\theta^\star,\omega^\star)$ is a \emph{pure} Nash Equilibrium (pure NE). If \eqref{eq:ne} holds only locally, we say that $(\theta^\star,\omega^\star)$ is a local pure NE. A major source of SPPs is the Generative Adversarial Networks (GANs) in deep learning \cite{goodfellow2014generative}, which give rise to a variety of algorithms. However, virtually all  search for a (local) pure NE; see Section~\ref{subsec:case.study-algs}.

\paragraph{Sampling for Mixed NE.}
Here, we review some of the key results from \cite{hsieh2018finding}. We denote the set of all probability measures on $\mathcal{Z}$ by $\mathcal{P}\br{\mathcal{Z}}$, and the set of all functions on $\mathcal{Z}$ by $\mathcal{F}\br{\mathcal{Z}}$. Given a (sufficiently regular) function $h: \Theta \times \Omega \rightarrow \mathbb{R}$, consider the following objective (a two-player game with mixed strategies):
\begin{equation}
\label{eq:abstract-game}
\max_{p \in \mathcal{P}\br{\Theta}} \min_{q \in \mathcal{P}\br{\Omega}} ~ f\br{p, q} ~:=~ \Ee{\theta \sim p}{\Ee{\omega \sim q}{h \br{\theta, \omega}}} .
\end{equation}
A pair $\br{p^\star,q^\star}$ achieving the max-min value in \eqref{eq:abstract-game} is called a \emph{mixed Nash Equilibrium} (mixed NE). 

Conceptually, problem (\ref{eq:abstract-game}) can be solved via several infinite-dimensional algorithms, such as the so-called entropic mirror descent or mirror-prox; see \cite{hsieh2018finding}. However, these algorithms are infinite-dimensional and require infinite computational power to implement. For practical interest, by leveraging the SGLD sampling techniques and using some practical relaxations, \cite{hsieh2018finding} features a simplified variant of these infinite-dimensional algorithms. 

For the robust RL formulation \eqref{eq:abstract-game}, it suffices to use the simplest algorithm in \cite{hsieh2018finding}. The pseudocode for their resulting algorithm, termed MixedNE-LD (\textbf{mixed NE} via \textbf{L}angevin \textbf{d}ynamics), can be found in \textbf{Algorithm~\ref{algo:approx-infinite-md}}. 



\begin{algorithm}[tb]
	\caption{MixedNE-LD}
	\label{algo:approx-infinite-md}
	\begin{algorithmic}
		\STATE \textbf{Input:} step-size $\bc{\eta_t}_{t=1}^T$, thermal noise $\bc{\epsilon_t}_{t=1}^T$, warmup steps $\bc{K_t}_{t=1}^T$, damping factor $\beta$.
		\STATE Initialize (randomly) $\omega_1 , \theta_1$
		\FOR{$t=1,2,\dots,T-1$}
		        \STATE $\bar \omega_t , \omega_t^{\br{1}} \gets \omega_t \,\, ; \,\, \bar \theta_t , \theta_t^{\br{1}} \gets \theta_t$
		        \FOR{$k=1,2,\dots,K_t$}
		        		\STATE $\theta_{t}^{\br{k+1}} \gets \theta_{t}^{\br{k}} + \eta_t \widehat{\nabla_\theta h \br{\theta_t^{\br{k}}, \omega_t}} + \sqrt{2 \eta_t} \epsilon_t \xi$, where $\xi \sim \mathcal{N}\br{0,I}$
				\STATE $\omega_{t}^{\br{k+1}} \gets \omega_{t}^{\br{k}} - \eta_t \widehat{\nabla_\omega h \br{\theta_t, \omega_t^{\br{k}}}} + \sqrt{2 \eta_t} \epsilon_t \xi'$, where $\xi' \sim \mathcal{N}\br{0,I}$
				\STATE $\bar \omega_t \gets \br{1 - \beta} \bar \omega_t + \beta \omega_{t}^{\br{k+1}} \,\, ; \,\, \bar \theta_t \gets \br{1 - \beta} \bar \theta_t + \beta \theta_{t}^{\br{k+1}}$
			\ENDFOR 
			\STATE $\omega_{t+1} \gets \br{1 - \beta} \omega_t + \beta \bar \omega_t \,\, ; \,\, \theta_{t+1} \gets \br{1 - \beta} \theta_t + \beta \bar \theta_t$
		\ENDFOR
		\STATE \textbf{Output:} $\omega_T$, $\theta_T$.
	\end{algorithmic}
\end{algorithm}

%% file: 3_two_player.tex
\section{Two-Player Markov Games}
\label{sec:markov.game}
\paragraph{Markov Decision Process.}
We consider a Markov Decision Process (MDP) represented by $\mathcal{M}_1 := \br{\mathcal{S},\mathcal{A},T_1,\gamma,P_0,R_1}$, where the state and action spaces are denoted by $\mathcal{S}$ and $\mathcal{A}$ respectively. We focus on continuous control tasks, where the actions are real-valued, \emph{i.e.}, $\mathcal{A} = \mathbb{R}^d$. $T_1: \mathcal{S} \times \mathcal{S} \times \mathcal{A} \rightarrow \bs{0,1}$ captures the state transition dynamics, \emph{i.e.}, $T_1\br{s' \mid s,a}$ denotes the probability of landing in state $s'$ by taking action $a$ from state $s$. Here $\gamma$ is the discounting factor, $P_0: \mathcal{S} \rightarrow \bs{0,1}$ is the initial distribution over $\mathcal{S}$, and $R_1: \mathcal{S} \times \mathcal{A} \rightarrow \mathbb{R}$ is the reward.

\paragraph{Two-Player Zero-Sum Markov Games.}
Consider a two-player zero-sum Markov game \cite{littman1994markov,perolat2015approximate}, where at each step of the game, both players simultaneously choose an action. The reward each player gets after one step depends on the state and the joint action of both players. Furthermore, the transition kernel of the game is controlled jointly by both players. In this work, we only consider simultaneous games, not the turn-based games.

This game can be described by an MDP $\mathcal{M}_2 = \br{\mathcal{S},\mathcal{A},\mathcal{A'},T_2,\gamma,R_2,P_0}$, where $\mathcal{A}$ and $\mathcal{A'}$ are the continuous set of actions the players can take, $T_2: \mathcal{S} \times \mathcal{A} \times \mathcal{A'} \times \mathcal{S} \rightarrow \mathbb{R}$ is the state transition probability, and $R_2: \mathcal{S} \times \mathcal{A} \times \mathcal{A'} \rightarrow \mathbb{R}$ is the reward for both players. Consider an agent executing a policy $\mu : \mathcal{S} \rightarrow \mathcal{A}$, and an adversary executing a policy $\nu : \mathcal{S} \rightarrow  \mathcal{A'}$ in the environment $\mathcal{M}$. At each time step $t$, both players observe the state $s_t$ and take actions $a_t = \mu\br{s_t}$ and $a'_t = \nu\br{s_t}$. In the zero-sum game, the agent gets a reward $r_t = R_2 \br{s_t , a_t , a'_t}$ while the adversary gets a negative reward $- r_t$. 

This two-player zero-sum Markov game formulation has been used to model the following robust RL settings:
\begin{itemize}
\item Robust Adversarial Reinforcement Learning (RARL)~\cite{pinto2017robust}, where the power of the adversary is limited by its action space $\mathcal{A'}$. 
\item Noisy Robust Markov Decision Process (NR-MDP)~\cite{tessler2019action}, where $\mathcal{A'} = \mathcal{A}$, $T_2 \br{s_{t+1} \mid s_t , a_t , a'_t} = T_1 \br{s_{t+1} \mid s_t , \bar a_t}$, and $R_2 \br{s_t, a_t , a'_t} = R_1 \br{s_t , \bar a_t}$, with $\bar a_t = (1-\delta) a_t + \delta a'_t$, for a chosen $\delta \in \br{0,1}$, which limits the adversary.

\end{itemize}

In our adversarial game, we consider the following performance objective:
\[
J \br{\mu , \nu} ~=~ \Ex{\sum_{t=1}^{\infty}{\gamma^{t-1}r_t} ~\bigg\vert~ \mu , \nu , \mathcal{M}_2} ,
\]
where $\sum_{t=1}^{\infty}{\gamma^{t-1}r_t}$ be the random cumulative return. In particular, we consider the parameterized policies $\bc{\mu_{\theta}: \theta \in \Theta}$, and $\bc{\nu_{\omega}: \omega \in \Omega}$. By an abuse of notation, we denote $J \br{\theta , \omega} = J \br{\mu_\theta , \nu_\omega}$. We consider the following objective:
\begin{equation}
\label{eq:two-player-pure}
\max_{\theta \in \Theta} \min_{\omega \in \Omega} ~ J \br{\theta , \omega} .
\end{equation}
Note that $J$ is non-convex-concave in both $\theta$ and $\omega$. Instead of solving \eqref{eq:two-player-pure} directly, we focus on the mixed strategy formulation of \eqref{eq:two-player-pure}. In other words, we consider the set of all probability distributions over $\Theta$ and $\Omega$, and we search for the optimal distribution that solves the following
program:
\begin{align}
\max_{p \in \mathcal{P}\br{\Theta}} \min_{q \in \mathcal{P}\br{\Omega}} ~ f\br{p, q} ~:=~& \Ee{\theta \sim p}{\Ee{\omega \sim q}{J\br{\theta, \omega}}} . \label{eq:robust-bilinear-problem}
\end{align}
Then, we can use the Algorithm~\ref{algo:approx-infinite-md} to solve the above problem.

%% file: 3.5_case_study.tex

\section{Simple Non-Convex-Concave SPPs}
\label{sec:case.study}

\begin{figure*}[ht!]
  \centering
        \includegraphics[width=0.35\textwidth]{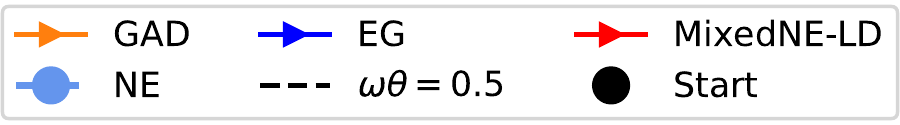}
        \label{fig:legends}\\
    \centering
    \begin{subfigure}[b]{0.237\textwidth}
        \includegraphics[width=\textwidth]{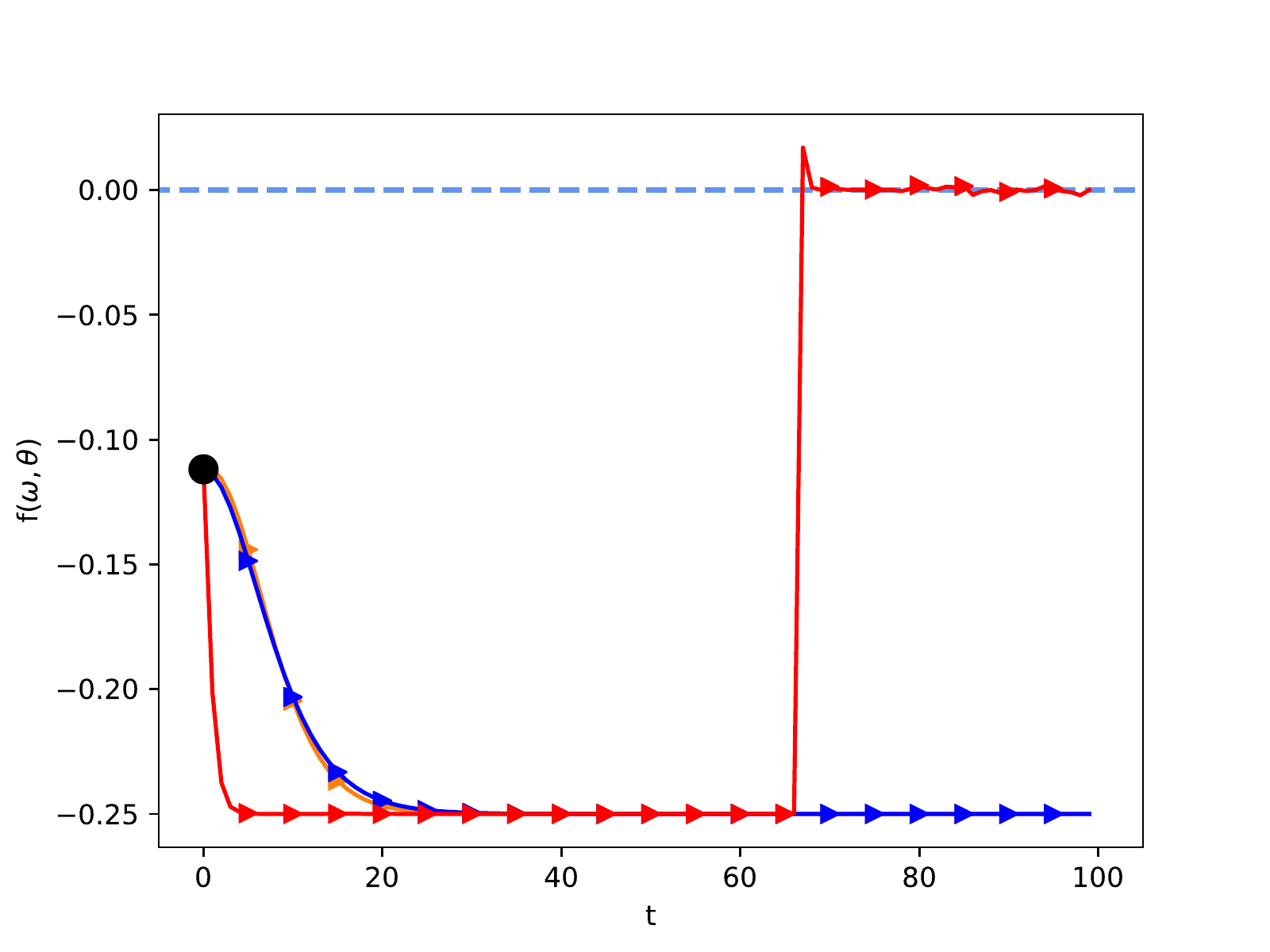}
        \caption{$f(\theta_t,\omega_t)$, far from NE.}
        \label{fig:fxy_1_1-t}
    \end{subfigure}
    ~ 
    \begin{subfigure}[b]{0.237\textwidth}
        \includegraphics[width=\textwidth]{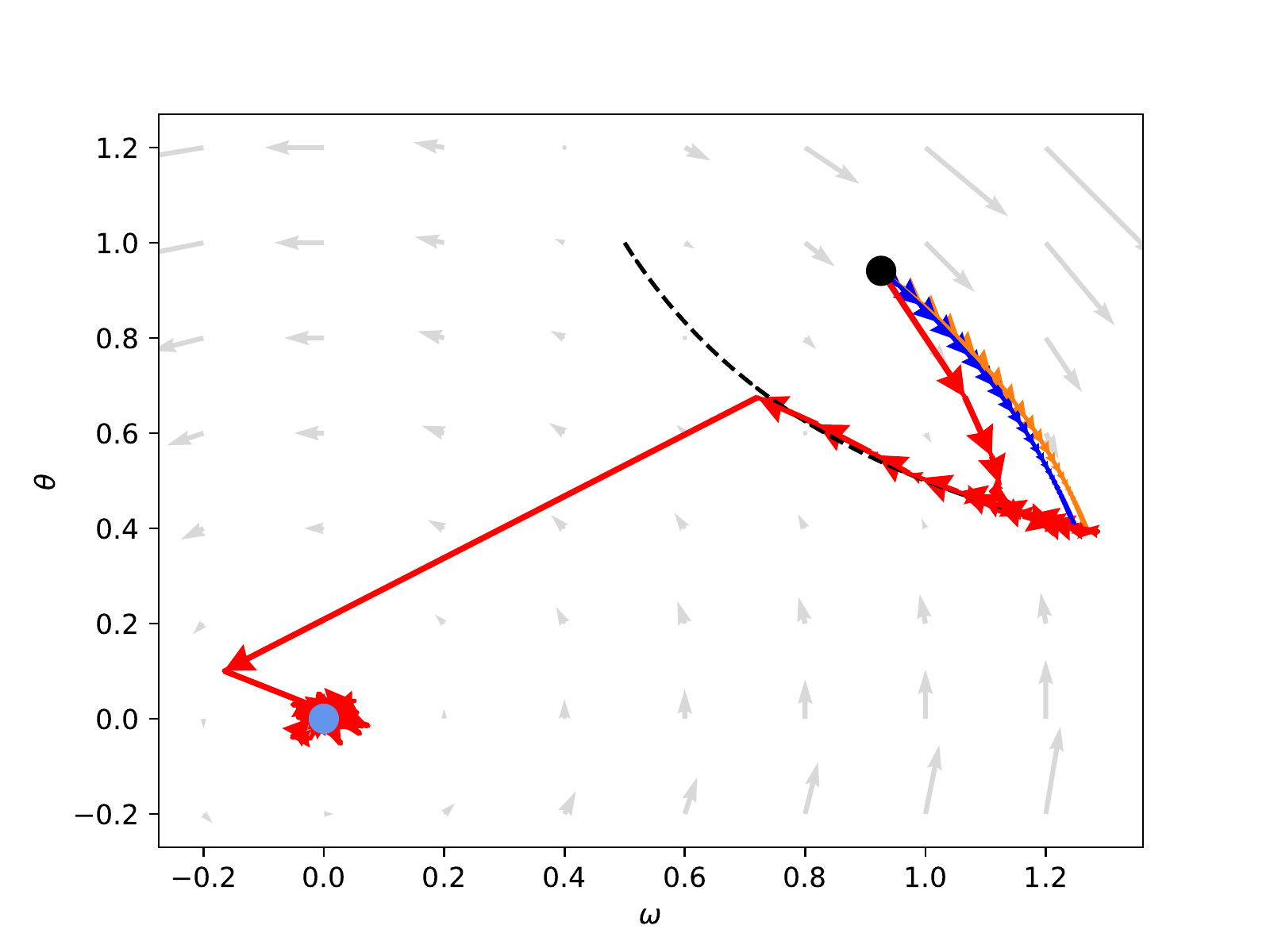}
        \caption{$(\theta_t, \omega_t)$, far from NE.}
        \label{fig:xy_1_1-t}
    \end{subfigure}
    ~
    \begin{subfigure}[b]{0.237\textwidth}
        \includegraphics[width=\textwidth]{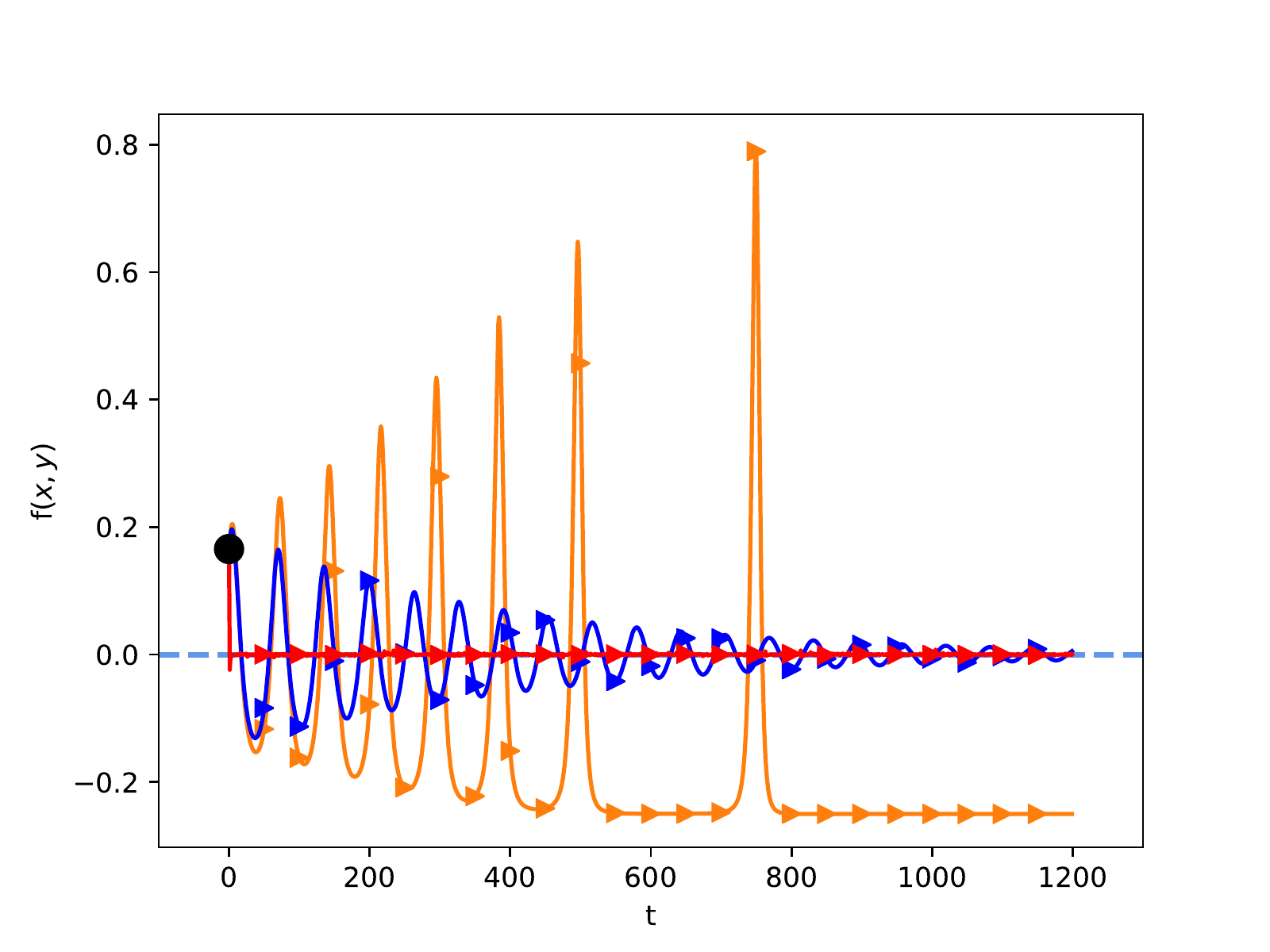}
        \caption{$f(\theta_t,\omega_t)$, close to NE.}
        \label{fig:fxy_1_2-t}
    \end{subfigure}
    ~ 
    \begin{subfigure}[b]{0.237\textwidth}
        \includegraphics[width=\textwidth]{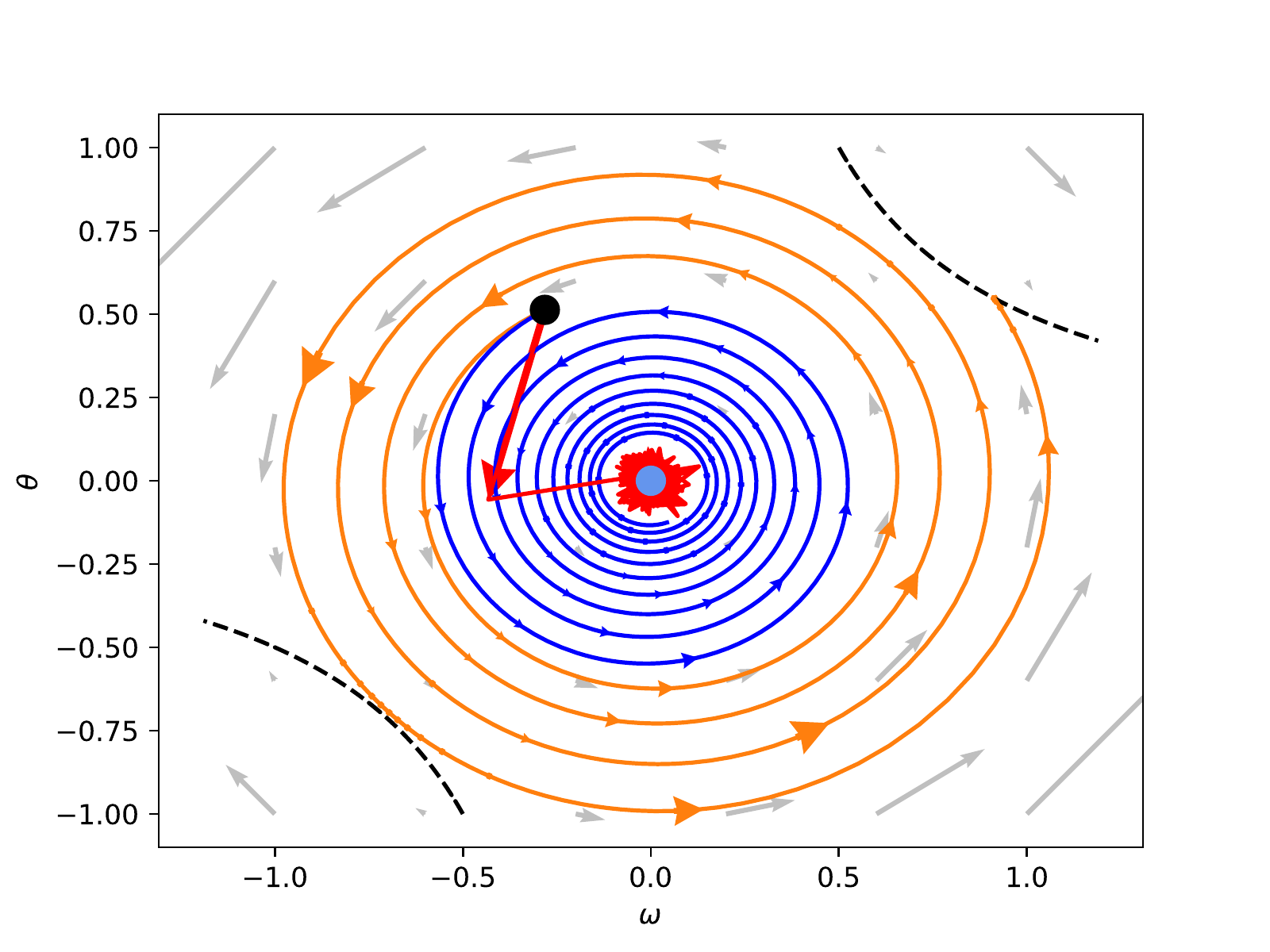}
        \caption{$(\theta_t, \omega_t)$, close to NE.}
        \label{fig:xy_1_2-t}
    \end{subfigure}
    \caption{$f(\theta ,\omega) = \theta^2 \omega^2 - \theta \omega$. The NE is $(0,0)$ with reward value 0. The dashed curve $\theta \omega=0.5$ describe all stationary points that are \emph{not} NE. (a), (b) shows the objective value and the training dynamics when initializing far away from NE. (c), (d) shows the objective value and the training dynamics when $(\theta_1,\omega_1)$ is initializing close to NE.}\label{fig:fxy_1_2}    
    \centering
    \begin{subfigure}[b]{0.237\textwidth}
        \includegraphics[width=\textwidth]{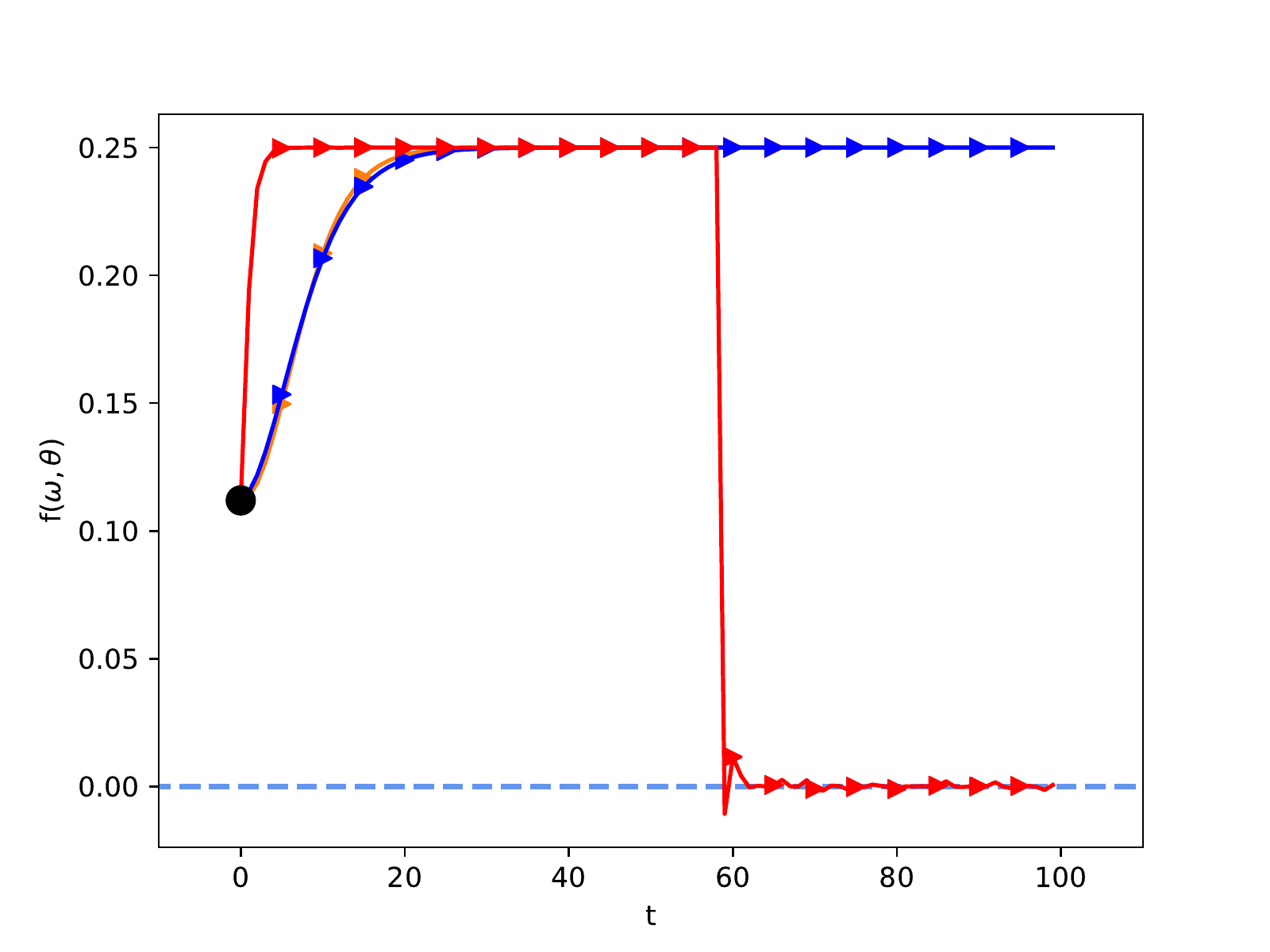}
        \caption{$f(\theta_t,\omega_t)$, far from NE.}
        \label{fig:fxy_3_1-t}
    \end{subfigure}
    ~ 
    \begin{subfigure}[b]{0.237\textwidth}
        \includegraphics[width=\textwidth]{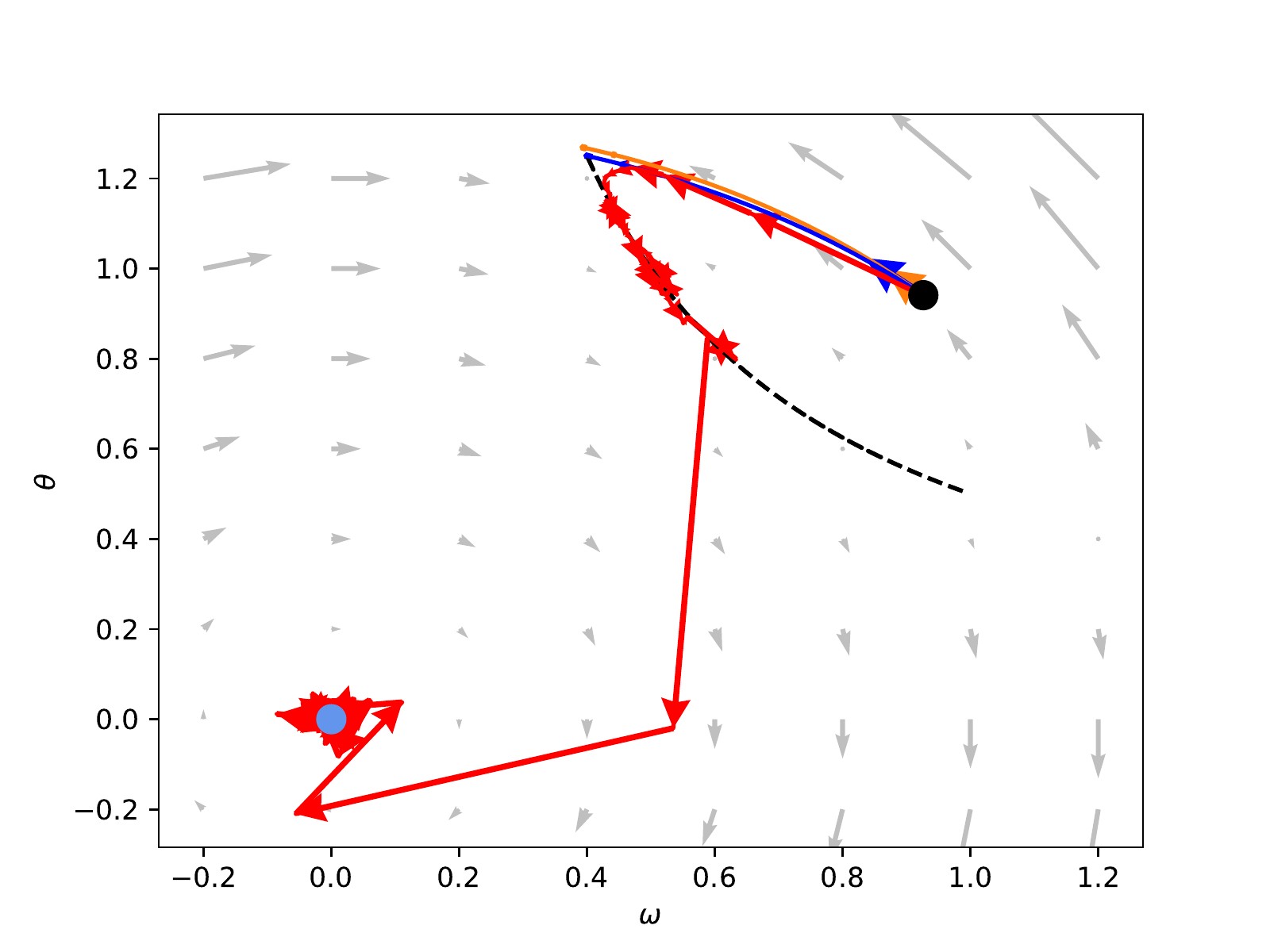}
        \caption{$(\theta_t,\omega_t)$, far from NE.}
        \label{fig:xy_3_1-t}
    \end{subfigure}
    ~
    \begin{subfigure}[b]{0.237\textwidth}
        \includegraphics[width=\textwidth]{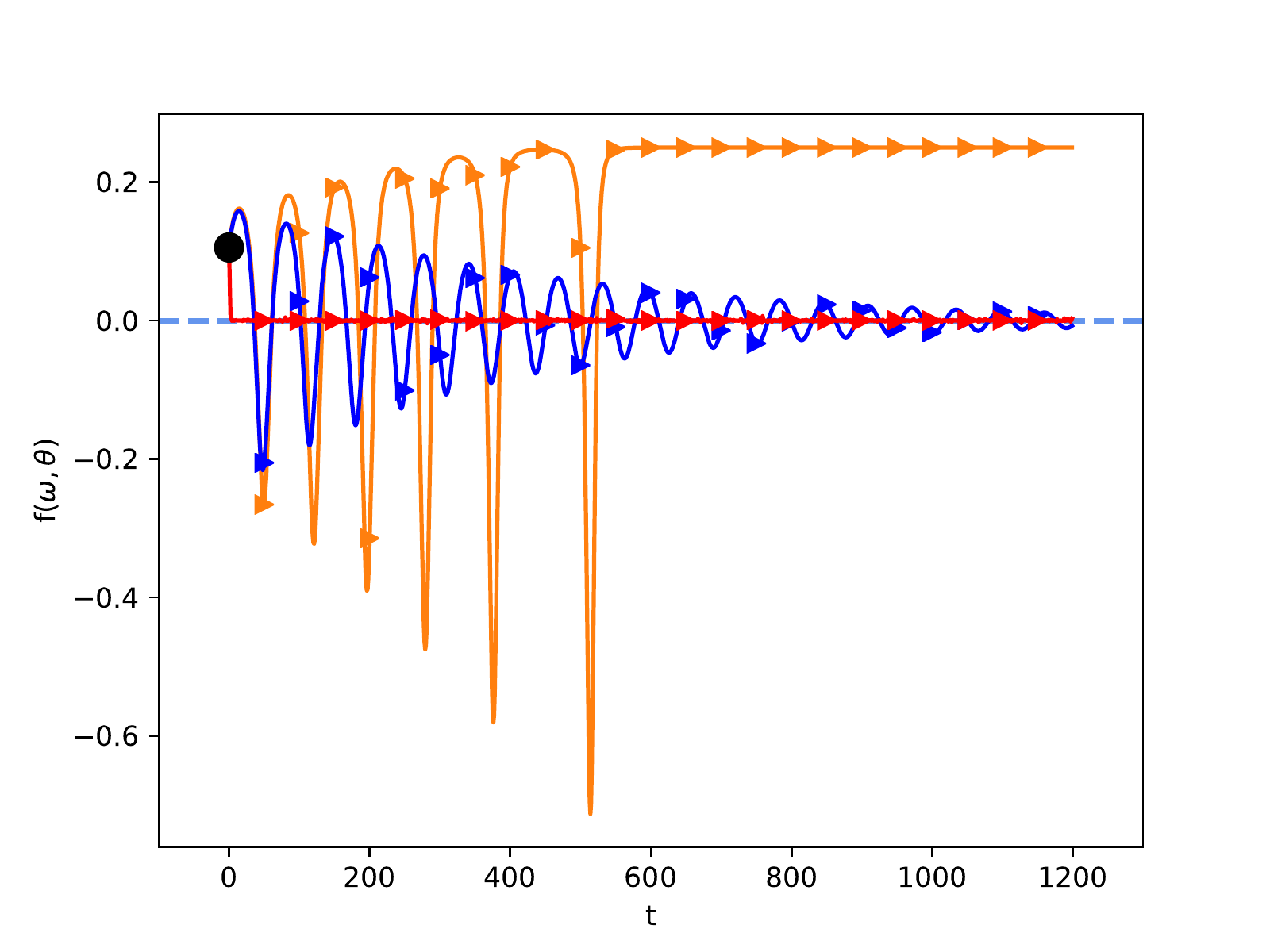}
        \caption{$f(\theta_t,\omega_t)$, close to NE.}
        \label{fig:fxy_3_2-t}
    \end{subfigure}
    ~ 
    \begin{subfigure}[b]{0.237\textwidth}
        \includegraphics[width=\textwidth]{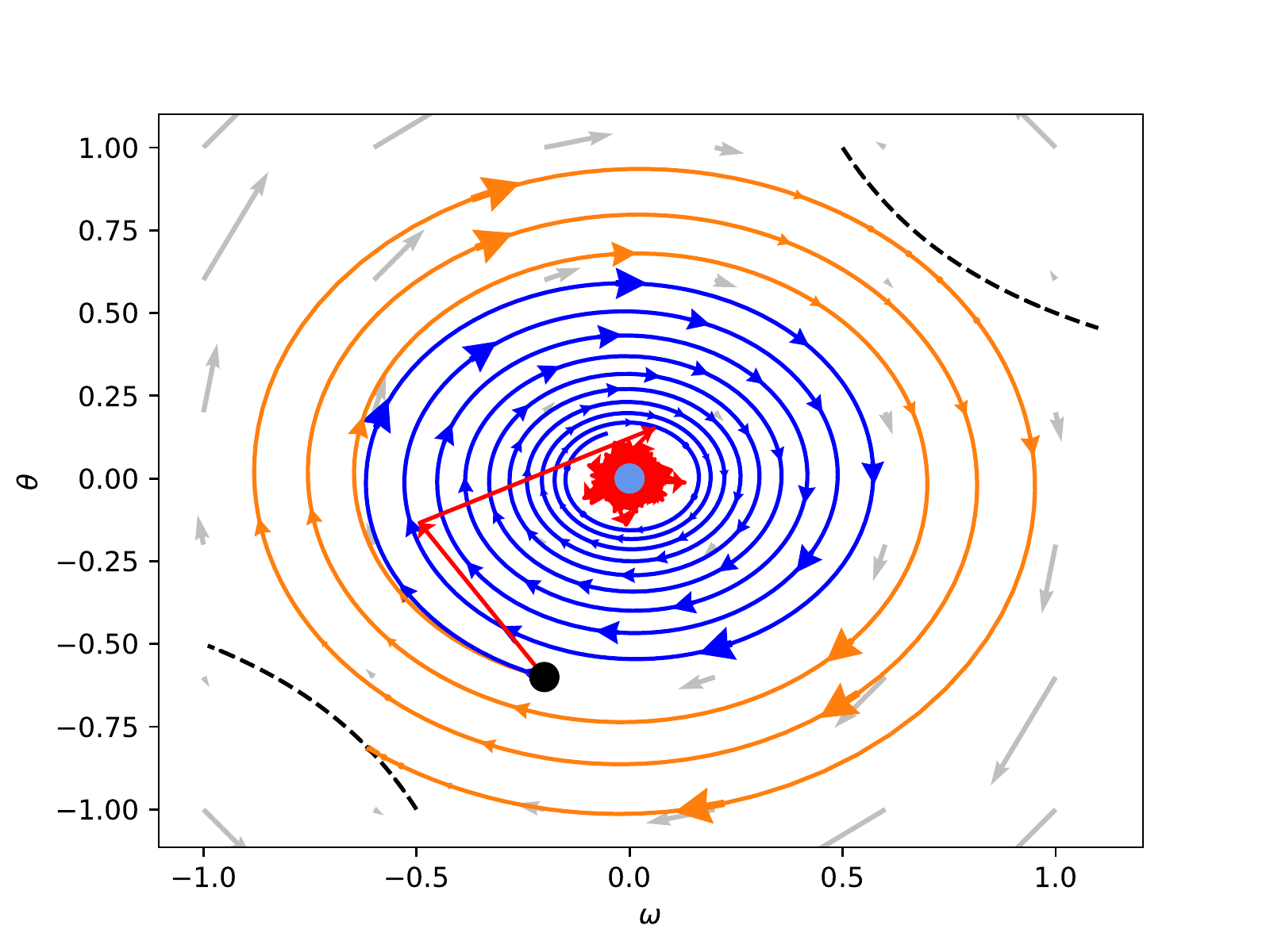}
        \caption{$(\theta_t,\omega_t)$, close to NE.}
        \label{fig:xy_3_2-t}
    \end{subfigure}
     \caption{$f(\theta,\omega) = \theta y - \theta^2 \omega^2$. The NE is (0,0) with reward value 0. The dashed curve $\theta \omega=0.5$ are stationary points that are \emph{not} NE. (a), (b) shows the objective value and the training dynamics when initializing far away from NE. (c), (d) shows the objective value and the training dynamics when initializing close to NE.}\label{fig:fxy_3_2}
\centering
    \begin{subfigure}[b]{0.237\textwidth}
        \includegraphics[width=\textwidth]{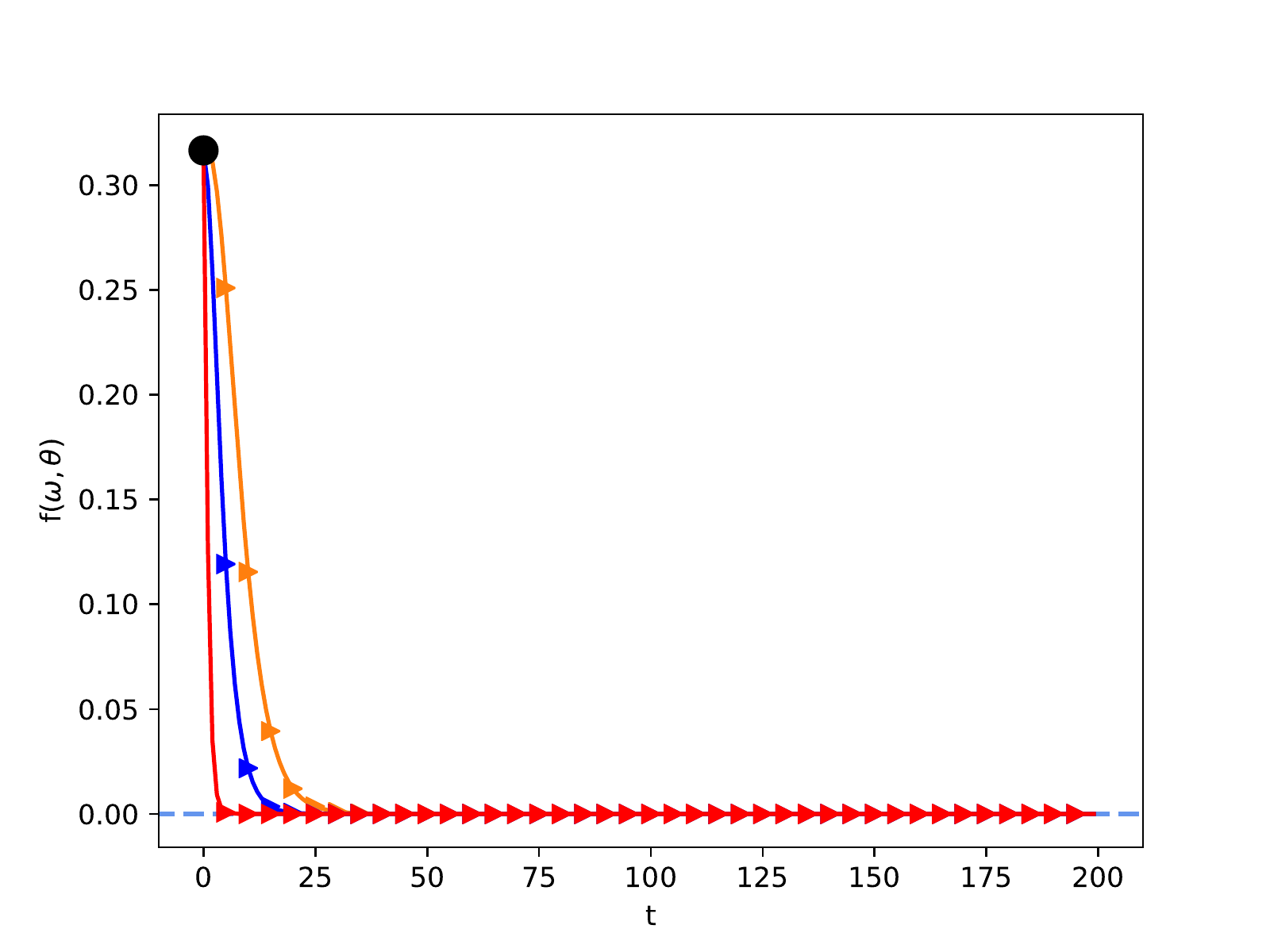}
        \caption{$f(\theta_t,\omega_t)$, far from NE.} 
        \label{fig:fxy_2_1-t}
    \end{subfigure}
    ~ 
    \begin{subfigure}[b]{0.237\textwidth}
        \includegraphics[width=\textwidth]{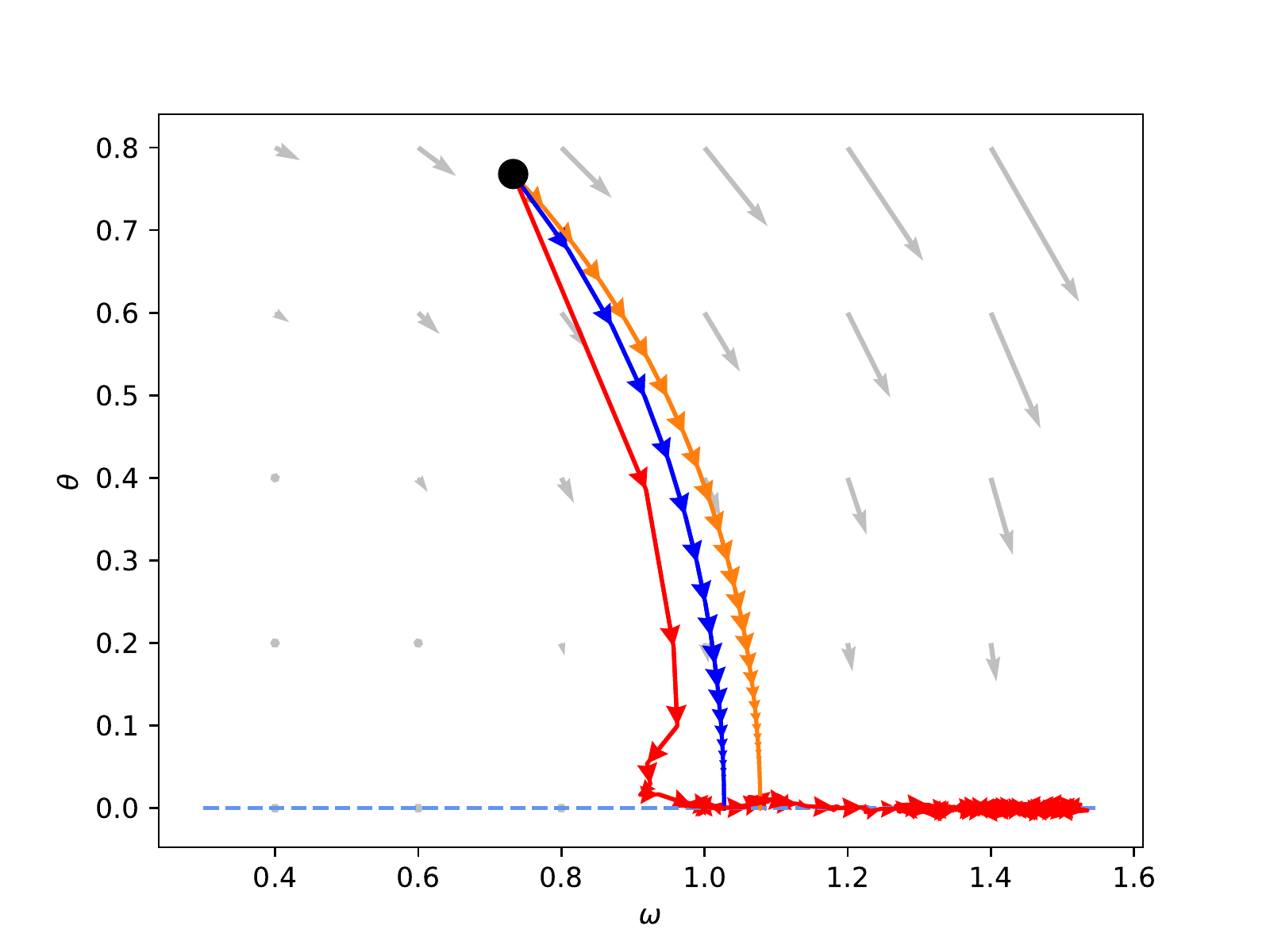}
        \caption{$(\theta_t, \omega_t)$, far from NE.}
        \label{fig:xy_2_1}
    \end{subfigure}
   ~
    \begin{subfigure}[b]{0.237\textwidth}
        \includegraphics[width=\textwidth]{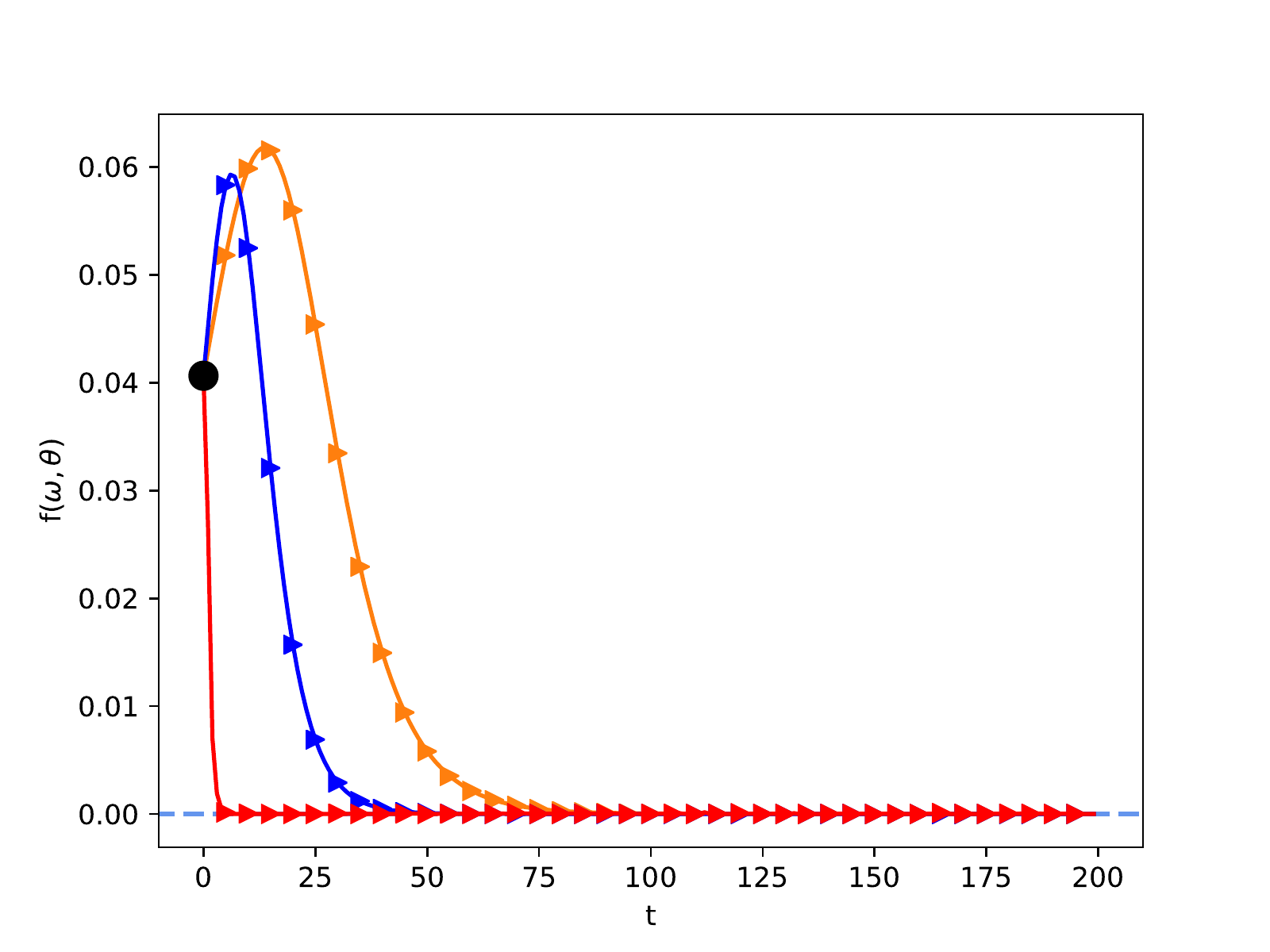}
         \caption{$f(\theta_t,\omega_t)$, close to NE.}
        \label{fig:fxy_2_2-t}
    \end{subfigure}
    ~ 
    \begin{subfigure}[b]{0.237\textwidth}
        \includegraphics[width=\textwidth]{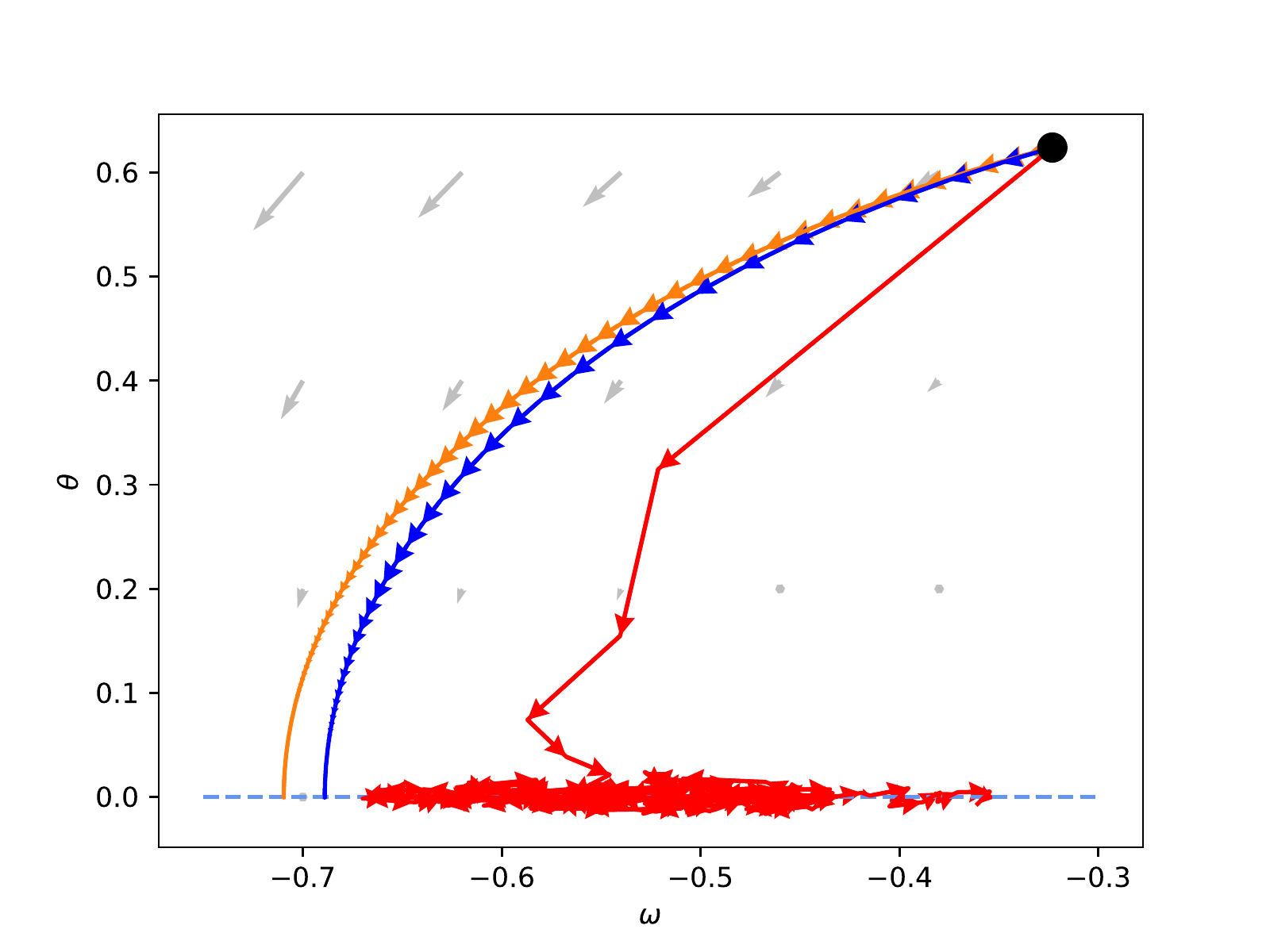}
         \caption{$(\theta_t, \omega_t)$, close to NE.}
        \label{fig:xy_2_2}
    \end{subfigure}
    \caption{$f(\theta,\omega) = \theta^2 \omega^2$. The NE are represented with the line $\{(\theta,0)\ |\ \theta \textup{ arbitrary}\}$ with reward value 0. (a), (b) shows the objective value and the training dynamics when initializing far away from NE. (c), (d) shows the objective value and the training dynamics when initializing close to NE.}\label{fig:fxy_2_2}     
     
\end{figure*}

In Section~\ref{sec:markov.game}, we have formulated the robust RL problems in either its pure strategy form~\eqref{eq:two-player-pure} and mixed strategy form~\eqref{eq:robust-bilinear-problem}. The goal of the present section is to demonstrate that solving \eqref{eq:robust-bilinear-problem} as a \emph{sampling} problem has superior performance over methods that seek pure NE for non-convex-concave SPPs. We do so by providing theoretical and empirical justifications on several simple, yet nontrivial, low-dimensional examples. Pseudocodes for all algorithms in the section and the omitted proofs can be found in Appendix~\ref{app:case-study}.

\begin{remark}
The goal of our examples is to show that sampling leads to a \textbf{stabler training algorithm}, rather than a \textbf{better solution concept}. In particular, whether mixed NE is more meaningful over pure NE (or any other equilibrium such as logistic stochastic best response equilibrium \cite{yu2019multi}) is outside of the scope of the this paper.
\end{remark}

\subsection{Existing Algorithms}
\label{subsec:case.study-algs}
We will consider three algorithmic frameworks:
\begin{enumerate}
\item GAD: Finding pure NE via \textbf{G}radient \textbf{a}scent-\textbf{d}escent.
\item EG: Finding pure NE via \textbf{E}xtra-\textbf{g}radient methods.
\item MixedNE-LD: Finding mixed NE via \textbf{Algorithm~\ref{algo:approx-infinite-md}}.
\end{enumerate}
Most existing methods to solving SPPs in deep learning can be classified as (adaptive) variants of these frameworks. For instance, Adam, being an adaptive version of GAD, is the predominant algorithm when it comes to learning GANs \cite{lucic2018gans}, which was also employed by \cite{tessler2019action} to train robust RL agents. EG was originally developed by Korpelevich in 1976 to solve variational inequalities for convex problems, and was recently shown to outperform (adaptive) GAD when it comes to training GANs \cite{gidel2018variational}. Finally, the MixedNE-LD framework was recently put forth by \cite{hsieh2018finding}, whose defining feature is to \emph{sample} from the mixed NE.

\def\drm{{\mathrm{d}}}
\def\EE{{\mathbb{E}}}
\def\RR{{\mathbb{R}}}

{It is common in practice to asymptotically decrease the step-size for GAD and EG to 0. According to the theory of \cite{panageas2019first}, these first-order methods with vanishing step-size behave asymptotically the same as their continuous-time counterpart (note that GAD and EG has the same continuous-time limit): 
\begin{equation}
\label{eq:gad.dynamics}
\begin{bmatrix}
\frac{\mathrm{d}\theta}{\drm t}(t) \\
\frac{\mathrm{d}\omega}{\drm t}(t)
\end{bmatrix} = 
\begin{bmatrix}
\nabla_\theta f(\theta,\omega) \\
-\nabla_\omega f(\theta,\omega) 
\end{bmatrix} 
\end{equation}Moreover, this result is robust to gradient noise, and so applies to stochastic variants of GAD and EG. Therefore, we will henceforth focus on \eqref{eq:gad.dynamics} in our theory.}

\subsection{Degree-2 Polynomials: Stationary Points \textit{v.s.}\ NE}
\label{subsec:degree2.poly}

We now turn to the objectives. Suppose that the objective $J$ in \eqref{eq:two-player-pure} or \eqref{eq:robust-bilinear-problem} is non-concave non-convex in $d$ directions. Since in practice one rarely acquires information higher than second-order, we will only consider quadratic local approximations of $J$. Finally, let us consider optimizing each dimension separately, each leading to a 2-dimensional subproblem. 

We will show, in \textbf{Theorem \ref{thm:gad_eg_trapped}} below, that even under this extremely simplified setting, and under simple non-convexity as in \eqref{eq:x2y2-xy} or \eqref{eq:xy-x2y2}, existing approaches can only succeed if the initialization is close enough to the equilibrium \textbf{along every direction}. As a result, the probability of successful training for existing algorithms will be exponential small in the number of non-convex non-concave directions.

We now construct nontrivial examples where there exist stationary points that are \emph{not} NE. To this end, we may simply use the degree-2 polynomials:
\begin{equation}
\max_{\theta\in[-2,2]}\min_{\omega\in [-2,2]}f(\theta,\omega) = \theta^2\omega^2 - \theta \omega \label{eq:x2y2-xy}
\end{equation}and 
\begin{equation}
\max_{\theta\in[-2,2]}\min_{\omega\in [-2,2]}f(\theta,\omega) = \theta \omega -\theta^2\omega^2. \label{eq:xy-x2y2}
\end{equation}
The constraint interval $[-2,2]$ is included only for ease of presentation; it has no impact on our conclusion. Moreover, the following facts can be readily verified:
\begin{itemize}
\item The pure and mixed NE are the same: $(0,0)$.
\item The curve $\bc{(\theta,\omega) \mid \theta \omega =0.5}$ presents stationary points that are \emph{not} NE.
\end{itemize}

Consider a single-player continuous bandit problem with 1d action space $\mathcal{A} = \mathbb{R}$, state space $\mathcal{S} = \bc{s_0}$, and reward function $R_1(s_0,a)$. For a policy $\pi_\theta (s_0) = \theta$, we have: $\max_a R_1(s_0,a) = \max_\theta R_1(s_0,\theta) = \max_{\pi_\theta} R_1(s_0,\pi_\theta (s_0))$. In light of this, one can easily see that Eqs.~\eqref{eq:x2y2-xy}~and~\eqref{eq:xy-x2y2} correspond to two-player bandit problems with $R_2(s_0,a, a') = f(\theta, \omega)$.

\subsection{Main Result}
\label{subsec:case.study-results}

We now present the main result in this section.
\begin{theorem}
\label{thm:gad_eg_trapped}
Consider the (continuous-time) GAD and EG dynamics:
\begin{equation}
\label{eq:gad.dynamics}
\begin{bmatrix}
\frac{\mathrm{d}\theta}{\drm t}(t) \\
\frac{\mathrm{d}\omega}{\drm t}(t)
\end{bmatrix} = 
\begin{bmatrix}
\nabla_\theta f(\theta,\omega) \\
-\nabla_\omega f(\theta,\omega) 
\end{bmatrix}
\end{equation}where $f(\theta,\omega)$ is either \eqref{eq:x2y2-xy} or \eqref{eq:xy-x2y2}. (Note that GAD and EG are different discretizations of the same continuous-time process \cite{diakonikolas2019approximate}.) Suppose that the initial point $(\theta(0), \omega(0))$ is far away from NE: $\theta(0)\cdot \omega(0) > 0.5$. Then \eqref{eq:gad.dynamics} converges to a non-equilibrium stationary point on $\{\theta \omega = 0.5\}$.

On the other hand, even when initialized at a stationary point such that $\theta_1\cdot \omega_1 = 0.5$, the MixedNE-LD still decreases the distance to NE in expectation: 
\begin{equation}
\label{eq:LD_decrease}
\EE \theta_3\cdot \omega_3 = \theta_1\omega_1 - 4\eta^2 \left(   \eta \left(\theta_1^2+\omega_1^2\right) + 14\eta^2 \right) < \theta_1\cdot \omega_1 
\end{equation}
where $\eta$ is the step-size, and the expectation is over the randomness of the algorithm.
\end{theorem}

In words, depending on the initialization, the (continuous-time) training dynamics of GAD and EG will either get trapped by non-equilibrium stationary points, or converge to NE. In contrast, the MixedNE-LD is always able to escape non-equilibrium stationary points in expectation.

Figures~\ref{fig:fxy_1_2}~and~\ref{fig:fxy_3_2} demonstrate the empirical behavior of the three algorithms, which is in perfect accordance with the theory. When initialized far away from NE, Figure~\eqref{fig:fxy_1_1-t}, \eqref{fig:xy_1_1-t}, \eqref{fig:fxy_3_1-t}, and \eqref{fig:xy_3_1-t} show that GAD and EG get trapped by local stationary points, while MixedNE-LD is able to escape after staying a few iterations near the non-equilibrium states. On the other hand, if initialized sufficiently close to NE, then EG tends to perform better than GAD, as indicated by previous work; see Figure~\eqref{fig:fxy_1_2-t}, \eqref{fig:xy_1_2-t}, \eqref{fig:fxy_3_2-t}, \eqref{fig:xy_3_2-t}.

Finally, one can ask whether the negative results for GAD and EG are sensitive to the choice of step-size. For instance, we have implemented the vanilla GAD and EG, while in practice one always uses adaptive step-size based on {approximate} second-order information \cite{duchi2011adaptive, kingma2014adam}. However, our next theorem shows that, even with \emph{perfect} second-order information, the training dynamics of GAD and EG still are unable to escape stationary points.
\begin{theorem}
\label{thm:second.order.doesnt.help}
Consider the Newton's dynamics for solving either \eqref{eq:x2y2-xy} or \eqref{eq:xy-x2y2}:
\begin{equation}
\label{eq:newton.dynamics}
\begin{bmatrix}
\frac{\mathrm{d}\theta}{\drm t}(t) \\
\frac{\mathrm{d}\omega}{\drm t}(t)
\end{bmatrix} =  
\begin{bmatrix}
\nabla^2_\theta f(\theta,\omega) & 0 \\
0 & \nabla^2_\omega f(\theta,\omega) ]
\end{bmatrix}^{-1}
\begin{bmatrix}
\nabla_\theta f(\theta,\omega) \\
-\nabla_\omega f(\theta,\omega) 
\end{bmatrix}.
\end{equation}Then we have $\theta(t) \cdot \omega(t) = \theta(0) \cdot \omega(0)$.
\end{theorem}
A consequence of of \textbf{Theorem~\ref{thm:second.order.doesnt.help}} is that if we initialize at any point such that $\theta(0) \cdot \omega(0) \neq 0$, the training dynamics will remain far away from $(0,0)$, which is the desired NE. Indeed, in Section~\ref{sec:experiments}, we shall see that MixedNE-LD outperforms GAD and EG even with adaptivity.
\subsection{A Digression: Sampling \textit{v.s.} Optimization}
We demonstrate an additional intriguing behavior of the sampling nature of MixedNE-LD, which we deem as a benefit over deterministic optimization algorithms. Consider the following SPP:
\begin{equation}
\max_{\theta\in[-2,2]}\min_{\omega\in [-2,2]}f(\theta,\omega) = \theta^2\omega^2. \label{eq:x2y2}
\end{equation}
This is a simple SPP where the stationary points $\{(\theta,0)\mid~\theta\in [-2,2]\}$ are all NE. Consequently, both GAD and EG succeed in finding an NE, regardless of the initialization; see Figure~\ref{fig:fxy_2_2}. The MixedNE-LD, nonetheless, does something slightly more than finding an NE: The MixedNE-LD \emph{explores} among all the NE, inducing a \emph{distribution} on the set of all equilibria; see Figure~\eqref{fig:xy_2_1}~and~\eqref{fig:xy_2_2}. 

Our theory suggests that, when there are many suboptimal stationary points, MixedNE-LD outperforms the GAD and EG baselines. We expect MixedNE-LD to perform slightly worse in the absence of this property since we can focus on converging to stationary points without adding the explorative noise in MixedNE-LD.

%% file: 4_experiments.tex

\section{Experiments}
\label{sec:experiments}

\begin{figure*}[ht!]
\centering
\includegraphics[width=0.9\linewidth]{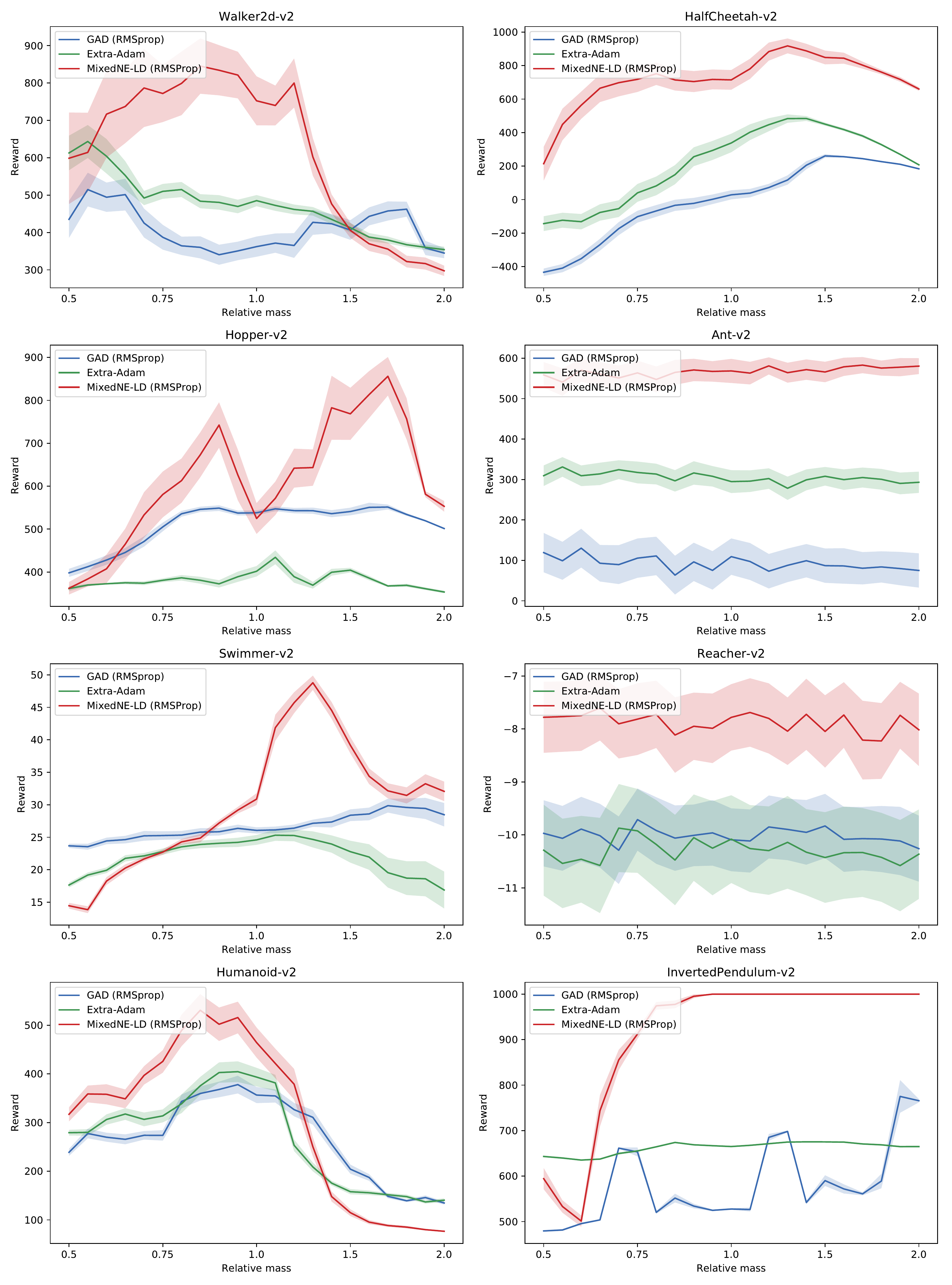}
\caption{Average performance (over 5 seeds) of Algorithm~\ref{alg:nar-sgld-ddpg} (DDPG with MixedNE-LD), and Algorithm~\ref{alg:nar-ddpg} (DDPG with GAD and Extra-Adam), under the NR-MDP setting with $\delta = 0.1$. The evaluation is performed without adversarial perturbations, on a range of mass values not encountered during training.}
\label{fig:Twoplayer_mass_comparison_average}
\end{figure*}

\begin{figure*}[ht!]
\centering
\includegraphics[width=1\linewidth,height=1\linewidth]{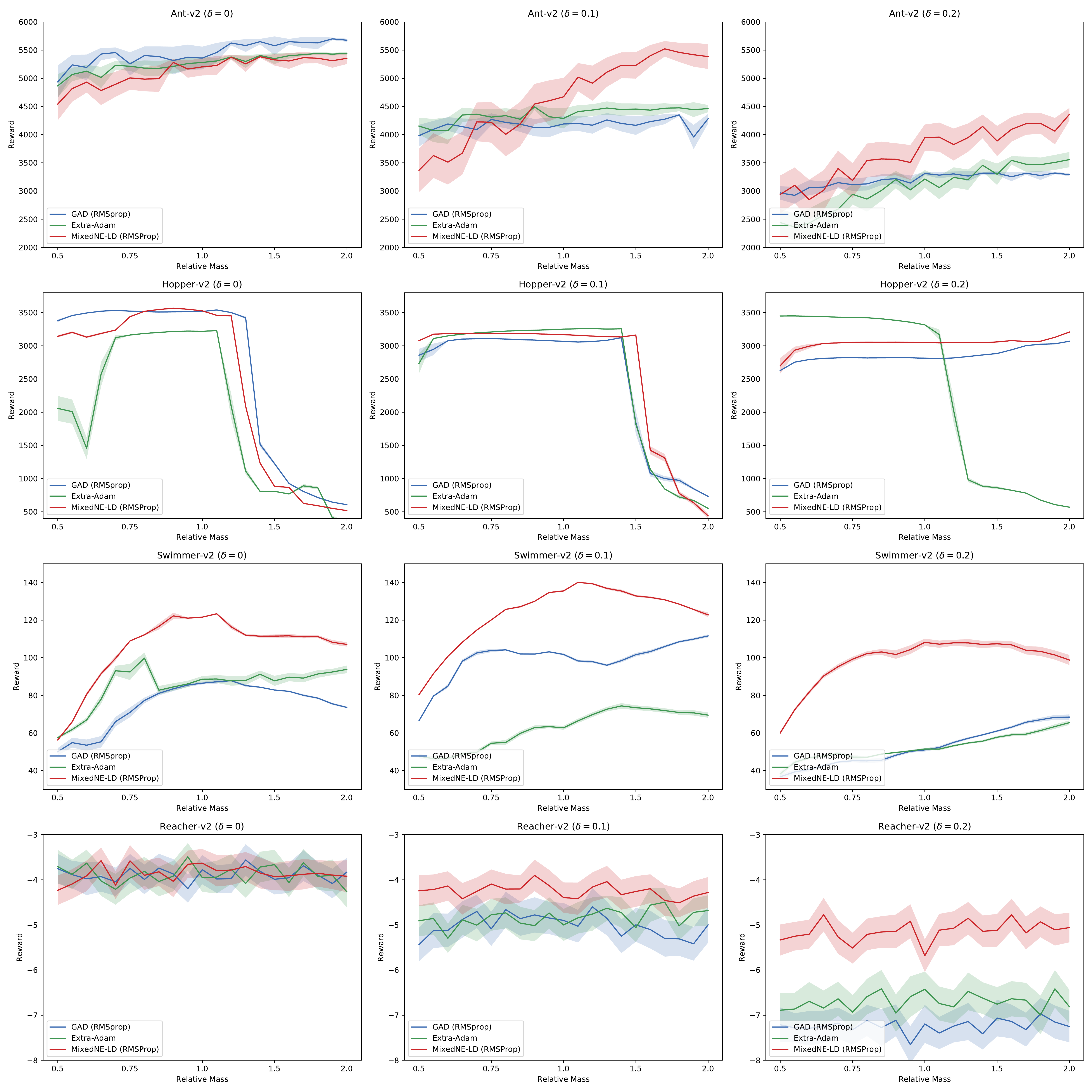}
\caption{Average performance (over 5 seeds) of Algorithm~\ref{alg:nar-sgld-td3} (TD3 with MixedNE-LD), and Algorithm~\ref{alg:nar-td3} (TD3 with GAD and Extra-Adam), under the NR-MDP setting with $\delta = 0,0.1,0.2$. The evaluation is performed without adversarial perturbations, on a range of mass values not encountered during training.}
\label{fig:td3_mass}
\end{figure*}

In this section, we demonstrate the effectiveness of using the MixedNE-LD framework to solve the robust RL problem. As a case study, we consider NR-MDP setting with $\delta = 0.1$ (as recommended in Section~6.3 of \cite{tessler2019action}). This setting can cover only the changes in the transition dynamics that can be simulated via the changes in the action. In the $H_\infty$ control literature~\cite{doyle2013feedback,morimoto2005robust}, an equivalence between environmental and action robustness has already been noted. The NR-MDP setting cannot handle: (i) the adversarial disturbances considered in~\cite{pinto2017robust}, as the action spaces of both the agent and adversary are same in the NR-MDP setting; and (ii) the feature changes like style, and illumination. Nevertheless, the MixedNE-LD framework applies to general two-player Markov Games as well.    

\paragraph{Two-Player DDPG.}
We design a two-player variant of DDPG~\cite{lillicrap2015continuous} algorithm by adapting the Algorithm~\ref{algo:approx-infinite-md}. As opposed to standard DDPG, in two-player DDPG two actor networks output two deterministic policies, the protagonist and adversary policies, denoted by $\mu_\theta$ and $\nu_{\omega}$. The critic is trained to estimate the Q-function of the joint-policy. The gradients of the protagonist and adversary parameters are given in Proposition~5 of \cite{tessler2019action}. The resulting algorithm is given in Algorithm~\ref{alg:nar-sgld-ddpg}. 

We compare the performance of our algorithm against the baseline algorithm proposed in~\cite{tessler2019action} (see Algorithm~\ref{alg:nar-ddpg} with GAD). \cite{tessler2019action} have suggested a training ratio of $1:1$ for actors and critic updates. Note that the action noise is injected while collecting transitions for the replay buffer. In \cite{fujimoto2018addressing}, authors noted that the action noise drawn from the Ornstein-Uhlenbeck \cite{uhlenbeck1930theory} process offered no performance benefits. Thus we also consider uncorrelated Gaussian noise. In addition to the baseline from~\cite{tessler2019action}, we have also considered another baseline, namely Algorithm~\ref{alg:nar-ddpg} with Extra-Adam~\cite{gidel2018variational}.

\paragraph{Setup.}
We evaluate the performance of Algorithm~\ref{alg:nar-sgld-ddpg} and Algorithm~\ref{alg:nar-ddpg} (with GAD and Extra-Adam) on standard continuous control benchmarks available on OpenAI Gym \cite{brockman2016openai} utilizing the MuJoCo environment \cite{todorov2012mujoco}. Specifically, we benchmark on eight tasks: Walker, Hopper, Half-Cheetah, Ant, Swimmer, Reacher, Humanoid, and InvertedPendulum. Details of these environments can be found in \cite{brockman2016openai} and on the GitHub website.

The Algorithm~\ref{alg:nar-sgld-ddpg} implementation is based on the codebase from~\cite{tessler2019action}. For all the algorithms, we use a two-layer feedforward neural network structure of (64, 64, tanh) for both actors (agent and adversary) and critic. The optimizer we use to update the critic is Adam \cite{kingma2015method} with a learning rate of $10^{-3}$. The target networks are soft-updated with $\tau = 0.999$. For the GAD baseline, the actors are trained with RMSProp optimizer. For our algorithm (MixedNE-LD), the actors are updated according to Algorithm~\ref{algo:approx-infinite-md} with warmup steps $K_t = \min \bc{ 15 , \floor{(1+10^{-5})^t} }$, and thermal noise $\sigma_t = \sigma_0 \times (1-5 \times 10^{-5})^t$. The hyperparameters that are not related to exploration (see Table~\ref{table:nonlin-fixed}) are identical to all the algorithms that are compared. And we tuned only the exploration-related hyperparameters (for all the algorithms) by grid search: (a) Algorithm~\ref{alg:nar-sgld-ddpg} with $\br{\sigma_0 , \sigma} \in \bc{10^{-2}, 10^{-3}, 10^{-4}, 10^{-5}} \times \bc{0, 0.01, 0.1, 0.2, 0.3, 0.4}$ ; (b) Algorithm~\ref{alg:nar-ddpg} with $\sigma \in \bc{0, 0.01, 0.1, 0.2, 0.3, 0.4}$.  For each algorithm-environment pair, we identified the best performing exploration hyperparameter configuration (see Tables~\ref{table:special-two}~and~\ref{table:special-one}). Each algorithm is trained on 0.5M samples (i.e., 0.5M time steps in the environment). We run our experiments, for each environment, with 5 different seeds. The exploration noise is turned off for evaluation.

\paragraph{Evaluation.} 
We evaluate the robustness of all the algorithms under different testing conditions, and in the presence of  adversarial disturbances in the testing environment. We train the algorithms with the standard mass and friction variables in OpenAI Gym. At test time, we evaluate the learned policies by changing the mass and friction values (without adversarial perturbations) and estimating the cumulative rewards. As shown in Figures~\ref{fig:Twoplayer_mass_comparison_average}~and~\ref{fig:Twoplayer_friction_comparison_average}, our Algorithm~\ref{alg:nar-sgld-ddpg} outperforms the baselines Algorithm~\ref{alg:nar-ddpg} (with GAD and Extra-Adam) in terms of robustness. Note that we obtain superior performance on the inverted pendulum, which is a failure case for~\cite{tessler2019action}. We also evaluate the robustness of the learned policies under both test condition changes, and adversarial disturbances (\emph{cf.} Appendix~\ref{sec:ddpg-exp-app}).

\paragraph{Additional Experiments.} 
We discuss with experimental evidence on improving the computation time of our algorithm in Appendix~\ref{sec:one-player-ddpg-app}. The One-Player DDPG with SGLD is a significant computational relaxation of DDPG with MixedNE-LD without compromising the empirical performance. We note that the MixedNE-LD can be adapted to any policy-gradient or actor-critic methods. We have already extended MixedNE-LD for TD3~\cite{fujimoto2018addressing} (cf. Appendix~\ref{sec:td3-exp-app} and Figure~\ref{fig:td3_mass}) and vanilla policy gradient~\cite{williams1992simple} (cf. Appendix~\ref{sec:vpg-exp-main}) methods. Also, we have compared the robust TD3 with MixedNE-LD against the non-robust SAC~\cite{haarnoja2018soft} algorithm (cf. Figures~\ref{fig:td3_sac_mass}~and~\ref{fig:td3_sac_friction} in the appendix). Except for the Walker environment, our robust TD3 algorithm outperforms the SAC baseline as well.

%% file: 5_conclusions.tex
\vspace{-3mm}
\section{Conclusion}
\label{sec:conclusion}
\vspace{-1mm}
In this work, we study the robust reinforcement learning problem. By adapting the approximate infinite-dimensional entropic mirror descent from~\cite{hsieh2018finding}, we design a robust variant of DDPG algorithm, under the NR-MDP setting. To the best of our knowledge, this is the first work to apply SGLD for the robust RL problem. In our experiments, we evaluate the robustness of our algorithm on several continuous control tasks, and found that our algorithm clearly outperforms the robust and non-robust baselines while tackling the failure case (i.e., inverted pendulum) for the earlier literature. Intriguingly, even the simple version of the algorithm with a single Langevin step results in competitive results with a desirable computational complexity.

%% file: 7_appendix.tex

\section{Appendix structure}
This Appendix provides additional proofs and experimental results. Here we provide an overview on the organization of the section.
\begin{itemize}
\item Appendix~\ref{app:case-study} reports omitted proofs of Section~\ref{sec:case.study}. 
\item Appendix~\ref{sec:one-player-ddpg-app} discusses the computational improvement for DDPG with MixedNE-LD.  
\item Appendix~\ref{sec:ddpg-exp-app} contains DDPG experimental results and hyperparameter details.
\item Appendix~\ref{sec:td3-exp-app} contains TD3 experimental results and hyperparameter details.
\item Appendix~\ref{sec:vpg-exp-main} discusses the experimental setup for VPG experiments. 
\item Appendix~\ref{sec:vpg-exp-app} contains VPG experimental results and hyperparameter details.
\end{itemize}

The code repository (for all the experiments):~\url{https://github.com/DaDaCheng/LIONS-RL/tree/master/Robust-Reinforcement-Learning-via-Adversarial-training-with-Langevin-Dynamics}. 

\input{8_appendix-case-study}

\newpage

\section{One-Player DDPG with SGLD}
\label{sec:one-player-ddpg-app}

\begin{figure*}[ht!]
\centering
\includegraphics[width=1\linewidth,height=1.4\linewidth]{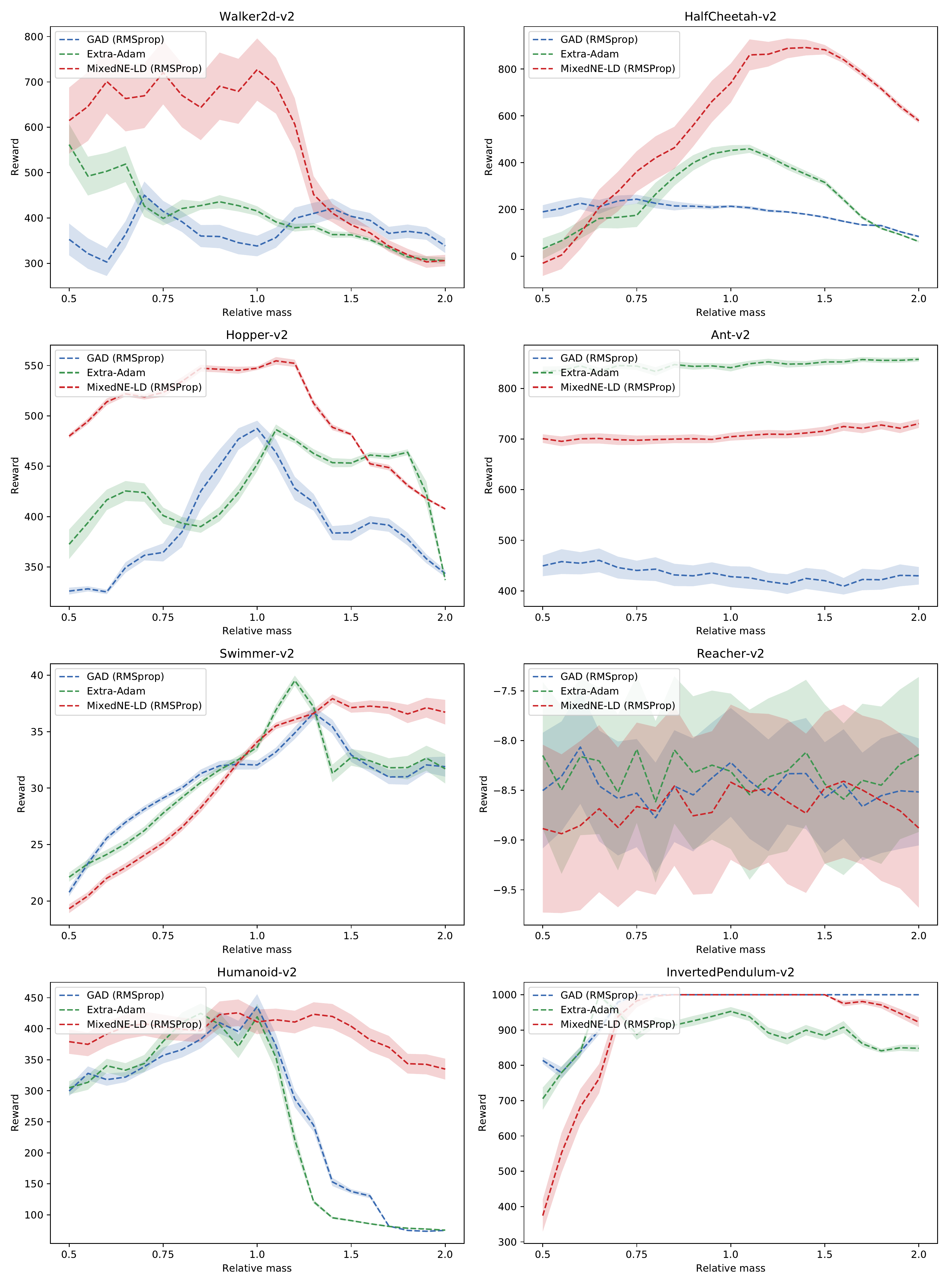}
\caption{Average performance (over 5 seeds) of Algorithm~\ref{alg:nar-sgld-ddpg} (DDPG with MixedNE-LD), and Algorithm~\ref{alg:nar-ddpg} (DDPG with GAD and Extra-Adam), under the NR-MDP setting with $\delta = 0$. The evaluation is performed without adversarial perturbations, on a range of mass values not encountered during training.}
\label{fig:Oneplayer_mass_comparison_average}
\end{figure*}

\begin{figure*}[ht!]
\centering
\includegraphics[width=1\linewidth,height=1.4\linewidth]{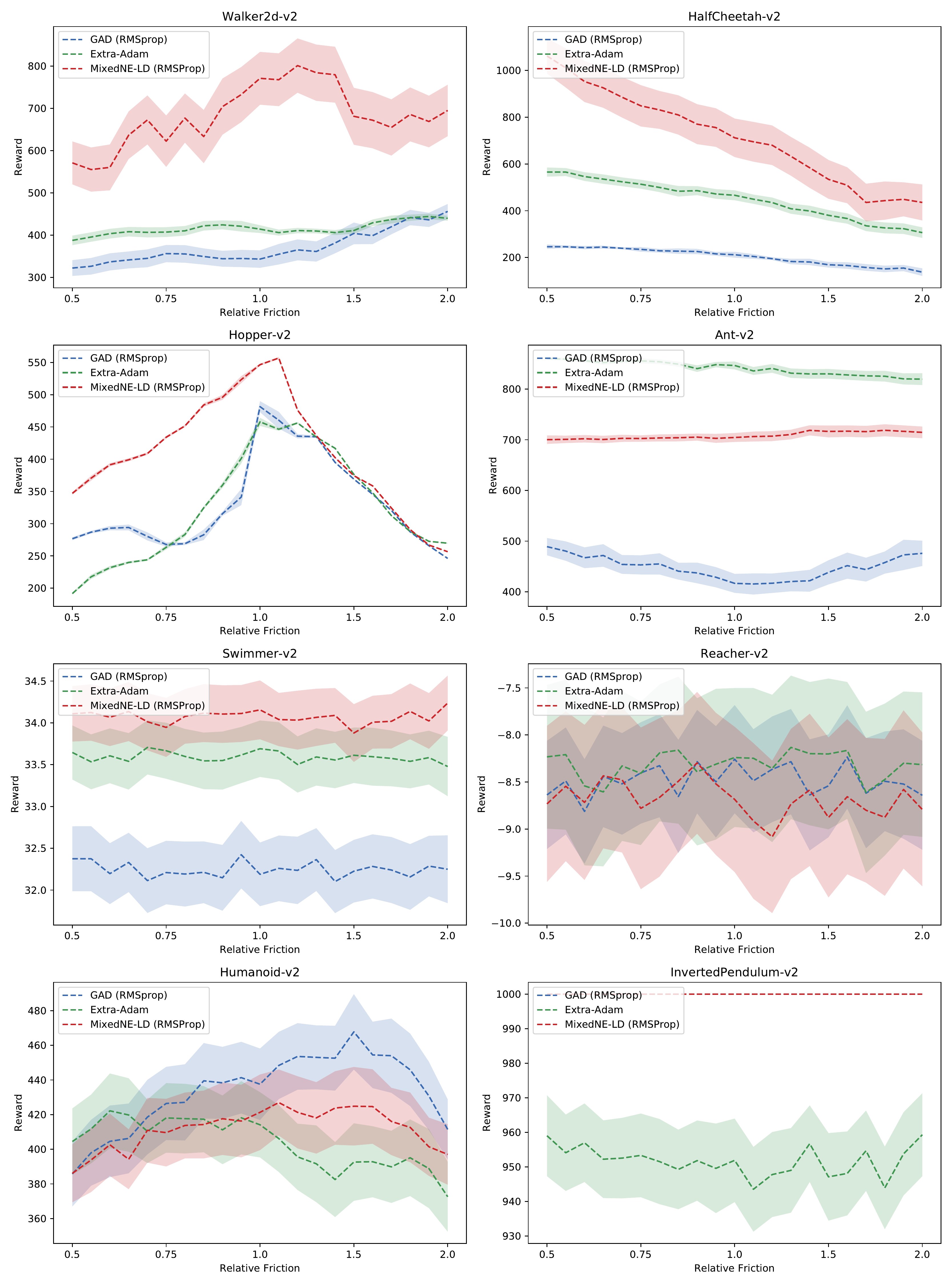}
\caption{Average performance (over 5 seeds) of Algorithm~\ref{alg:nar-sgld-ddpg} (DDPG with MixedNE-LD), and Algorithm~\ref{alg:nar-ddpg} (DDPG with GAD and Extra-Adam), under the NR-MDP setting with $\delta = 0$. The evaluation is performed without adversarial perturbations, on a range of friction values not encountered during training.}
\label{fig:Oneplayer_friction_comparison_average}
\end{figure*}

We evaluate the robustness of one-player variants of Algorithm~\ref{alg:nar-sgld-ddpg}, and Algorithm~\ref{alg:nar-ddpg} (with GAD and Extra-Adam), i.e., we consider the NR-MDP setting with $\delta = 0$. In this case, we set $K_t = 1$ for Algorithm~\ref{alg:nar-sgld-ddpg} (this choice of $K_t$ makes the computational complexity of both algorithms equal). The corresponding results are presented in Figures~\ref{fig:Oneplayer_mass_comparison_average}~and~\ref{fig:Oneplayer_friction_comparison_average} (also \emph{cf.}  Appendix~\ref{sec:ddpg-exp-app}).

Here, we remark that Algorithm~\ref{alg:nar-sgld-ddpg} with $\delta = 0$, and $K_t = 1$ is simply the standard DDPG with actor being updated by preconditioned version of SGLD. Thus we achieve robustness under different testing conditions with just a simple change in the DDPG algorithm and without additional computational cost.

\subsection{Robustness of One-Player MixedNE-LD}
\label{sec:one-player-mix}

Consider the standard (non-robust) RL objective of maximizing $J \br{\theta} = \Ex{\sum_{t=1}^{\infty}{\gamma^{t-1}r_t} ~\vert~ \mu_\theta , \mathcal{M}_1}$. We can translate this non-convex problem into an infinite dimensional convex-problem by considering a distribution over deterministic policies as follows~\cite{liu2017stein}:
\[
\max_{p \in \mathcal{P}\br{\Theta}} ~ \Ee{\theta \sim p}{J\br{\theta}} + \lambda H\br{p} ,
\]
where $H\br{p} = \Ee{\theta \sim p}{ - \log p\br{\theta}}$ is the entropy of the distribution $p$. The robust behavior of this objective (in the context of loss surface) is discussed in~\cite{chaudhari2019entropy}. The optimal solution to the above problem takes the form: $p_\lambda^*\br{\theta} \propto \exp \br{\frac{1}{\lambda} J \br{\theta}}$. For a given $\lambda$, SGLD can be used to draw samples from $p_\lambda^*\br{\theta}$. 

Then, the resulting algorithm is equivalent to Algorithm~\ref{alg:nar-sgld-ddpg} with $\delta = 0$. Note that in our one-player DDPG experiments, we obtained significant improvement over both robust and non-robust baselines even with single inner loop iteration ($K_t = 1$). Since the Algorithm~\ref{alg:nar-sgld-ddpg} is computationally demanding even though it uses a mean approximation in its inner loop, this new approximation by setting $\delta = 0$ and $K_t = 1$ is preferred in practice. 

\section{DDPG Experiments: Algorithms, Hyperparameters, and Results}
\label{sec:ddpg-exp-app}

\begin{figure*}[t!]
	\centering
	\begin{subfigure}[t]{1\textwidth}
		\centering
		\includegraphics[width=\linewidth]{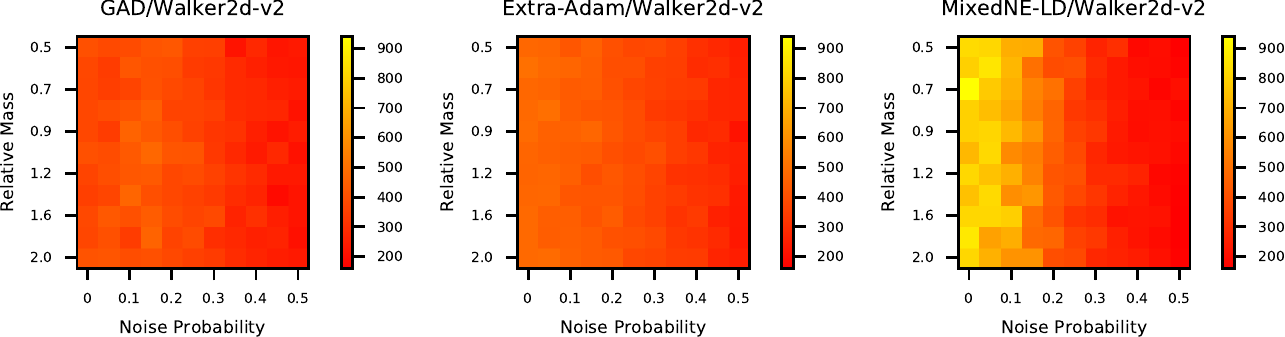}
		\label{fig:Heat_map_Walker}
	\end{subfigure}%
    \\
	\begin{subfigure}[t]{1\textwidth}
		\centering
		\includegraphics[width=\linewidth]{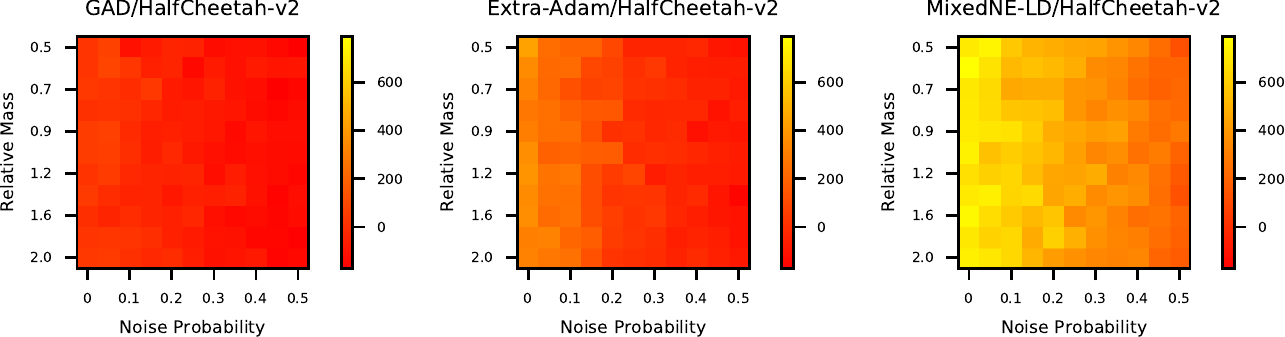}
		\label{fig:Heat_map_HalfCheetah}
	\end{subfigure}%
    \\
    	\begin{subfigure}[t]{1\textwidth}
		\centering
		\includegraphics[width=\linewidth]{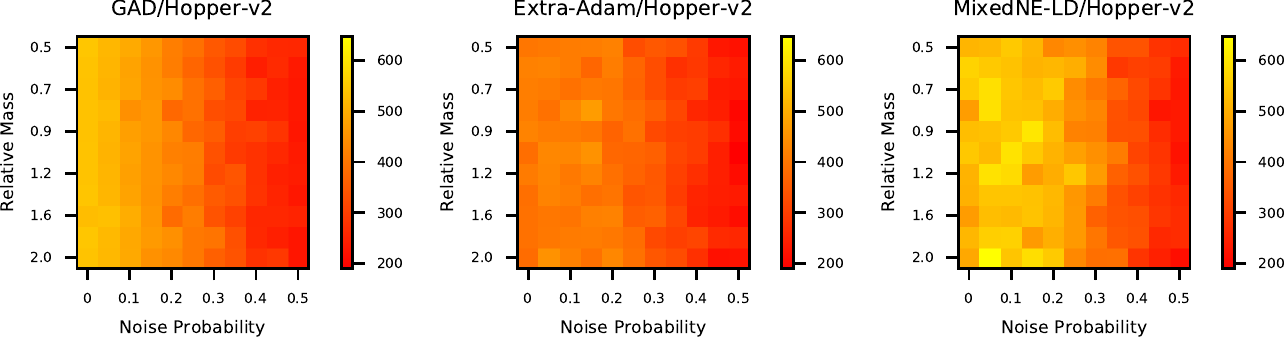}
		\label{fig:Heat_map_Hopper}
	\end{subfigure}%
    \\
	\begin{subfigure}[t]{1\textwidth}
		\centering
		\includegraphics[width=\linewidth]{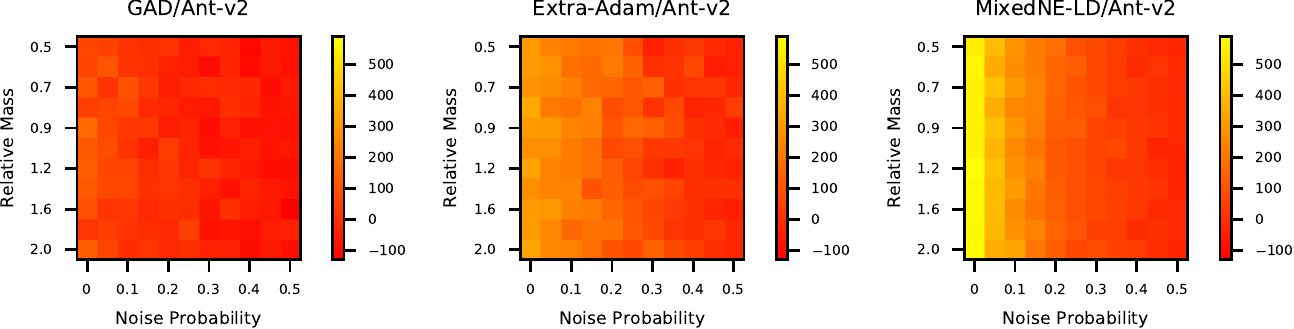}
		\label{fig:Heat_map_Ant}
	\end{subfigure}

	\caption{Average performance (over 5 seeds) of Algorithm~\ref{alg:nar-sgld-ddpg}, and Algorithm~\ref{alg:nar-ddpg} (with GAD and Extra-Adam), under the NR-MDP setting with $\delta = 0.1$. The evaluation is performed on a range of noise probability and mass values not encountered during training. Environments: Walker, HalfCheetah, Hopper, and Ant.}
	\label{fig:TwoPlayer_Heat_map_comparison_average_a}
\end{figure*}

\begin{figure*}[t!]
	\centering
	    	\begin{subfigure}[t]{1\textwidth}
		\centering
		\includegraphics[width=\linewidth]{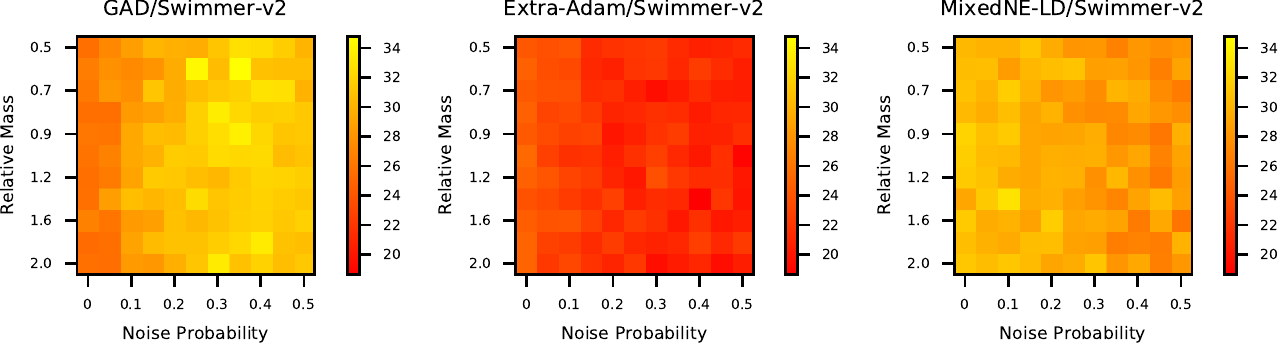}
		\label{fig:Heat_map_Swimmer}
	\end{subfigure}%
    \\
	\begin{subfigure}[t]{1\textwidth}
		\centering
		\includegraphics[width=\linewidth]{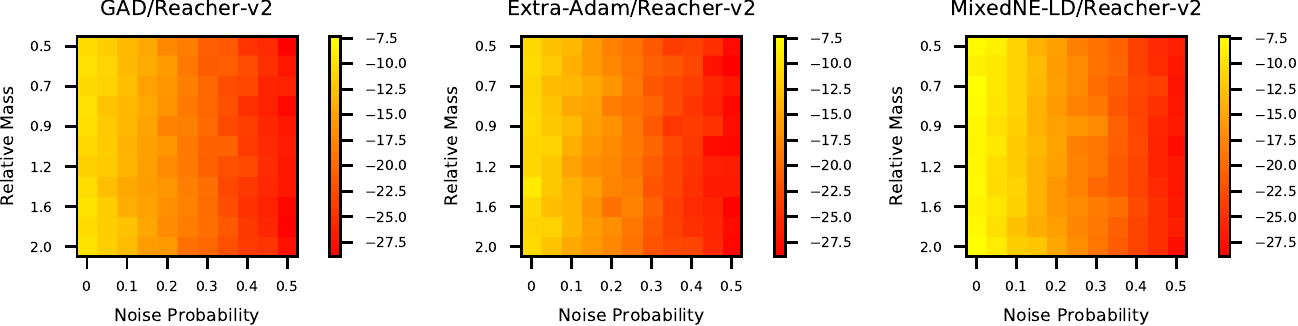}
		\label{fig:Heat_map_Reacher}
	\end{subfigure}%
    \\
    	\begin{subfigure}[t]{1\textwidth}
		\centering
		\includegraphics[width=\linewidth]{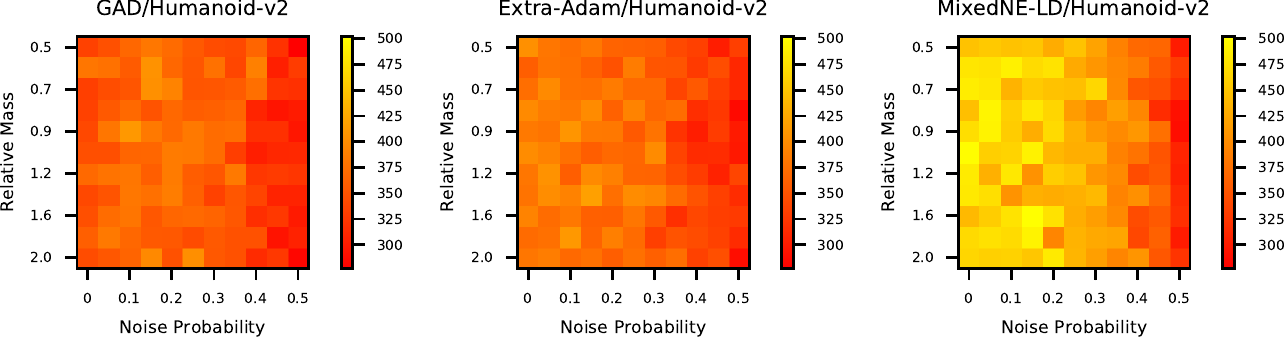}
		\label{fig:Heat_map_Humanoid}
	\end{subfigure}%
    \\
	\begin{subfigure}[t]{1\textwidth}
		\centering
		\includegraphics[width=\linewidth]{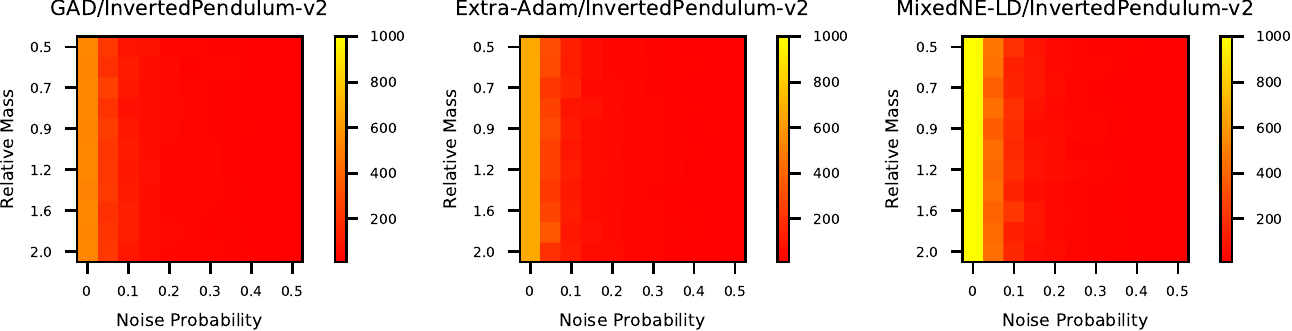}
		\label{fig:Heat_map_InvertedPendulum}
	\end{subfigure}

	\caption{Average performance (over 5 seeds) of Algorithm~\ref{alg:nar-sgld-ddpg}, and Algorithm~\ref{alg:nar-ddpg} (with GAD and Extra-Adam), under the NR-MDP setting with $\delta = 0.1$. The evaluation is performed on a range of noise probability and mass values not encountered during training. Environments: Swimmer, Reacher, Humanoid, and InvertedPendulum.}
	\label{fig:TwoPlayer_Heat_map_comparison_average_b}
\end{figure*}

\begin{figure*}[t!]
	\centering
	\begin{subfigure}[t]{1\textwidth}
		\centering
		\includegraphics[width=\linewidth]{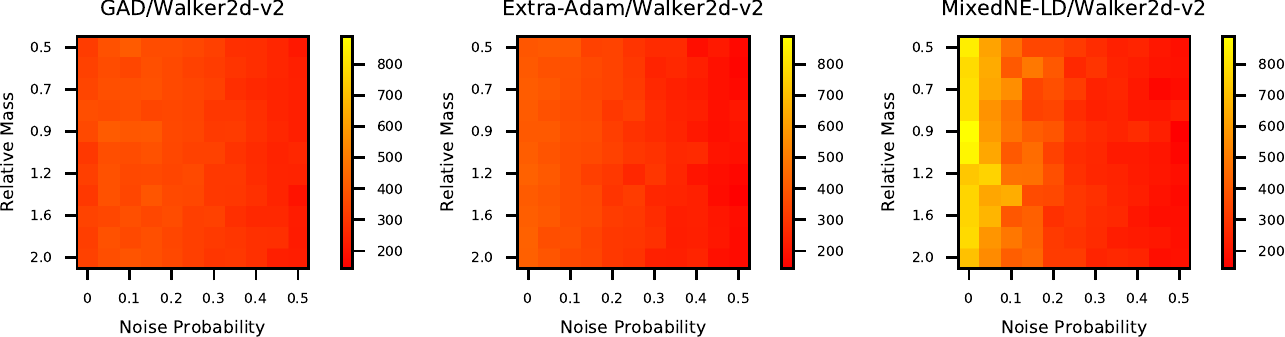}
		\label{fig:Heat_map_Walker-1}
	\end{subfigure}%
    \\
	\begin{subfigure}[t]{1\textwidth}
		\centering
		\includegraphics[width=\linewidth]{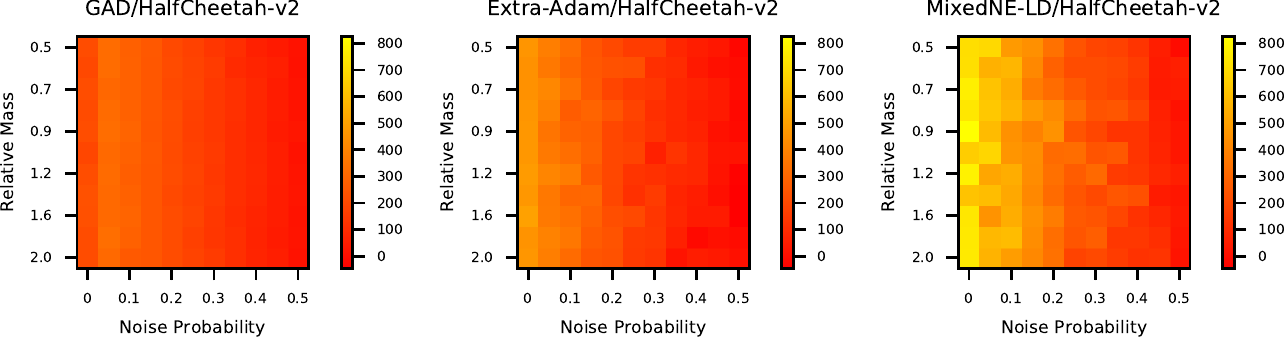}
		\label{fig:Heat_map_HalfCheetah-1}
	\end{subfigure}%
    \\
    	\begin{subfigure}[t]{1\textwidth}
		\centering
		\includegraphics[width=\linewidth]{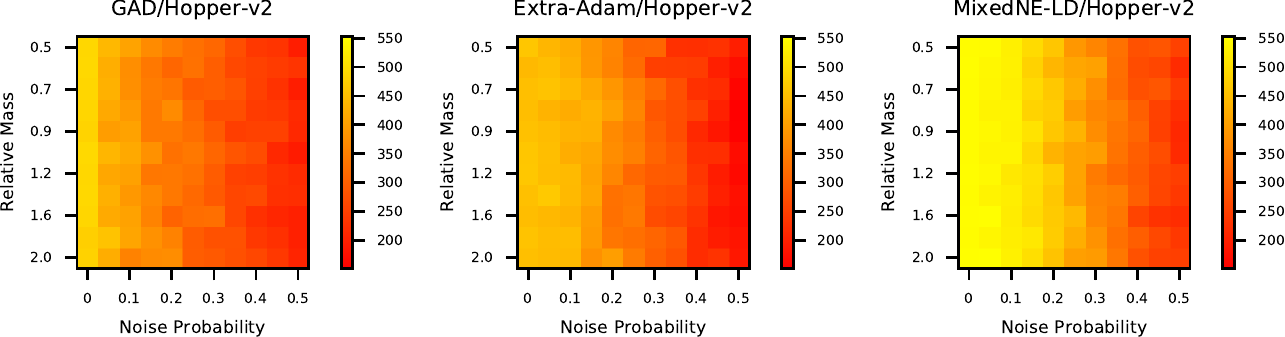}
		\label{fig:Heat_map_Hopper-1}
	\end{subfigure}%
    \\
	\begin{subfigure}[t]{1\textwidth}
		\centering
		\includegraphics[width=\linewidth]{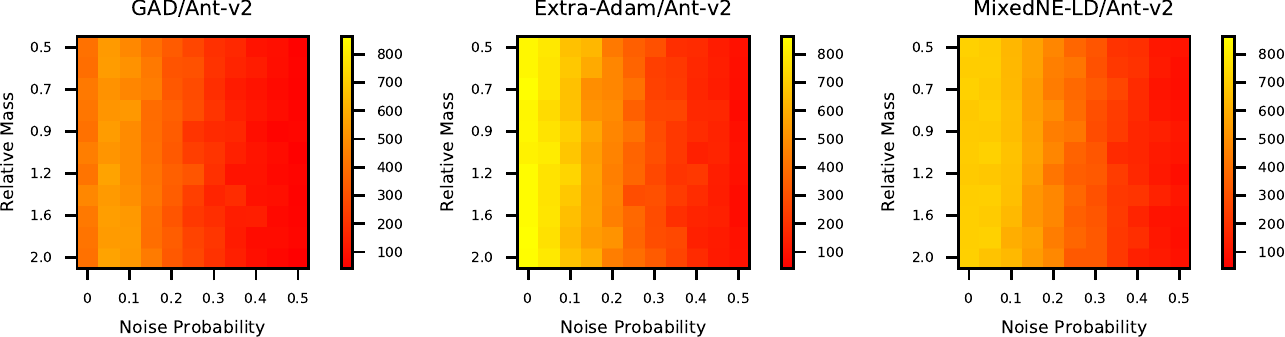}
		\label{fig:Heat_map_Ant-1}
	\end{subfigure}

	\caption{Average performance (over 5 seeds) of Algorithm~\ref{alg:nar-sgld-ddpg}, and Algorithm~\ref{alg:nar-ddpg} (with GAD and Extra-Adam), under the NR-MDP setting with $\delta = 0$. The evaluation is performed on a range of noise probability and mass values not encountered during training. Environments: Walker, HalfCheetah, Hopper, and Ant.}
	\label{fig:OnePlayer_Heat_map_comparison_average_a}
\end{figure*}

\begin{figure*}[t!]
	\centering
	    	\begin{subfigure}[t]{1\textwidth}
		\centering
		\includegraphics[width=\linewidth]{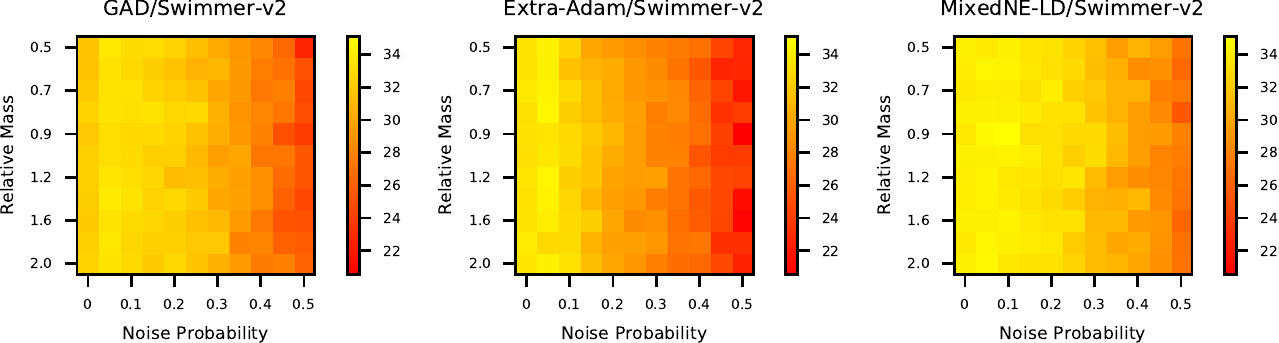}
		\label{fig:Heat_map_Swimmer-1}
	\end{subfigure}%
    \\
	\begin{subfigure}[t]{1\textwidth}
		\centering
		\includegraphics[width=\linewidth]{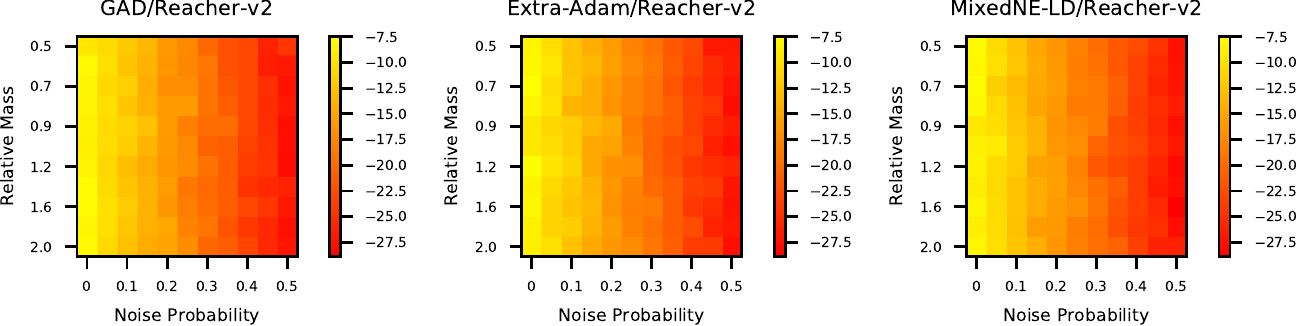}
		\label{fig:Heat_map_Reacher-1}
	\end{subfigure}%
    \\
    	\begin{subfigure}[t]{1\textwidth}
		\centering
		\includegraphics[width=\linewidth]{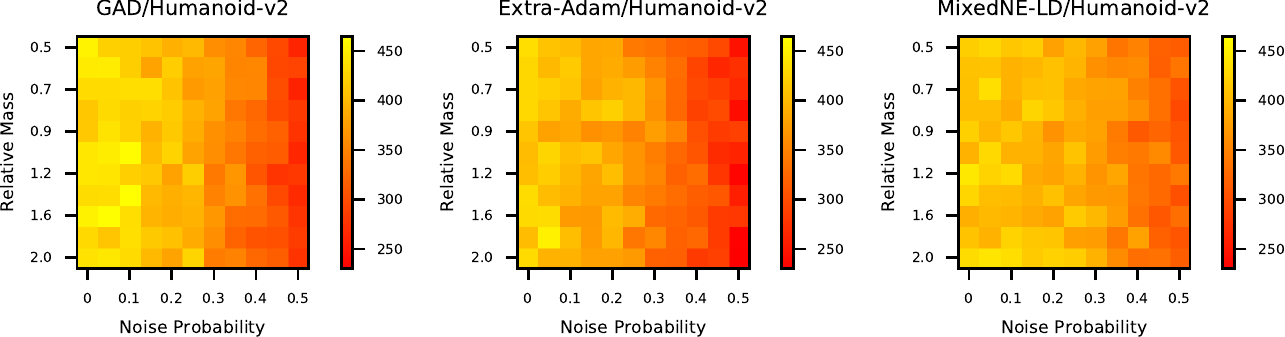}
		\label{fig:Heat_map_Humanoid-1}
	\end{subfigure}%
    \\
	\begin{subfigure}[t]{1\textwidth}
		\centering
		\includegraphics[width=\linewidth]{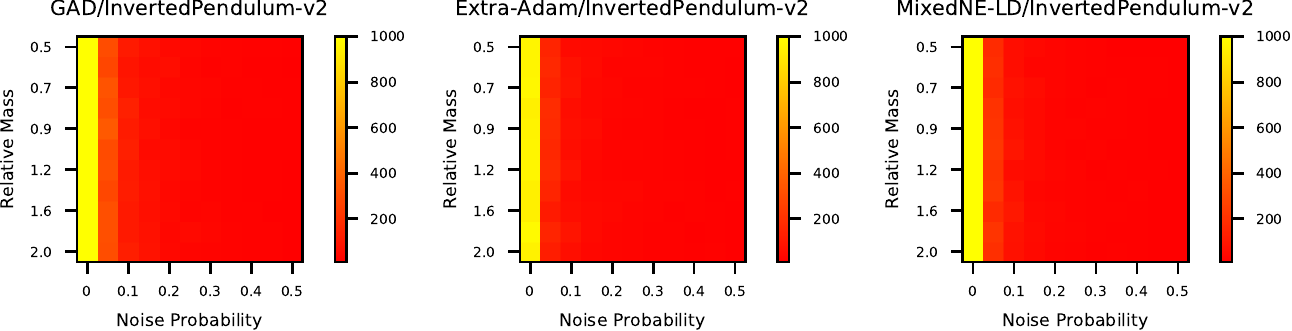}
		\label{fig:Heat_map_InvertedPendulum-1}
	\end{subfigure}

	\caption{Average performance (over 5 seeds) of Algorithm~\ref{alg:nar-sgld-ddpg}, and Algorithm~\ref{alg:nar-ddpg} (with GAD and Extra-Adam), under the NR-MDP setting with $\delta = 0$. The evaluation is performed on a range of noise probability and mass values not encountered during training. Environments: Swimmer, Reacher, Humanoid, and InvertedPendulum.}
	\label{fig:OnePlayer_Heat_map_comparison_average_b}
\end{figure*}

\begin{figure*}[t!]
	\centering
	\begin{subfigure}[t]{1\textwidth}
		\centering
		\includegraphics[width=\linewidth]{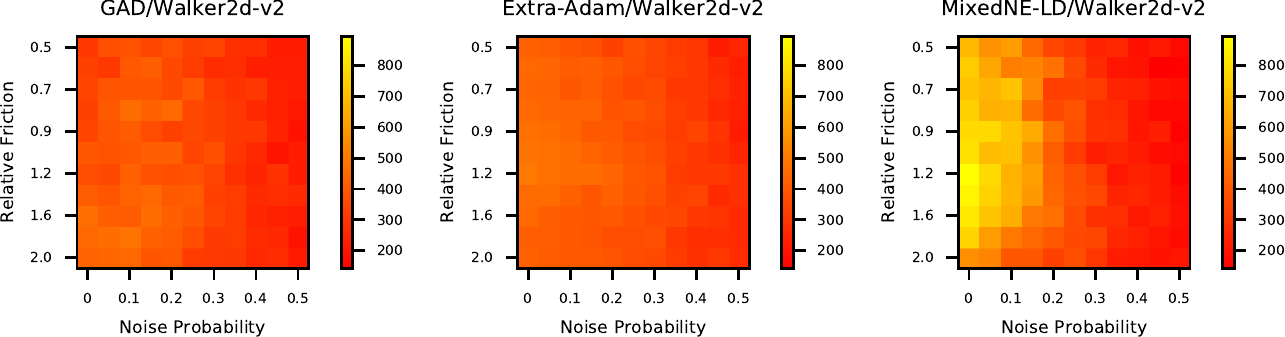}
		\label{fig:Heat_map_Walker_friction}
	\end{subfigure}%
    \\
	\begin{subfigure}[t]{1\textwidth}
		\centering
		\includegraphics[width=\linewidth]{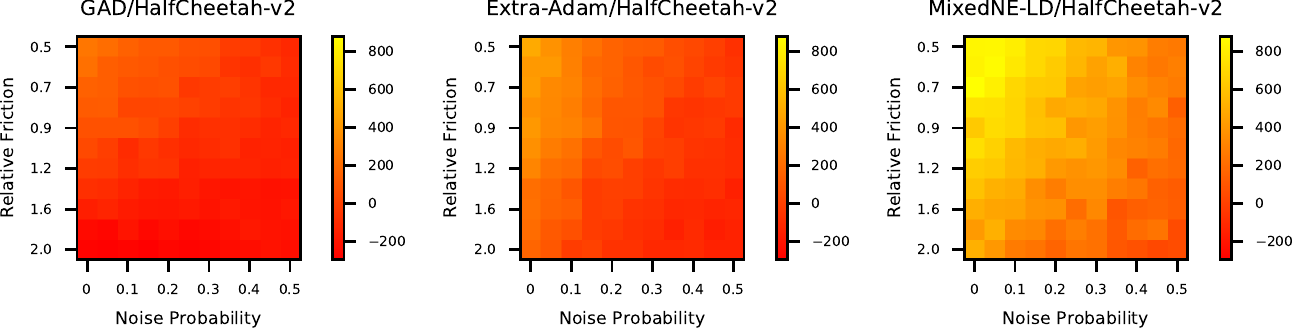}
		\label{fig:Heat_map_HalfCheetah_friction}
	\end{subfigure}%
    \\
    	\begin{subfigure}[t]{1\textwidth}
		\centering
		\includegraphics[width=\linewidth]{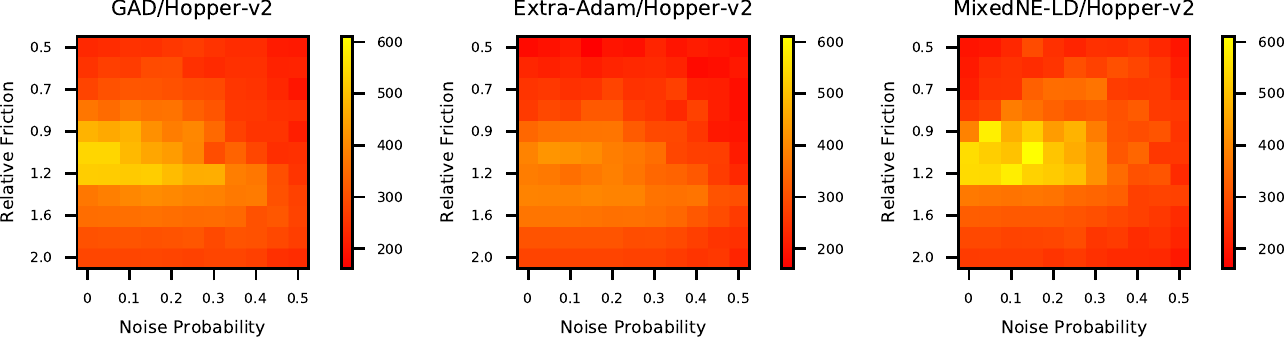}
		\label{fig:Heat_map_Hopper_friction}
	\end{subfigure}%
    \\
	\begin{subfigure}[t]{1\textwidth}
		\centering
		\includegraphics[width=\linewidth]{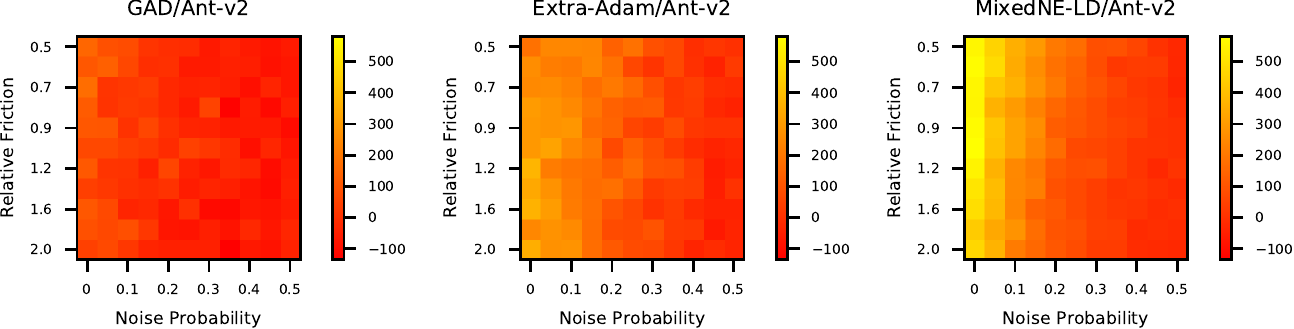}
		\label{fig:Heat_map_Ant_friction}
	\end{subfigure}

	\caption{Average performance (over 5 seeds) of Algorithm~\ref{alg:nar-sgld-ddpg}, and Algorithm~\ref{alg:nar-ddpg} (with GAD and Extra-Adam), under the NR-MDP setting with $\delta = 0.1$. The evaluation is performed on a range of noise probability and friction values not encountered during training. Environments: Walker, HalfCheetah, Hopper, and Ant.}
	\label{fig:TwoPlayer_Heat_map_friction_comparison_average_a}
\end{figure*}

\begin{figure*}[t!]
	\centering
	    	\begin{subfigure}[t]{1\textwidth}
		\centering
		\includegraphics[width=\linewidth]{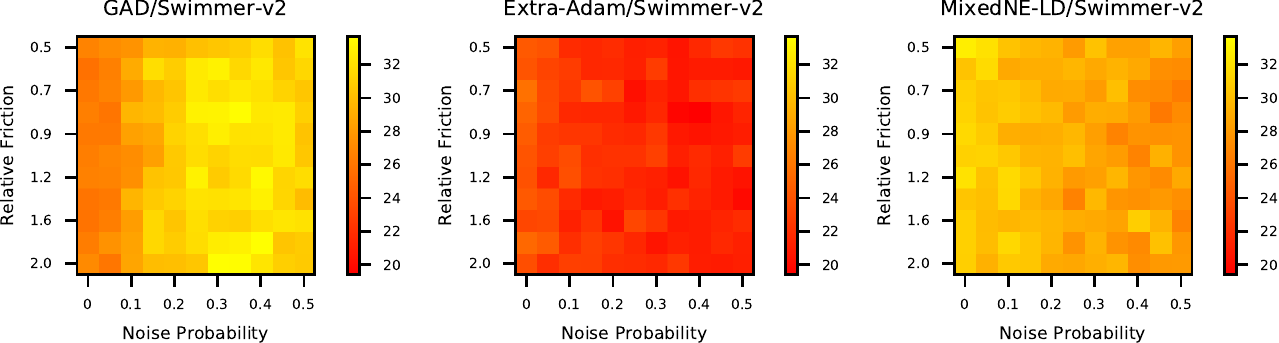}
		\label{fig:Heat_map_Swimmer_friction}
	\end{subfigure}%
    \\
	\begin{subfigure}[t]{1\textwidth}
		\centering
		\includegraphics[width=\linewidth]{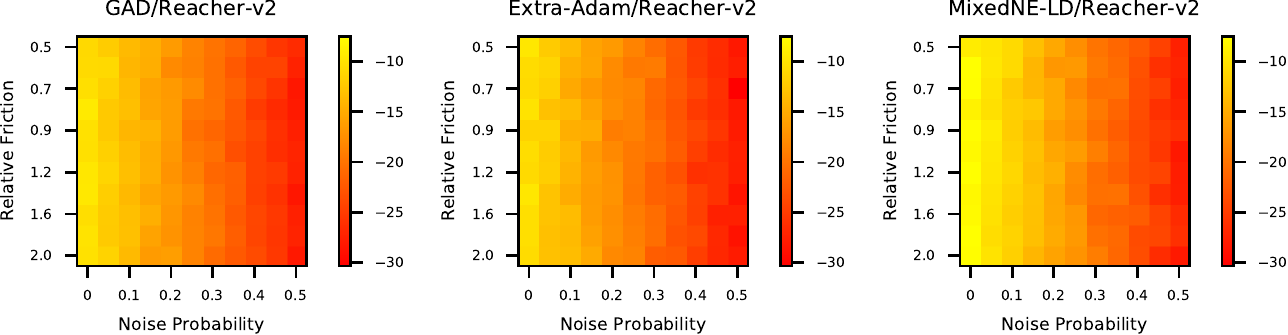}
		\label{fig:Heat_map_Reacher_friction}
	\end{subfigure}%
    \\
    	\begin{subfigure}[t]{1\textwidth}
		\centering
		\includegraphics[width=\linewidth]{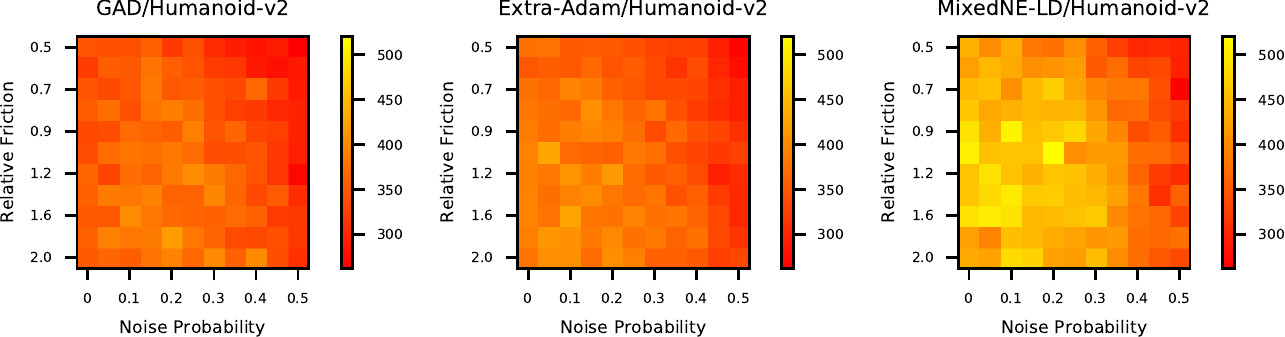}
		\label{fig:Heat_map_Humanoid_friction}
	\end{subfigure}%
    \\
	\begin{subfigure}[t]{1\textwidth}
		\centering
		\includegraphics[width=\linewidth]{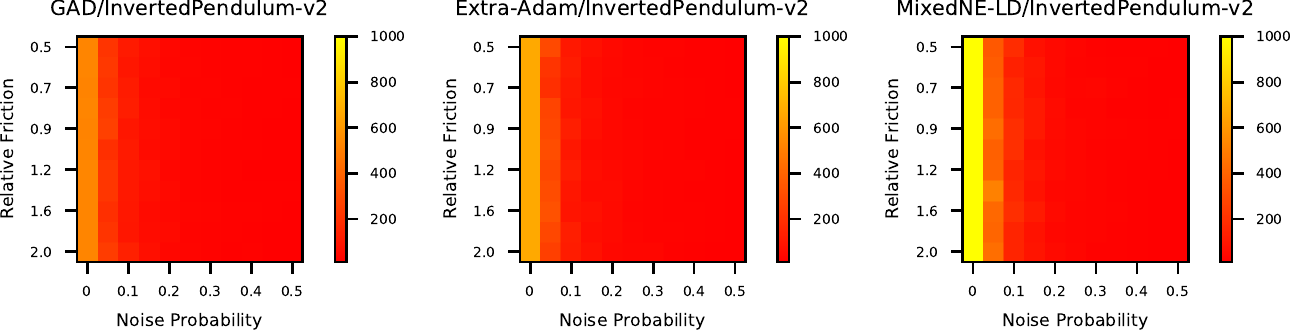}
		\label{fig:Heat_map_InvertedPendulum_friction}
	\end{subfigure}

	\caption{Average performance (over 5 seeds) of Algorithm~\ref{alg:nar-sgld-ddpg}, and Algorithm~\ref{alg:nar-ddpg} (with GAD and Extra-Adam), under the NR-MDP setting with $\delta = 0.1$. The evaluation is performed on a range of noise probability and friction values not encountered during training. Environments: Swimmer, Reacher, Humanoid, and InvertedPendulum.}
	\label{fig:TwoPlayer_Heat_map_friction_comparison_average_b}
\end{figure*}

\begin{figure*}[t!]
	\centering
	\begin{subfigure}[t]{1\textwidth}
		\centering
		\includegraphics[width=\linewidth]{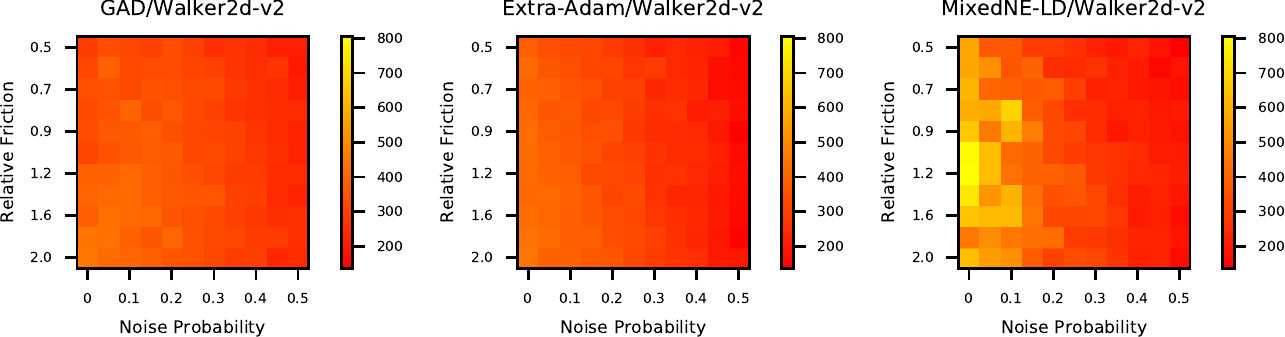}
		\label{fig:Heat_map_Walker-1_friction}
	\end{subfigure}%
    \\
	\begin{subfigure}[t]{1\textwidth}
		\centering
		\includegraphics[width=\linewidth]{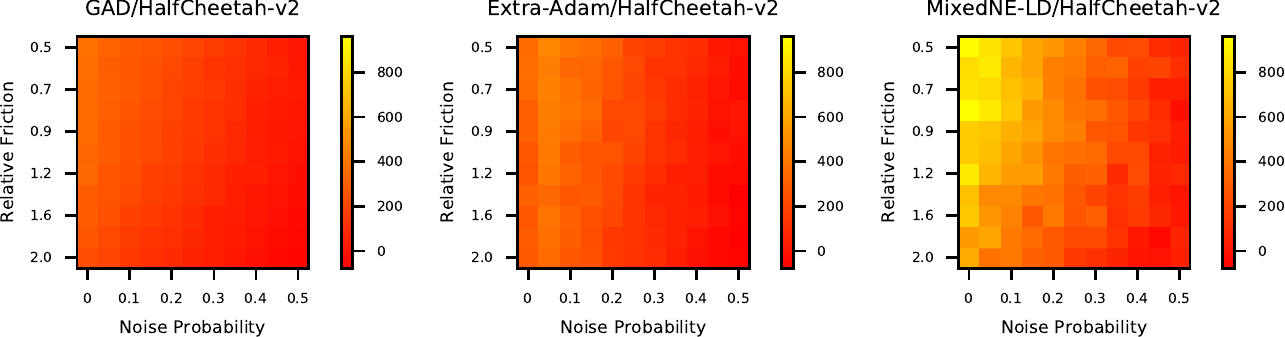}
		\label{fig:Heat_map_HalfCheetah-1_friction}
	\end{subfigure}%
    \\
    	\begin{subfigure}[t]{1\textwidth}
		\centering
		\includegraphics[width=\linewidth]{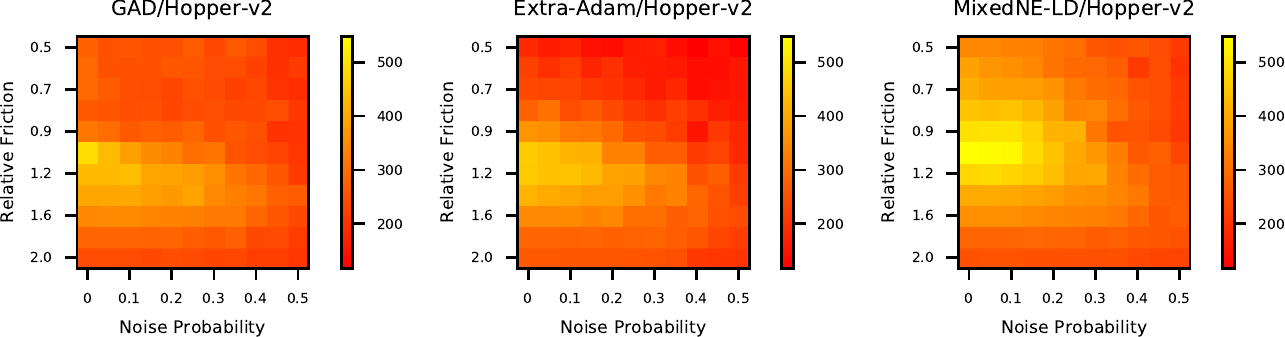}
		\label{fig:Heat_map_Hopper-1_friction}
	\end{subfigure}%
    \\
	\begin{subfigure}[t]{1\textwidth}
		\centering
		\includegraphics[width=\linewidth]{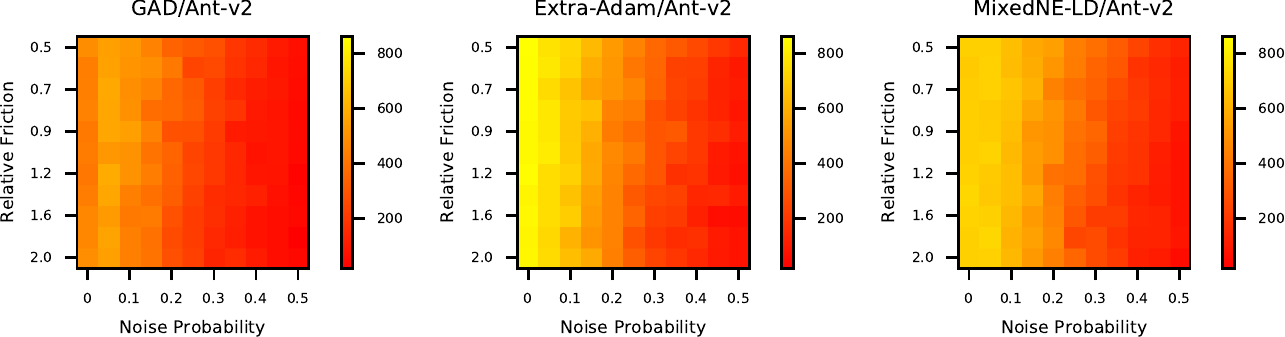}
		\label{fig:Heat_map_Ant-1_friction}
	\end{subfigure}

	\caption{Average performance (over 5 seeds) of Algorithm~\ref{alg:nar-sgld-ddpg}, and Algorithm~\ref{alg:nar-ddpg} (with GAD and Extra-Adam), under the NR-MDP setting with $\delta = 0$. The evaluation is performed on a range of noise probability and friction values not encountered during training. Environments: Walker, HalfCheetah, Hopper, and Ant.}
	\label{fig:OnePlayer_Heat_map_friction_comparison_average_a}
\end{figure*}

\begin{figure*}[t!]
	\centering
	    	\begin{subfigure}[t]{1\textwidth}
		\centering
		\includegraphics[width=\linewidth]{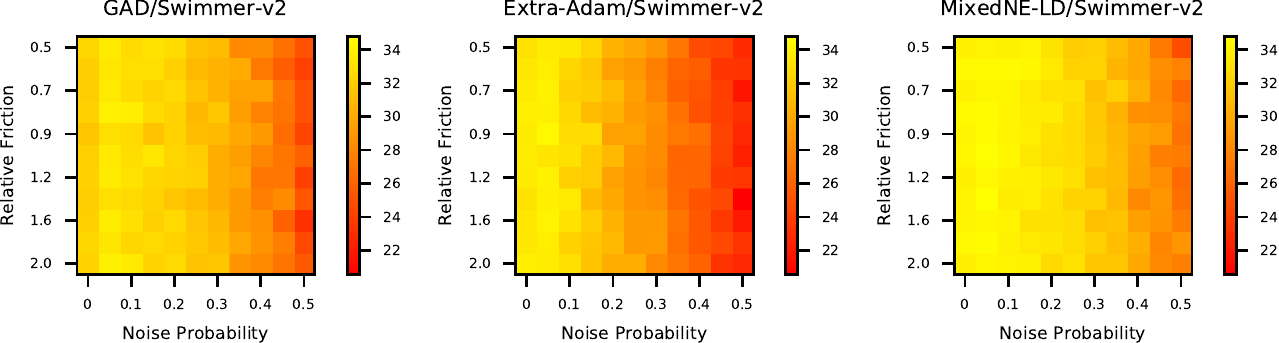}
		\label{fig:Heat_map_Swimmer-1_friction}
	\end{subfigure}%
    \\
	\begin{subfigure}[t]{1\textwidth}
		\centering
		\includegraphics[width=\linewidth]{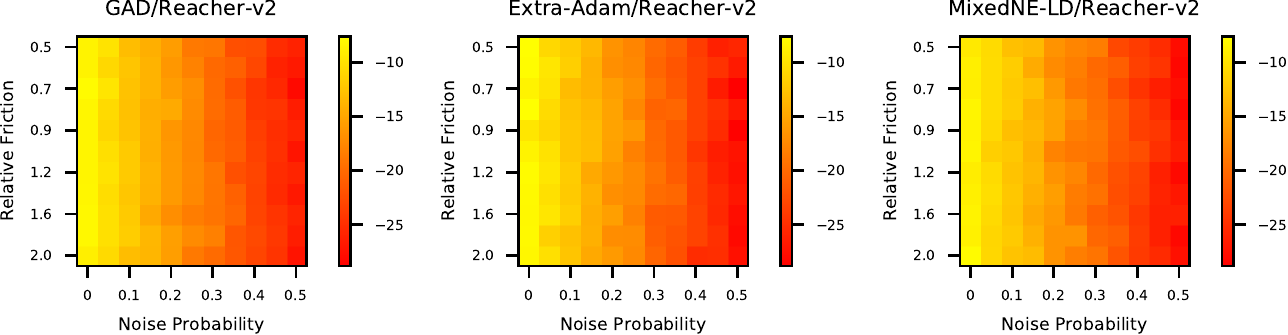}
		\label{fig:Heat_map_Reacher-1_friction}
	\end{subfigure}%
    \\
    	\begin{subfigure}[t]{1\textwidth}
		\centering
		\includegraphics[width=\linewidth]{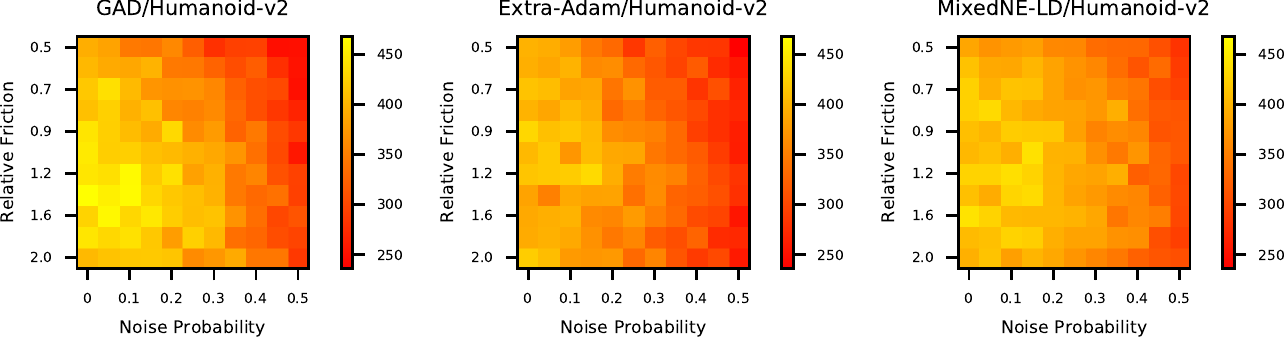}
		\label{fig:Heat_map_Humanoid-1_friction}
	\end{subfigure}%
    \\
	\begin{subfigure}[t]{1\textwidth}
		\centering
		\includegraphics[width=\linewidth]{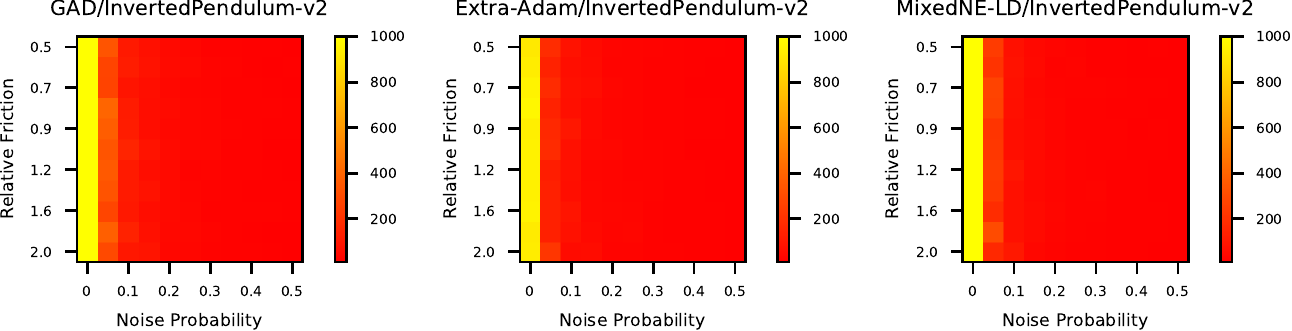}
		\label{fig:Heat_map_InvertedPendulum-1_friction}
	\end{subfigure}

	\caption{Average performance (over 5 seeds) of Algorithm~\ref{alg:nar-sgld-ddpg}, and Algorithm~\ref{alg:nar-ddpg} (with GAD and Extra-Adam), under the NR-MDP setting with $\delta = 0$. The evaluation is performed on a range of noise probability and friction values not encountered during training. Environments: Swimmer, Reacher, Humanoid, and InvertedPendulum.}
	\label{fig:OnePlayer_Heat_map_friction_comparison_average_b}
\end{figure*}

\begin{figure*}[t!]
	\centering
	\begin{subfigure}[t]{1\textwidth}
		\centering
		\includegraphics[width=\linewidth]{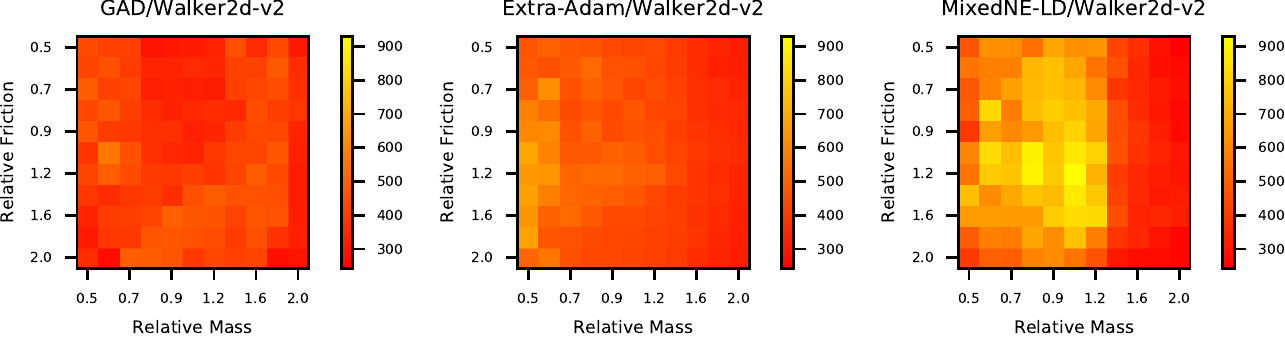}
		\label{fig:Heat_map_Walker_both}
	\end{subfigure}%
    \\
	\begin{subfigure}[t]{1\textwidth}
		\centering
		\includegraphics[width=\linewidth]{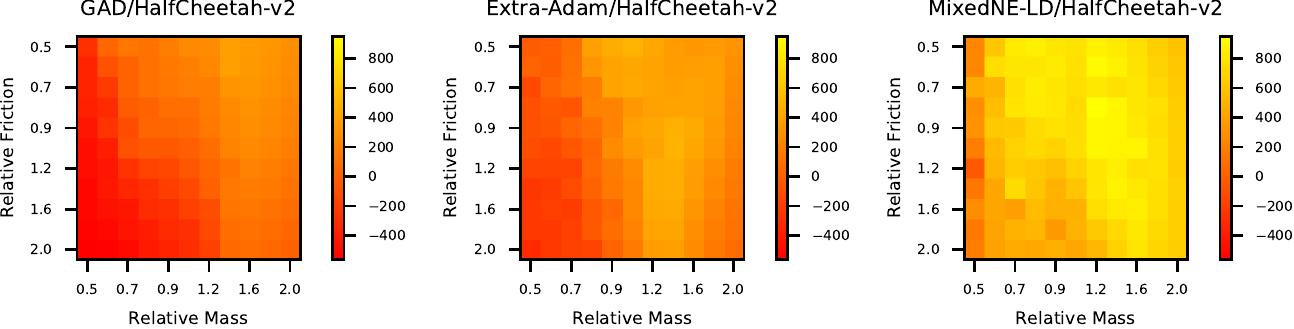}
		\label{fig:Heat_map_HalfCheetah_both}
	\end{subfigure}%
    \\
    	\begin{subfigure}[t]{1\textwidth}
		\centering
		\includegraphics[width=\linewidth]{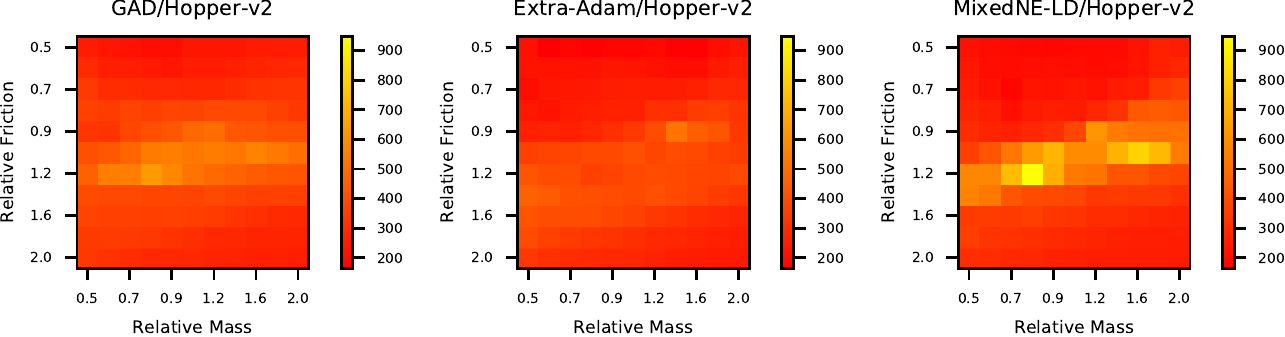}
		\label{fig:Heat_map_Hopper_both}
	\end{subfigure}%
    \\
	\begin{subfigure}[t]{1\textwidth}
		\centering
		\includegraphics[width=\linewidth]{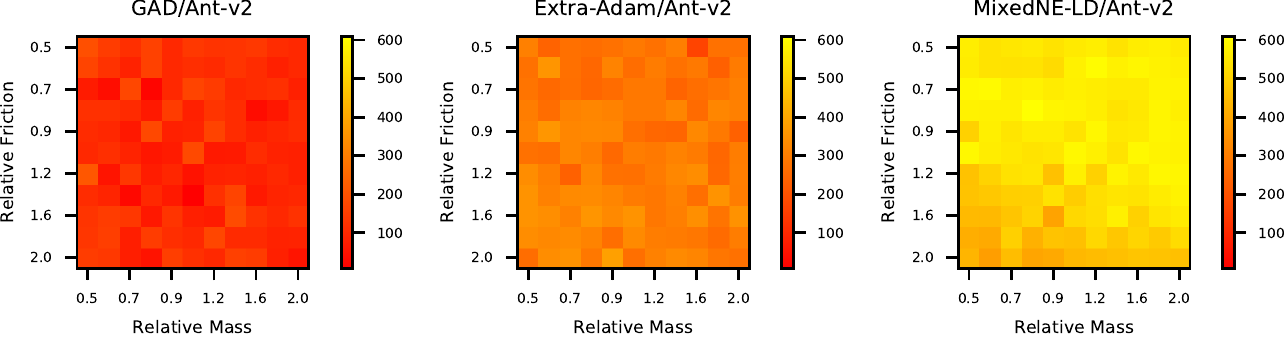}
		\label{fig:Heat_map_Ant_both}
	\end{subfigure}

	\caption{Average performance (over 5 seeds) of Algorithm~\ref{alg:nar-sgld-ddpg}, and Algorithm~\ref{alg:nar-ddpg} (with GAD and Extra-Adam), under the NR-MDP setting with $\delta = 0.1$. The evaluation is performed on a range of friction and mass values not encountered during training. Environments: Walker, HalfCheetah, Hopper, and Ant.}
	\label{fig:TwoPlayer_Heat_map_both_comparison_average_a}
\end{figure*}

\begin{figure*}[t!]
	\centering
	    	\begin{subfigure}[t]{1\textwidth}
		\centering
		\includegraphics[width=\linewidth]{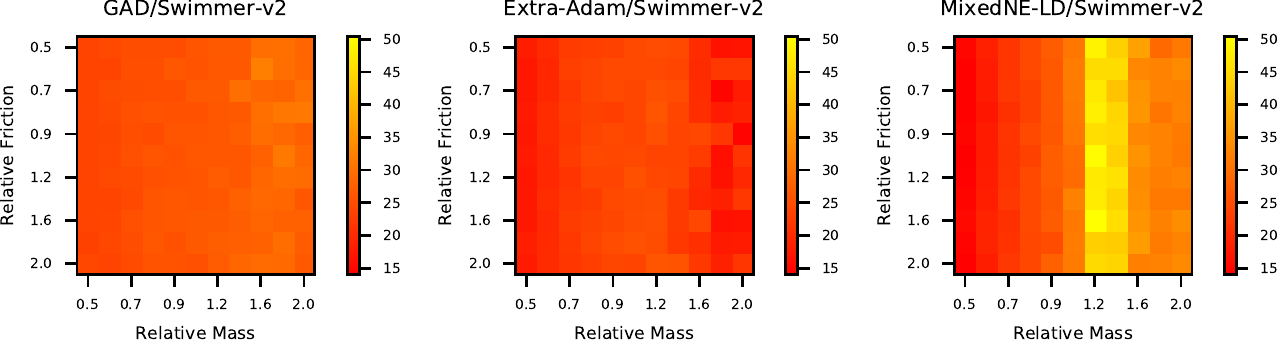}
		\label{fig:Heat_map_Swimmer_both}
	\end{subfigure}%
    \\
	\begin{subfigure}[t]{1\textwidth}
		\centering
		\includegraphics[width=\linewidth]{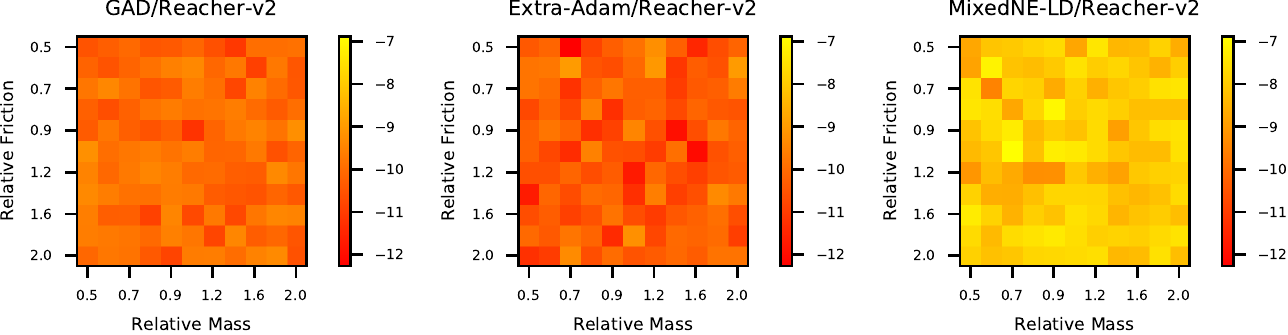}
		\label{fig:Heat_map_Reacher_both}
	\end{subfigure}%
    \\
    	\begin{subfigure}[t]{1\textwidth}
		\centering
		\includegraphics[width=\linewidth]{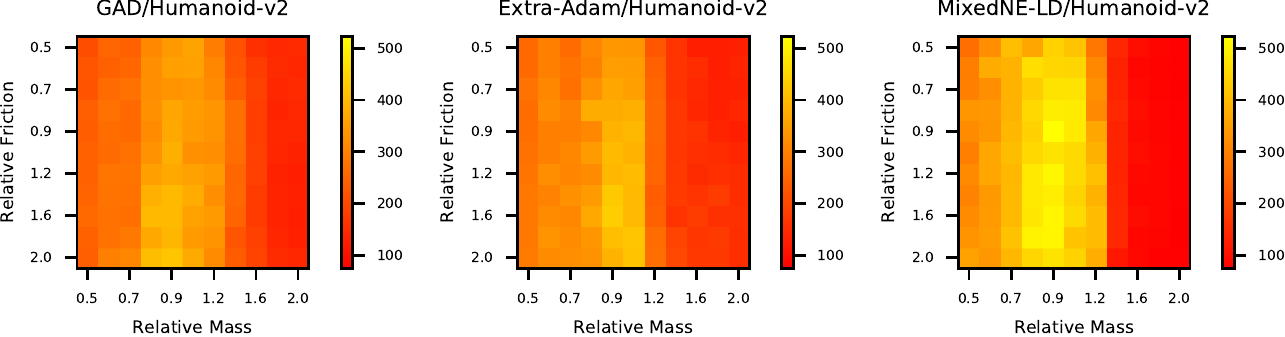}
		\label{fig:Heat_map_Humanoid_both}
	\end{subfigure}%
    \\
	\begin{subfigure}[t]{1\textwidth}
		\centering
		\includegraphics[width=\linewidth]{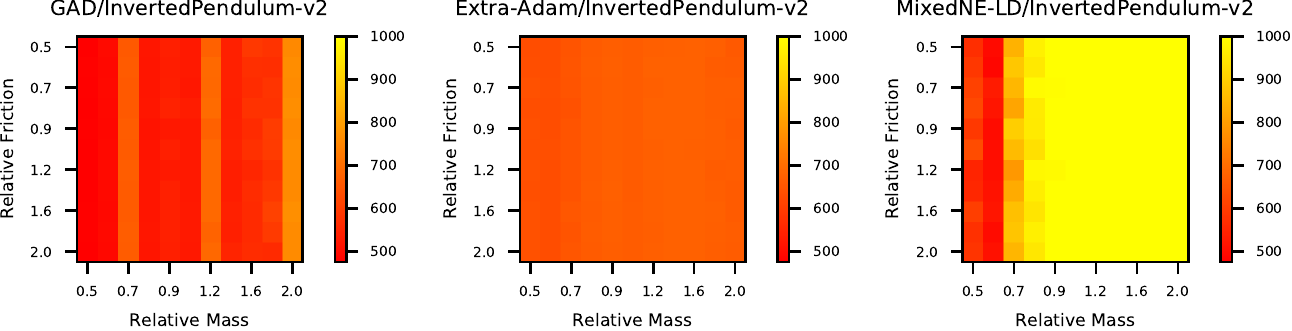}
		\label{fig:Heat_map_InvertedPendulum_both}
	\end{subfigure}

	\caption{Average performance (over 5 seeds) of Algorithm~\ref{alg:nar-sgld-ddpg}, and Algorithm~\ref{alg:nar-ddpg} (with GAD and Extra-Adam), under the NR-MDP setting with $\delta = 0.1$. The evaluation is performed on a range of friction and mass values not encountered during training. Environments: Swimmer, Reacher, Humanoid, and InvertedPendulum.}
	\label{fig:TwoPlayer_Heat_map_both_comparison_average_b}
\end{figure*}

\begin{figure*}[t!]
	\centering
	\begin{subfigure}[t]{1\textwidth}
		\centering
		\includegraphics[width=\linewidth]{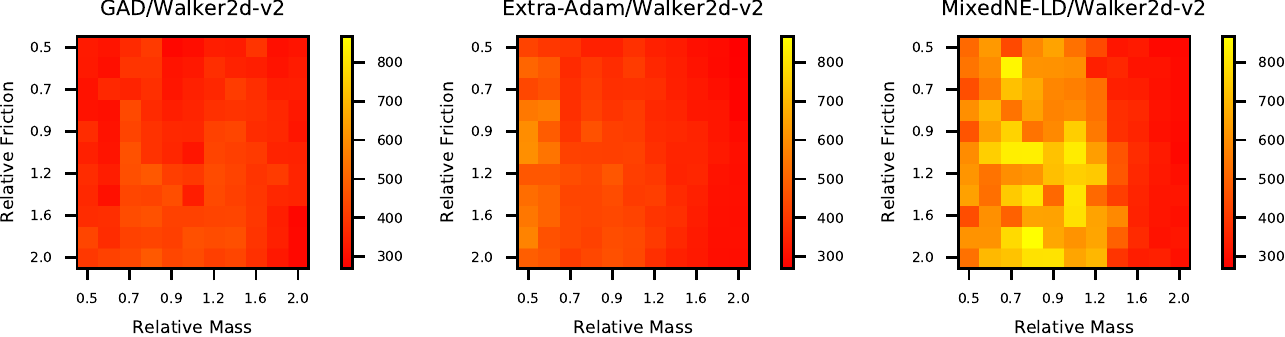}
		\label{fig:Heat_map_Walker-1_both}
	\end{subfigure}%
    \\
	\begin{subfigure}[t]{1\textwidth}
		\centering
		\includegraphics[width=\linewidth]{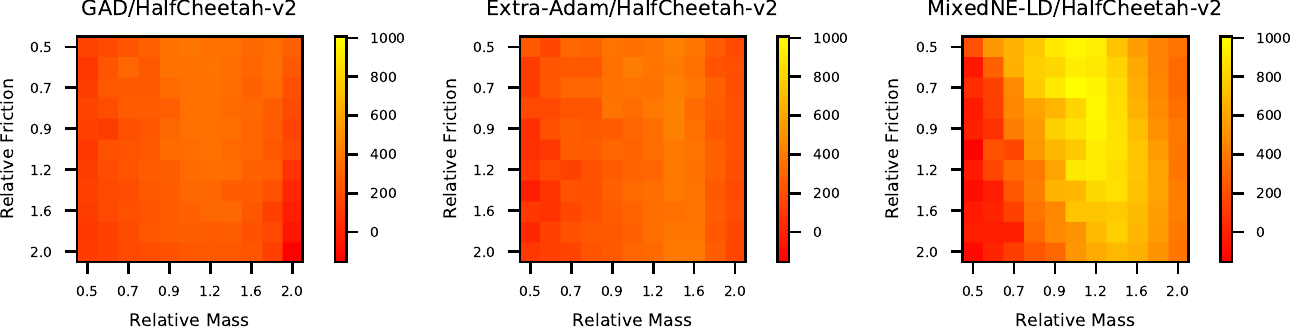}
		\label{fig:Heat_map_HalfCheetah-1_both}
	\end{subfigure}%
    \\
    	\begin{subfigure}[t]{1\textwidth}
		\centering
		\includegraphics[width=\linewidth]{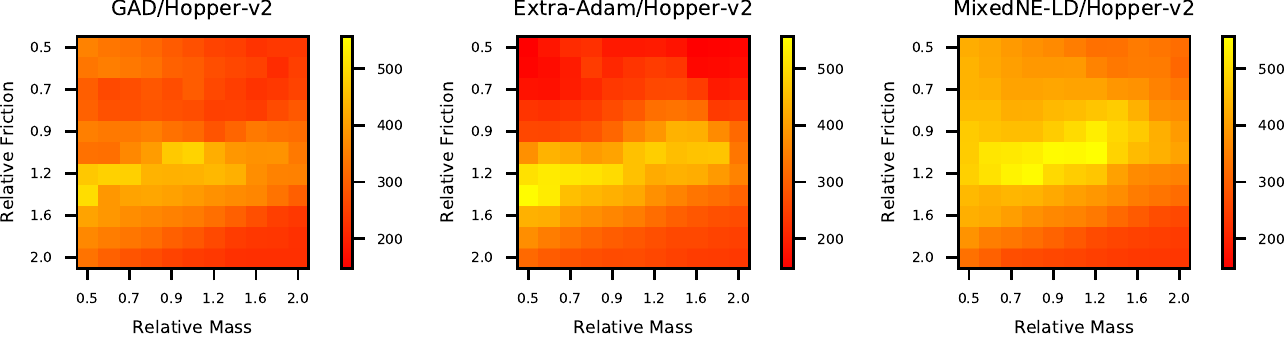}
		\label{fig:Heat_map_Hopper-1_both}
	\end{subfigure}%
    \\
	\begin{subfigure}[t]{1\textwidth}
		\centering
		\includegraphics[width=\linewidth]{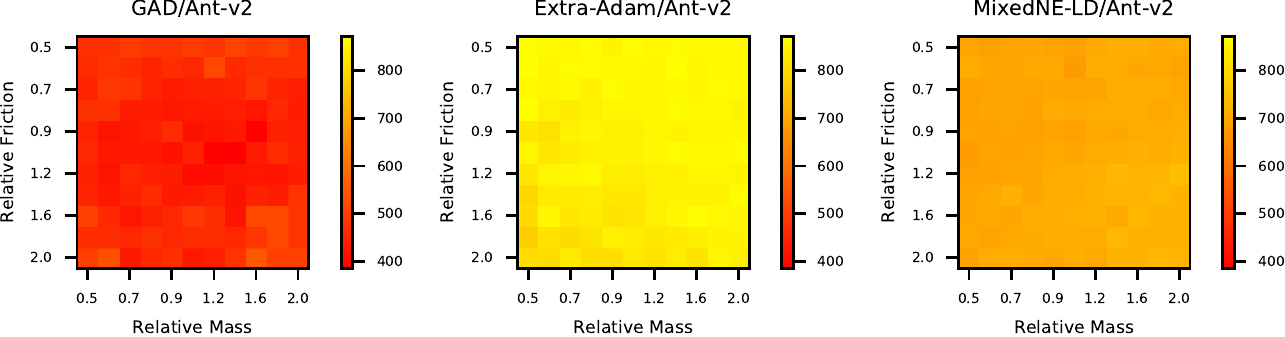}
		\label{fig:Heat_map_Ant-1_both}
	\end{subfigure}

	\caption{Average performance (over 5 seeds) of Algorithm~\ref{alg:nar-sgld-ddpg}, and Algorithm~\ref{alg:nar-ddpg} (with GAD and Extra-Adam), under the NR-MDP setting with $\delta = 0$. The evaluation is performed on a range of friction and mass values not encountered during training. Environments: Walker, HalfCheetah, Hopper, and Ant.}
	\label{fig:OnePlayer_Heat_map_both_comparison_average_a}
\end{figure*}

\begin{figure*}[t!]
	\centering
	    	\begin{subfigure}[t]{1\textwidth}
		\centering
		\includegraphics[width=\linewidth]{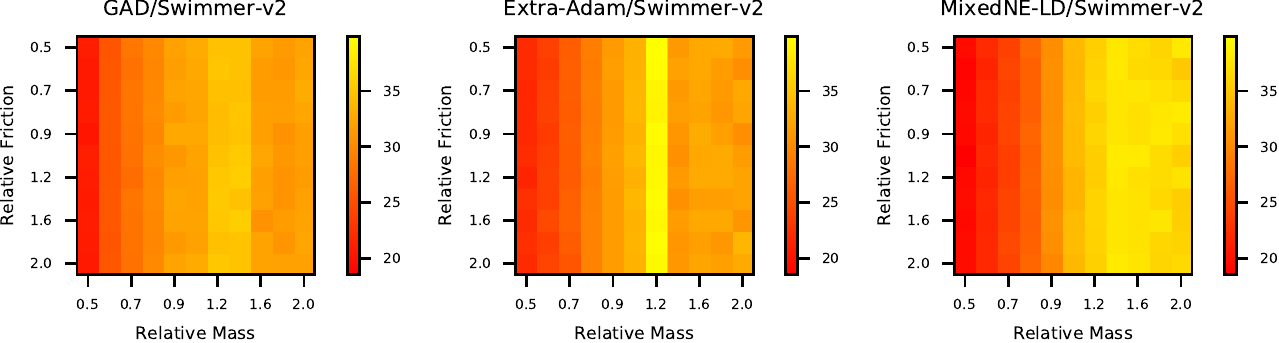}
		\label{fig:Heat_map_Swimmer-1_both}
	\end{subfigure}%
    \\
	\begin{subfigure}[t]{1\textwidth}
		\centering
		\includegraphics[width=\linewidth]{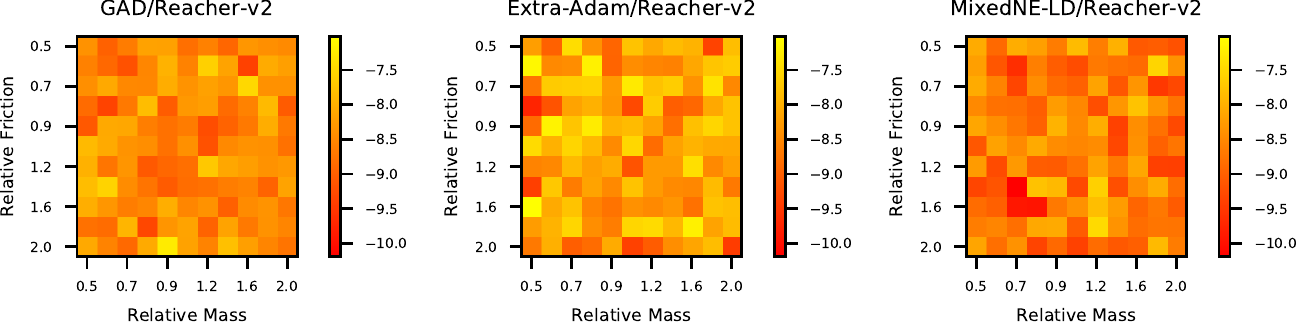}
		\label{fig:Heat_map_Reacher-1_both}
	\end{subfigure}%
    \\
    	\begin{subfigure}[t]{1\textwidth}
		\centering
		\includegraphics[width=\linewidth]{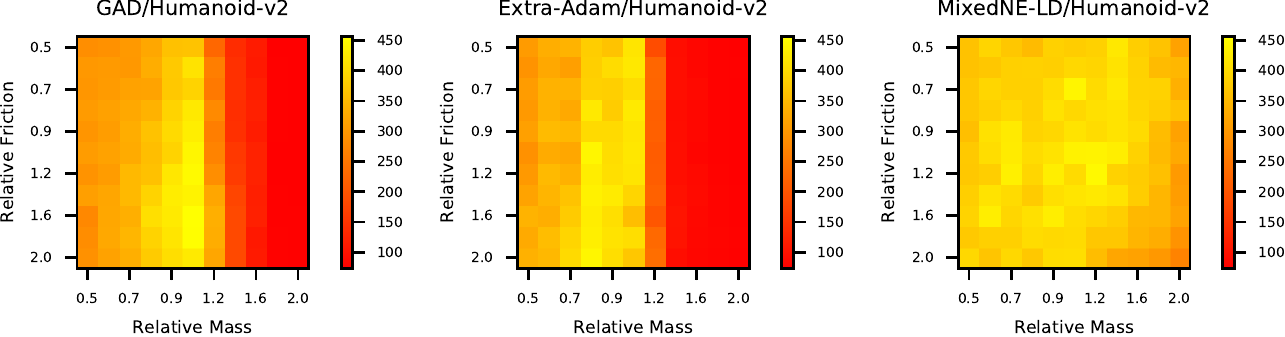}
		\label{fig:Heat_map_Humanoid-1_both}
	\end{subfigure}%
    \\
	\begin{subfigure}[t]{1\textwidth}
		\centering
		\includegraphics[width=\linewidth]{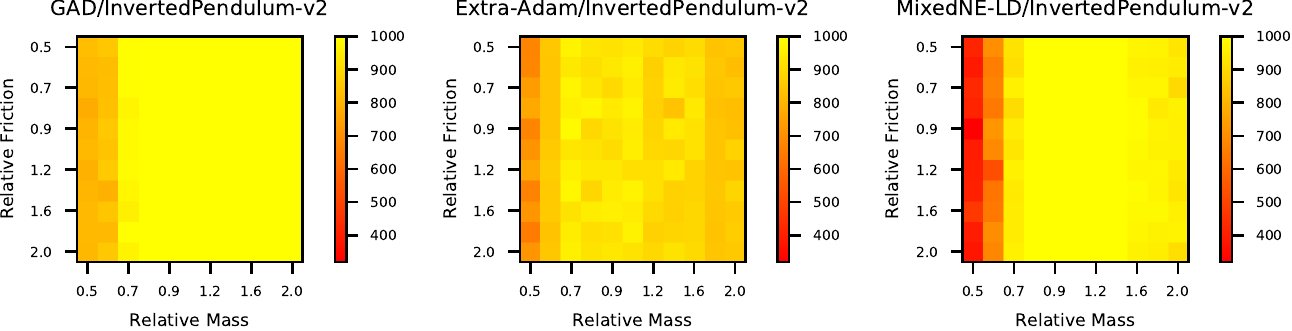}
		\label{fig:Heat_map_InvertedPendulum-1_both}
	\end{subfigure}

	\caption{Average performance (over 5 seeds) of Algorithm~\ref{alg:nar-sgld-ddpg}, and Algorithm~\ref{alg:nar-ddpg} (with GAD and Extra-Adam), under the NR-MDP setting with $\delta = 0$. The evaluation is performed on a range of friction and mass values not encountered during training. Environments: Swimmer, Reacher, Humanoid, and InvertedPendulum.}
	\label{fig:OnePlayer_Heat_map_both_comparison_average_b}
\end{figure*}

\begin{itemize}
\item Algorithms:  
\begin{enumerate}
\item MixedNE-LD: Algorithm~\ref{alg:nar-sgld-ddpg}
\item Baselines: Algorithm~\ref{alg:nar-ddpg} (with GAD and Extra-Adam)
\end{enumerate}
\item Hyperparameters: 
\begin{enumerate}
\item Common hyperparameters for Algorithm~\ref{alg:nar-sgld-ddpg} and Algorithm~\ref{alg:nar-ddpg}: Table~\ref{table:nonlin-fixed}
\item Exploration-related hyperparameters for Algorithm~\ref{alg:nar-sgld-ddpg} and Algorithm~\ref{alg:nar-ddpg} (the best performing values for every environment are presented): Tables~\ref{table:special-two}~and~\ref{table:special-one} 
\end{enumerate}
\item Results: 
\begin{enumerate}
\item Heat maps (mass-noise) for NR-MDP setting with $\delta = 0.1$ (Figures~\ref{fig:TwoPlayer_Heat_map_comparison_average_a}~and~\ref{fig:TwoPlayer_Heat_map_comparison_average_b})
\item Heat maps (mass-noise) for NR-MDP setting with $\delta = 0$ (Figures~\ref{fig:OnePlayer_Heat_map_comparison_average_a}~and~\ref{fig:OnePlayer_Heat_map_comparison_average_b})
\item Heat maps (friction-noise) for NR-MDP setting with $\delta = 0.1$ (Figures~\ref{fig:TwoPlayer_Heat_map_friction_comparison_average_a}~and~\ref{fig:TwoPlayer_Heat_map_friction_comparison_average_b})
\item Heat maps (friction-noise) for NR-MDP setting with $\delta = 0$ (Figures~\ref{fig:OnePlayer_Heat_map_friction_comparison_average_a}~and~\ref{fig:OnePlayer_Heat_map_friction_comparison_average_b})
\item Heat maps (mass-friction) for NR-MDP setting with $\delta = 0.1$ (Figures~\ref{fig:TwoPlayer_Heat_map_both_comparison_average_a}~and~\ref{fig:TwoPlayer_Heat_map_both_comparison_average_b})
\item Heat maps (mass-friction) for NR-MDP setting with $\delta = 0$ (Figures~\ref{fig:OnePlayer_Heat_map_both_comparison_average_a}~and~\ref{fig:OnePlayer_Heat_map_both_comparison_average_b})
\end{enumerate}
\end{itemize}


\newpage


\section{TD3 Experiments: Algorithms, Hyperparameters, and Results}
\label{sec:td3-exp-app}

\begin{figure*}[ht!]
\centering
\includegraphics[width=1\linewidth,height=1\linewidth]{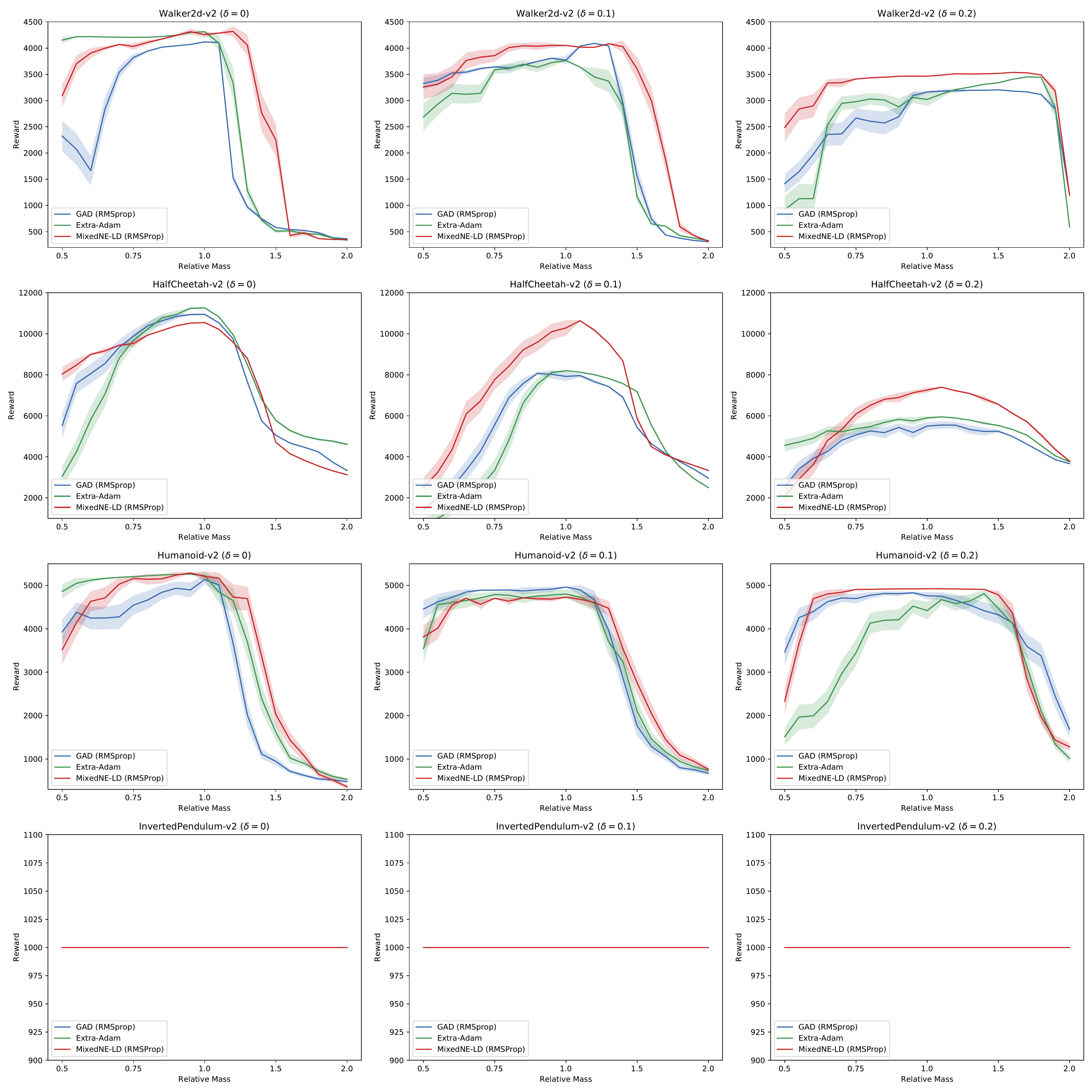}
\caption{Average performance (over 5 seeds) of Algorithm~\ref{alg:nar-sgld-td3} (TD3 with MixedNE-LD), and Algorithm~\ref{alg:nar-td3} (TD3 with GAD and Extra-Adam), under the NR-MDP setting with $\delta = 0,0.1,0.2$. The evaluation is performed without adversarial perturbations, on a range of mass values not encountered during training.}
\label{fig:td3_mass-2}
\end{figure*}

\begin{figure*}[ht!]
\centering
\includegraphics[width=1\linewidth,height=1\linewidth]{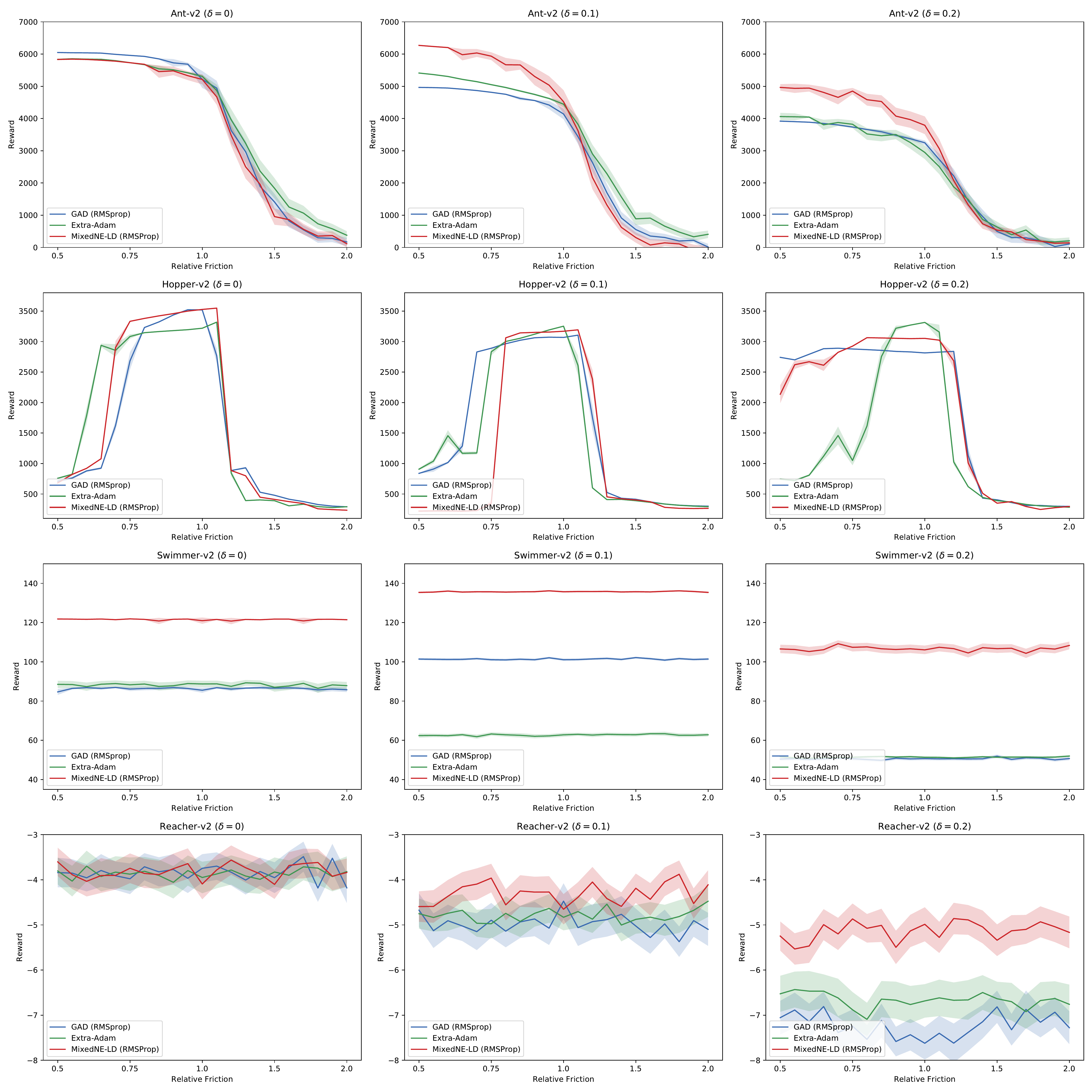}
\caption{Average performance (over 5 seeds) of Algorithm~\ref{alg:nar-sgld-td3} (TD3 with MixedNE-LD), and Algorithm~\ref{alg:nar-td3} (TD3 with GAD and Extra-Adam), under the NR-MDP setting with $\delta = 0,0.1,0.2$. The evaluation is performed without adversarial perturbations, on a range of friction values not encountered during training.}
\label{fig:td3_friction}
\end{figure*}

\begin{figure*}[ht!]
\centering
\includegraphics[width=1\linewidth,height=1\linewidth]{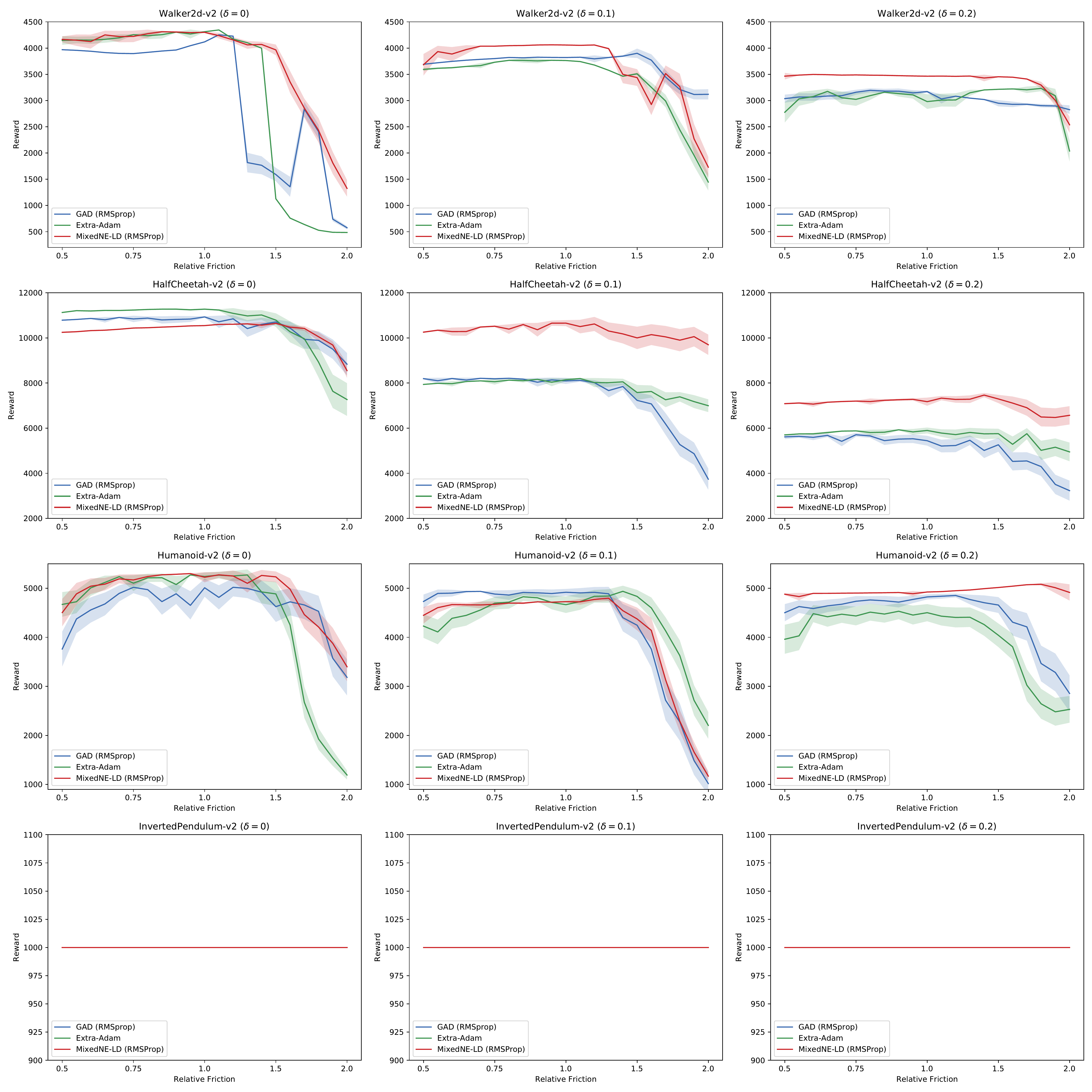}
\caption{Average performance (over 5 seeds) of Algorithm~\ref{alg:nar-sgld-td3} (TD3 with MixedNE-LD), and Algorithm~\ref{alg:nar-td3} (TD3 with GAD and Extra-Adam), under the NR-MDP setting with $\delta = 0,0.1,0.2$. The evaluation is performed without adversarial perturbations, on a range of friction values not encountered during training.}
\label{fig:td3_friction-2}
\end{figure*}

\begin{figure*}[ht!]
\centering
\includegraphics[width=1\linewidth,height=1.4\linewidth]{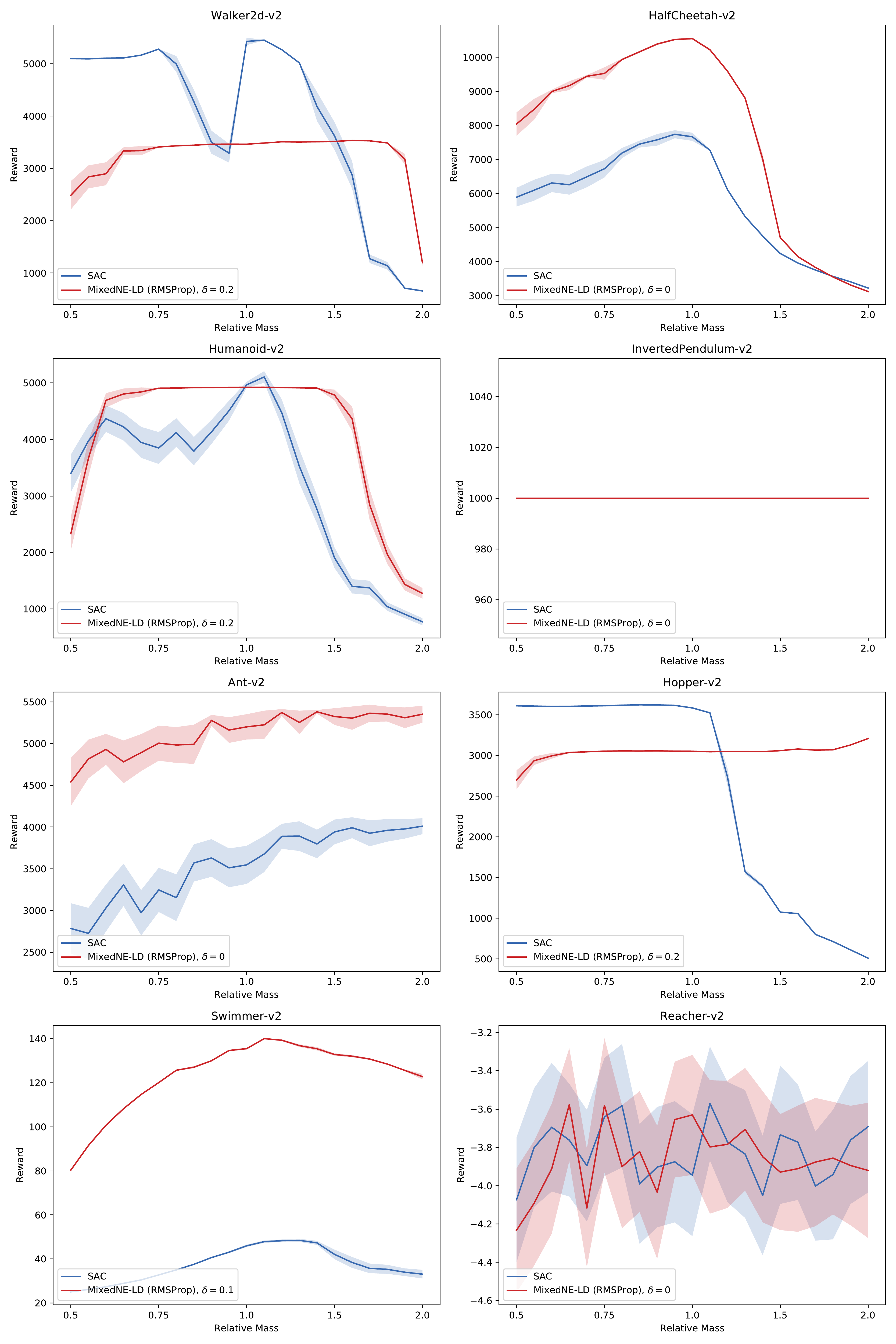}
\caption{Average performance (over 5 seeds) of Algorithm~\ref{alg:nar-sgld-td3} (TD3 with MixedNE-LD), and SAC, under the NR-MDP setting. The evaluation is performed without adversarial perturbations, on a range of mass values not encountered during training.}
\label{fig:td3_sac_mass}
\end{figure*}

\begin{figure*}[ht!]
\centering
\includegraphics[width=1\linewidth,height=1.4\linewidth]{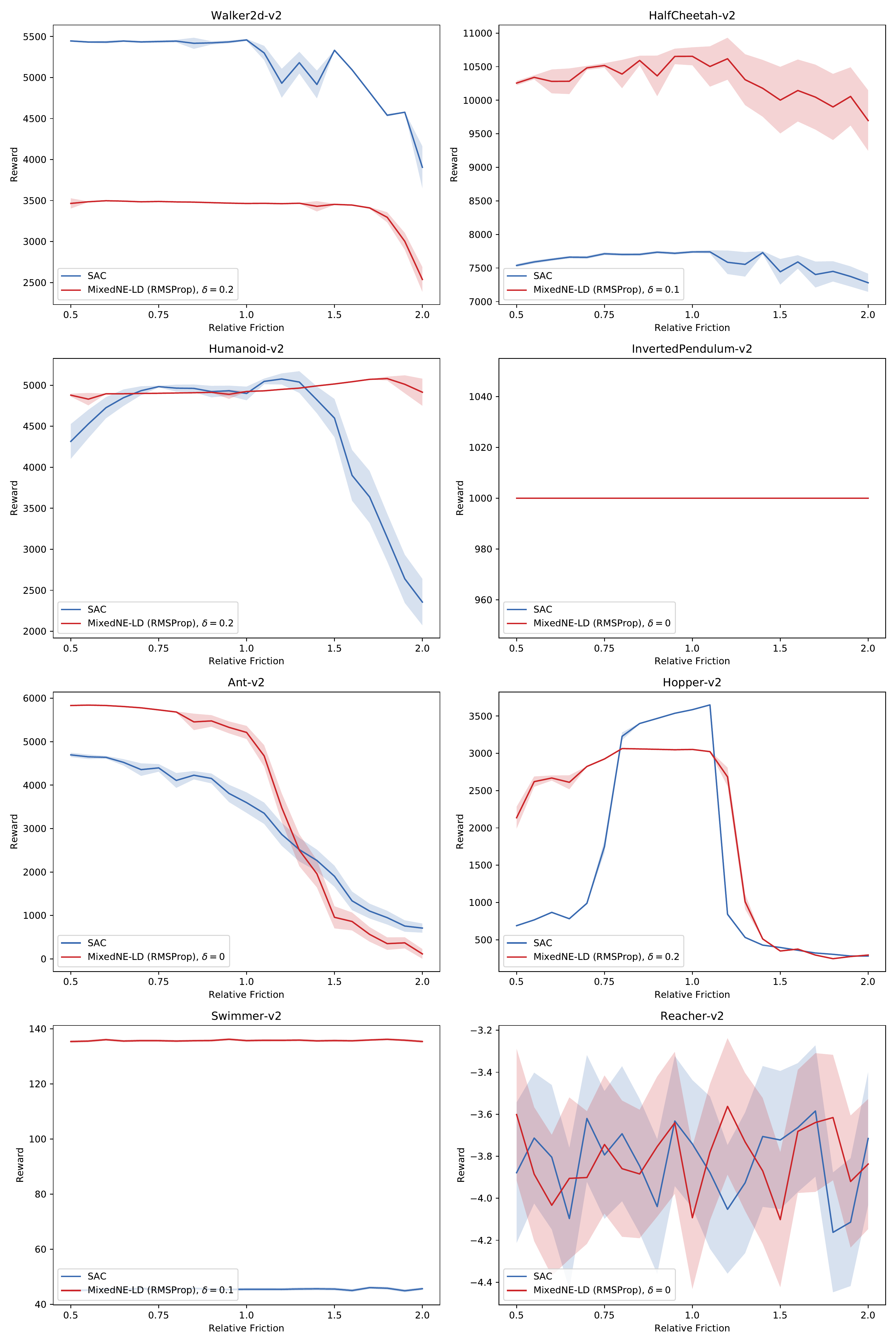}
\caption{Average performance (over 5 seeds) of Algorithm~\ref{alg:nar-sgld-td3} (TD3 with MixedNE-LD), and SAC, under the NR-MDP setting. The evaluation is performed without adversarial perturbations, on a range of friction values not encountered during training.}
\label{fig:td3_sac_friction}
\end{figure*}

\begin{itemize}
\item Algorithms:  
\begin{enumerate}
\item MixedNE-LD: Algorithm~\ref{alg:nar-sgld-td3}
\item Baselines: Algorithm~\ref{alg:nar-td3} (with GAD and Extra-Adam)
\end{enumerate}
\item Hyperparameters: 
\begin{enumerate}
\item Common hyperparameters for Algorithm~\ref{alg:nar-sgld-td3} and Algorithm~\ref{alg:nar-td3}: Table~\ref{table:nonlin-fixed-td3}
\item Exploration-related hyperparameters for Algorithm~\ref{alg:nar-sgld-td3} and Algorithm~\ref{alg:nar-td3} (the best performing values for every environment are presented): Tables~\ref{table:special-two-td3-1}, ~\ref{table:special-two-td3-2} and~\ref{table:special-one-td3} 
\end{enumerate}
\item Results: 
\begin{enumerate}
\item Mass uncertainty: Figures~\ref{fig:td3_mass}~and~\ref{fig:td3_mass-2}
\item Friction uncertainty: Figures~\ref{fig:td3_friction}~and~\ref{fig:td3_friction-2}
\item Comparison with SAC: Figures~\ref{fig:td3_sac_mass}~and~\ref{fig:td3_sac_friction}
\end{enumerate}
\end{itemize}


\newpage

\section{VPG Experiments: Setup and Evaluation}
\label{sec:vpg-exp-main}
In addition to the DDPG (off-policy) experiments, we test the effectiveness of the MixedNE-LD strategy with the vanilla policy gradient (VPG) method on a toy MDP problem. In particular, we design a two-player variant of VPG~\cite{sutton2000policy} algorithm (\emph{cf.} Algorithm~\ref{alg:vpg-two-sgld}) by adapting the Algorithm~\ref{algo:approx-infinite-md}.
\paragraph{Setup.} 
We compare the performance of Algorithm~\ref{alg:vpg-two-sgld} and Algorithm~\ref{alg:vpg-two-baseline} (with GAD and Extra-Adam) on a parametrized class of MDPs $\bc{\mathcal{M}_\rho = \br{\mathcal{S},\mathcal{A},T_\rho,\gamma,P_0,R} : \rho \in \bs{0 , 0.4}}$. Here $\mathcal{S} = \bs{-10 , 10}$, $\mathcal{A} = \bs{-1 , 1}$, and $R(s) = \sin (\sqrt{1.7} s) + \cos (\sqrt{0.3} s) + 3$. The transition dynamics $T_\rho$ is defined as follows: given the current state and action $\br{s_t , a_t}$, the next state is $s_{t+1} = s_t + a_t$ with probability $1-\rho$, and $s_{t+1} = s_t + a'$ (where $a' \sim \mathrm{unif}([-1,1])$) with probability $\rho$. We also ensure that $s_{t+1} \in [-10,10]$.

For all the algorithms, we use a two-layer feedforward neural network structure of (16, 16, relu) for both actors (agent and adversary). The relevant hyperparameters are given in Tables~\ref{table:nonlin-fixed-vpg},~\ref{table:special-three},~and~\ref{table:special-four}. Each algorithm is trained for 5000 steps. We run our experiments with 5 different seeds.
\paragraph{Evaluation.} 
We train the algorithms with a nominal environment parameter $\rho = 0.2$, and evaluate the learned policies on a range of $\rho \in \bs{0 , 0.4}$ values. As shown in Figure~\ref{fig:vpg-one-two-player_mass_comparison_average} (\emph{cf.}  Appendix~\ref{sec:vpg-exp-app}), our Algorithm~\ref{alg:vpg-two-sgld} outperforms the baselines Algorithm~\ref{alg:vpg-two-baseline} (with GAD and Extra-Adam) in terms of robustness (in both two-player and one-player settings).

\section{VPG Experiments: Algorithms, and Hyperparameters, and Results}
\label{sec:vpg-exp-app}

\begin{figure*}[t!]
	\centering
	\begin{subfigure}[t]{0.5\textwidth}
		\centering
		\includegraphics[width=1\linewidth]{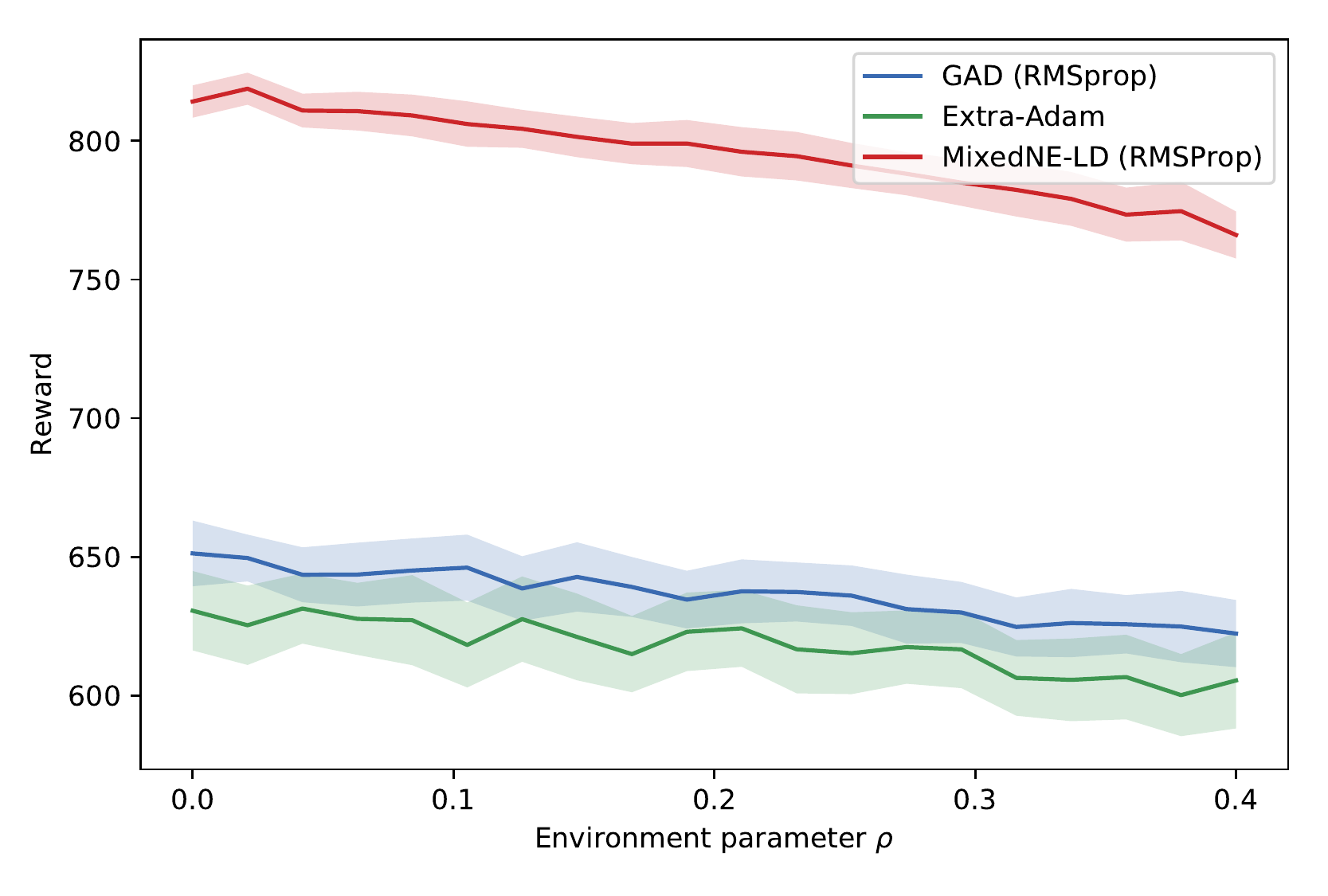}
		\caption{$\delta = 0.1$}
		\label{fig:vpg-twoplayer_mass_comparison_average}
	\end{subfigure}%
    ~
	\begin{subfigure}[t]{0.5\textwidth}
		\centering
		\includegraphics[width=\linewidth]{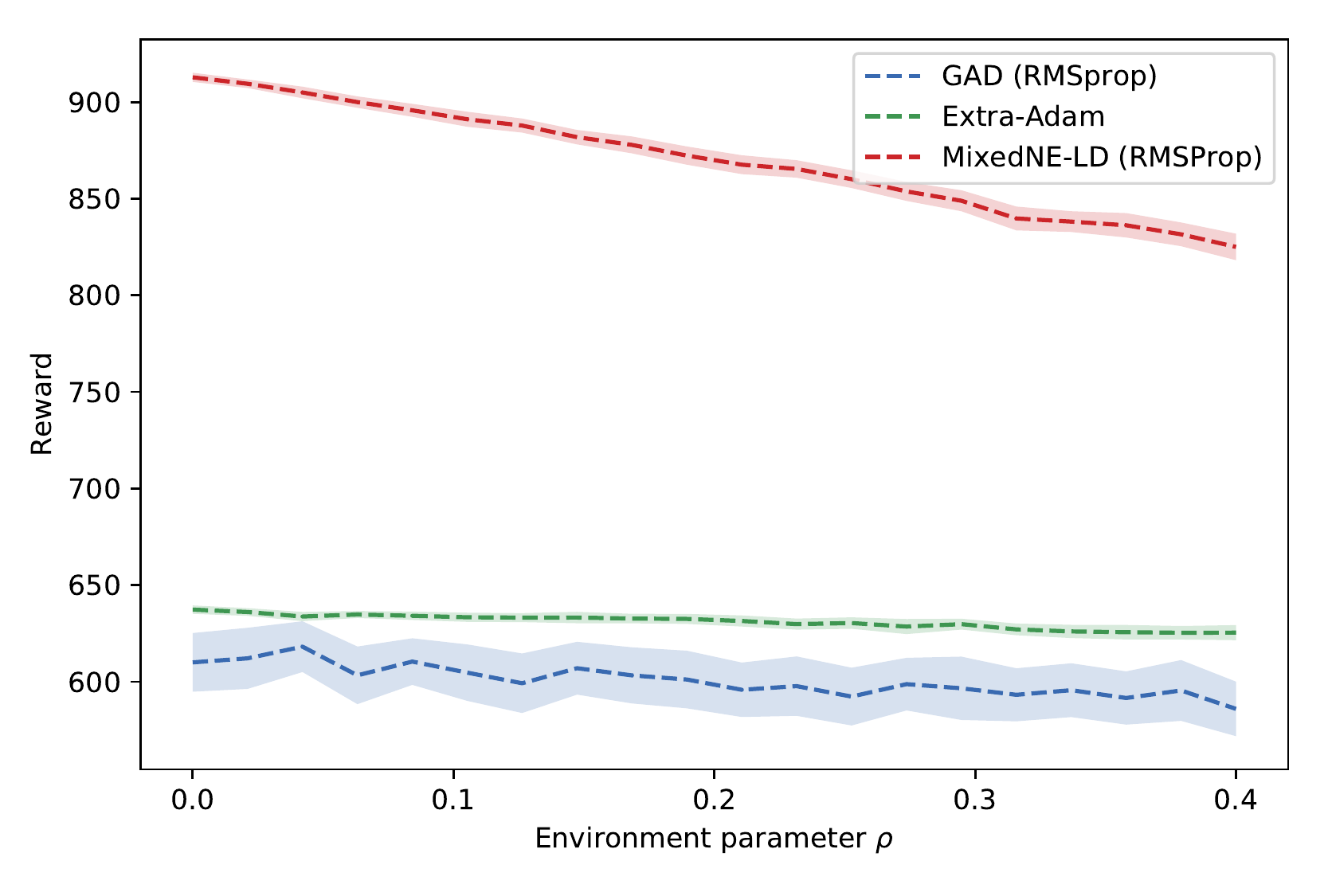}
		\caption{$\delta = 0$}
		\label{fig:vpg-oneplayer_mass_comparison_average}
	\end{subfigure}

	\caption{Average performance (over 5 seeds) of Algorithm~\ref{alg:vpg-two-sgld}, and Algorithm~\ref{alg:vpg-two-baseline} (with GAD and Extra-Adam), under the NR-MDP setting with $\delta = 0.1 \text{ and } 0$ (training on nominal environment $\rho_0 = 0.2$). The evaluation is performed without adversarial perturbations, on a range of environment parameters not encountered during training.}
	\label{fig:vpg-one-two-player_mass_comparison_average}
\end{figure*}

\begin{itemize}
\item Algorithms:  
\begin{enumerate}
\item MixedNE-LD: Algorithm~\ref{alg:vpg-two-sgld}
\item Baselines: Algorithm~\ref{alg:vpg-two-baseline} (with GAD and Extra-Adam)
\end{enumerate}
\item Hyperparameters: 
\begin{enumerate}
\item Common hyperparameters for Algorithm~\ref{alg:vpg-two-sgld} and Algorithm~\ref{alg:vpg-two-baseline}: Table~\ref{table:nonlin-fixed-vpg}
\item Additional hyperparameters for Algorithm~\ref{alg:vpg-two-sgld} and Algorithm~\ref{alg:vpg-two-baseline} (the best performing values are presented): Tables~\ref{table:special-three}~and~\ref{table:special-four} 
\end{enumerate}
\item Results: 
\begin{enumerate}
\item NR-MDP setting with $\delta = 0.1$ (Figure~\ref{fig:vpg-twoplayer_mass_comparison_average})
\item NR-MDP setting with $\delta = 0$ (Figure~\ref{fig:vpg-oneplayer_mass_comparison_average})
\end{enumerate}
\end{itemize}


%


\newpage

\begin{algorithm}[tb]
	\caption{DDPG with MixedNE-LD (pre-conditioner = RMSProp)}
	\label{alg:nar-sgld-ddpg}
	\begin{algorithmic}
	        \STATE \textbf{Hyperparameters:} see Table~\ref{table:nonlin-fixed}
		\STATE Initialize (randomly) policy parameters $\omega_1 , \theta_1$, and Q-function parameter $\phi$. 
		\STATE Initialize the target network parameters $\omega_\mathrm{targ} \gets \omega_1$, $\theta_\mathrm{targ} \gets \theta_1$, and $\phi_\mathrm{targ} \gets \phi$.
		\STATE Initialize replay buffer $\mathcal{D}$.
		\STATE Initialize $m \gets \mathbf{0} \,\, ; \,\, m' \gets \mathbf{0}$.
		\STATE $t \gets 1$.
		\REPEAT
			\STATE Observe state $s$, and select actions $a = \mu_{\theta_t}(s) + \xi$ ; $a' = \nu_{\omega_t}(s) + \xi'$, where $\xi , \xi' \sim \mathcal{N}\br{0,\sigma I}$
			\STATE Execute the action $\bar a = (1-\delta) a + \delta a'$ in the environment.
			\STATE Observe reward $r$, next state $s'$, and done signal $d$ to indicate whether $s'$ is terminal.
			\STATE Store $\br{s, \bar a, r, s',d}$ in replay buffer $\mathcal{D}$.
			\STATE If $s'$ is terminal, reset the environment state. 
			\IF{it's time to update}
				\FOR{however many updates}
					\STATE $\bar \omega_t , \omega_t^{\br{1}} \gets \omega_t \,\, ; \,\, \bar \theta_t , \theta_t^{\br{1}} \gets \theta_t$
		        \FOR{$k=1,2,\dots,K_t$}
					\STATE Sample a random minibatch of $N$ transitions $B=\bc{\br{s, \bar a, r, s',d}}$ from $\mathcal{D}$.
					\STATE Compute targets $y\br{r,s',d} = r + \gamma \br{1-d} Q_{\phi_\mathrm{targ}} \br{s' , (1-\delta) \mu_{\theta_\mathrm{targ}}\br{s'} + \delta \nu_{\omega_\mathrm{targ}}\br{s'}}$. 
					\STATE Update critic by one step of (preconditioned) gradient descent using $\nabla_\phi L\br{\phi}$, where  
					\[
						L\br{\phi} ~=~ \frac{1}{N} \sum_{\br{s, \bar a, r, s',d} \in B} \br{y\br{r,s',d} - Q_\phi\br{s, \bar a}}^2 .
					\]
					\STATE Compute the (agent and adversary) policy gradient estimates:
					\begin{align*}
					\widehat{\nabla_\theta J \br{\theta , \omega_t}} ~=~&\frac{1-\delta}{N} \sum_{s \in \mathcal{D}}{\nabla_\theta \mu_\theta \br{s} \nabla_{\bar a} Q_\phi\br{s, \bar a} \vert_{\bar a = (1-\delta) \mu_\theta \br{s} + \delta \nu_{\omega_t} \br{s}}} \\
					\widehat{\nabla_{\omega} J \br{\theta_t , \omega}} ~=~& \frac{\delta}{N} \sum_{s \in \mathcal{D}}{\nabla_{\omega} \nu_{\omega} \br{s} \nabla_{\bar a} Q_\phi\br{s, \bar a} \vert_{\bar a = (1-\delta) \mu_{\theta_t} \br{s} + \delta \nu_{\omega} \br{s}}} .
					\end{align*}
		        		\STATE $g \gets \bs{\widehat{\nabla_\theta J \br{\theta, \omega_t}}}_{\theta = \theta_t^{\br{k}}} \,\, ; \,\, m \gets \alpha m + \br{1 - \alpha} g \odot g \,\, ; \,\, C \gets \mathrm{diag}\br{\sqrt{m + \epsilon}}$
		        		\STATE $\theta_{t}^{\br{k+1}} \gets \theta_{t}^{\br{k}} + \eta C^{-1} g + \sqrt{2 \eta} \sigma_t C^{-\frac{1}{2}} \xi$, where $\xi \sim \mathcal{N}\br{0,I}$
				\STATE $g' \gets \bs{\widehat{\nabla_\omega J \br{\theta_t, \omega}}}_{\omega = \omega_t^{\br{k}}} \,\, ; \,\, m' \gets \alpha m' + \br{1 - \alpha} g' \odot g' \,\, ; \,\, D \gets \mathrm{diag}\br{\sqrt{m' + \epsilon}}$
				\STATE $\omega_{t}^{\br{k+1}} \gets \omega_{t}^{\br{k}} - \eta D^{-1} g' + \sqrt{2 \eta} \sigma_t D^{-\frac{1}{2}} \xi'$, where $\xi' \sim \mathcal{N}\br{0,I}$
				\STATE $\bar \omega_t \gets \br{1 - \beta} \bar \omega_t + \beta \omega_{t}^{\br{k+1}} \,\, ; \,\, \bar \theta_t \gets \br{1 - \beta} \bar \theta_t + \beta \theta_{t}^{\br{k+1}}$
				\STATE Update the target networks:
		\begin{align*}
		\phi_\mathrm{targ} ~\gets~& \tau \phi_\mathrm{targ} + (1-\tau) \phi \\
		\theta_\mathrm{targ} ~\gets~& \tau \theta_\mathrm{targ} + (1 - \tau) \theta_{t}^{\br{k+1}} \\
		\omega_\mathrm{targ} ~\gets~& \tau \omega_\mathrm{targ} + (1 - \tau) \omega_{t}^{\br{k+1}}
		\end{align*}
			\ENDFOR 
			\STATE $\omega_{t+1} \gets \br{1 - \beta} \omega_t + \beta \bar \omega_t \,\, ; \,\, \theta_{t+1} \gets \br{1 - \beta} \theta_t + \beta \bar \theta_t$
					\STATE $t \gets t+1$.
				\ENDFOR
			\ENDIF
		\UNTIL{convergence}
		\STATE \textbf{Output:} $\omega_T$, $\theta_T$.
	\end{algorithmic}
\end{algorithm}

\begin{algorithm}[tb]
	\caption{DDPG with GAD (pre-conditioner = RMSProp) / Extra-Adam}
	\label{alg:nar-ddpg}
	\begin{algorithmic}
	        \STATE \textbf{Hyperparameters:} see Table~\ref{table:nonlin-fixed}
		\STATE Initialize (randomly) policy parameters $\omega_1 , \theta_1$, and Q-function parameter $\phi$. 
		\STATE Initialize the target network parameters $\omega_\mathrm{targ} \gets \omega_1$, $\theta_\mathrm{targ} \gets \theta_1$, and $\phi_\mathrm{targ} \gets \phi$.
		\STATE Initialize replay buffer $\mathcal{D}$.
		\STATE Initialize $m \gets \mathbf{0} \,\, ; \,\, m' \gets \mathbf{0}$.
		\STATE $t \gets 1$.
		\REPEAT
			\STATE Observe state $s$, and select actions $a = \mu_{\theta_t}(s) + \xi$ ; $a' = \nu_{\omega_t}(s) + \xi'$, where $\xi , \xi' \sim \mathcal{N}\br{0,\sigma I}$
			\STATE Execute the action $\bar a = (1-\delta) a + \delta a'$ in the environment.
			\STATE Observe reward $r$, next state $s'$, and done signal $d$ to indicate whether $s'$ is terminal.
			\STATE Store $\br{s, \bar a, r, s',d}$ in replay buffer $\mathcal{D}$.
			\STATE If $s'$ is terminal, reset the environment state. 
			\IF{it's time to update} 
				\FOR{however many updates}
					\STATE Sample a random minibatch of $N$ transitions $B=\bc{\br{s, \bar a, r, s',d}}$ from $\mathcal{D}$.
					\STATE Compute targets $y\br{r,s',d} = r + \gamma \br{1-d} Q_{\phi_\mathrm{targ}} \br{s' , (1-\delta) \mu_{\theta_\mathrm{targ}}\br{s'} + \delta \nu_{\omega_\mathrm{targ}}\br{s'}}$. 
					\STATE Update critic by one step of (preconditioned) gradient descent using $\nabla_\phi L\br{\phi}$, where  
					\[
						L\br{\phi} ~=~ \frac{1}{N} \sum_{\br{s, \bar a, r, s',d} \in B} \br{y\br{r,s',d} - Q_\phi\br{s, \bar a}}^2 .
					\]
					\STATE Compute the (agent and adversary) policy gradient estimates:
					\begin{align*}
					\widehat{\nabla_\theta J \br{\theta , \omega_t}} ~=~& \frac{1-\delta}{N} \sum_{s \in \mathcal{D}}{\nabla_\theta \mu_\theta \br{s} \nabla_{\bar a} Q_\phi\br{s, \bar a} \vert_{\bar a = (1-\delta) \mu_\theta \br{s} + \delta \nu_{\omega_t} \br{s}}} \\
					\widehat{\nabla_{\omega} J \br{\theta_t , \omega}} ~=~& \frac{\delta}{N} \sum_{s \in \mathcal{D}}{\nabla_{\omega} \nu_{\omega} \br{s} \nabla_{\bar a} Q_\phi\br{s, \bar a} \vert_{\bar a = (1-\delta) \mu_{\theta_t} \br{s} + \delta \nu_{\omega} \br{s}}} .
					\end{align*}
					\STATE \textbf{GAD (pre-conditioner = RMSProp):}
		       \STATE $g \gets \bs{\widehat{\nabla_\theta J \br{\theta, \omega_t}}}_{\theta = \theta_t} \,\, ; \,\, m \gets \alpha m + \br{1 - \alpha} g \odot g \,\, ; \,\, C \gets \mathrm{diag}\br{\sqrt{m + \epsilon}}$
		        		\STATE $\theta_{t+1} \gets \theta_{t} + \eta C^{-1} g$
				\STATE $g' \gets \bs{\widehat{\nabla_\omega J \br{\theta_t, \omega}}}_{\omega = \omega_t} \,\, ; \,\, m' \gets \alpha m' + \br{1 - \alpha} g' \odot g' \,\, ; \,\, D \gets \mathrm{diag}\br{\sqrt{m' + \epsilon}}$
				\STATE $\omega_{t+1} \gets \omega_{t} - \eta D^{-1} g'$
				\STATE \textbf{Extra-Adam:} use Algorithm~4 from \cite{gidel2018variational}.
				\STATE Update the target networks:
		\begin{align*}
		\phi_\mathrm{targ} ~\gets~& \tau \phi_\mathrm{targ} + (1-\tau) \phi \\
		\theta_\mathrm{targ} ~\gets~& \tau \theta_\mathrm{targ} + (1 - \tau) \theta_{t+1} \\
		\omega_\mathrm{targ} ~\gets~& \tau \omega_\mathrm{targ} + (1 - \tau) \omega_{t+1}
		\end{align*}
					\STATE $t \gets t+1$.
				\ENDFOR
			\ENDIF
		\UNTIL{convergence}
		\STATE \textbf{Output:} $\omega_T$, $\theta_T$.
	\end{algorithmic}
\end{algorithm}



\begin{algorithm}[tb]
	\caption{TD3 with MixedNE-LD (pre-conditioner = RMSProp)}
	\label{alg:nar-sgld-td3}
	\begin{algorithmic}
	      \STATE \textbf{Hyperparameters:} see Table~\ref{table:nonlin-fixed-td3}
		\STATE Initialize (randomly) policy parameters $\omega_1 , \theta_1$, and Q-function parameters $\phi_1$, $\phi_2$. 
		\STATE Initialize the target network parameters $\omega_\mathrm{targ} \gets \omega_1$, $\theta_\mathrm{targ} \gets \theta_1$, and $\phi_\mathrm{targ, 1} \gets \phi_1$, $\phi_\mathrm{targ, 2} \gets \phi_2$.
		\STATE Initialize replay buffer $\mathcal{D}$.
		\STATE Initialize $m \gets \mathbf{0} \,\, ; \,\, m' \gets \mathbf{0}$.
		\STATE $t \gets 1$.
		\REPEAT
			\STATE Observe state $s$, and select actions $a =$ clip $(\mu_{\theta_t}(s) + \xi, a_{\mathrm{Low}}, a_{\mathrm{High}})$ ; $a' =$ clip$( \nu_{\omega_t}(s) + \xi', a_{\mathrm{Low}}, a_{\mathrm{High}})$, where $\xi , \xi' \sim \mathcal{N}\br{0,\sigma I}$
			\STATE Execute the action $\bar a = (1-\delta) a + \delta a'$ in the environment.
			\STATE Observe reward $r$, next state $s'$, and done signal $d$ to indicate whether $s'$ is terminal.
			\STATE Store $\br{s, \bar a, r, s',d}$ in replay buffer $\mathcal{D}$.
			\STATE If $s'$ is terminal, reset the environment state. 
			\IF{it's time to update}
				\FOR{however many updates}
					\STATE $\bar \omega_t , \omega_t^{\br{1}} \gets \omega_t \,\, ; \,\, \bar \theta_t , \theta_t^{\br{1}} \gets \theta_t$
		        \FOR{$k=1,2,\dots,K_t$}
					\STATE Sample a random minibatch of $N$ transitions $B=\bc{\br{s, \bar a, r, s',d}}$ from $\mathcal{D}$.
					\STATE Compute target actions
					\[
					    \tilde{a} = \text{clip}((1-\delta) \mu_{\theta_\mathrm{targ}}\br{s'} + \delta \nu_{\omega_\mathrm{targ}}\br{s'} + \epsilon, a_{\mathrm{Low}}, a_{\mathrm{High}}), \text{where} \hspace{0.2cm} \epsilon \sim \text{clip}(\mathcal{N}\br{0,\sigma I}, -c, c)
					\]
					\STATE Compute targets $y\br{r,s',d} = r + \gamma \min_{i=1, 2} \br{1-d} Q_{\phi_\mathrm{targ, i}} \br{s' , \tilde{a}}$. 
					\STATE Update critic by one step of (preconditioned) gradient descent using $\nabla_\phi L\br{\phi}$, where  
					\[
						L\br{\phi} ~=~ \frac{1}{N} \sum_{\br{s, \bar a, r, s',d} \in B} \br{y\br{r,s',d} - Q_{\phi_{i}}\br{s, \bar a}}^2. \hspace{1cm} \text{for $i = 1, 2$}
					\]
					\IF{$t$ mod policy\textunderscore delay = 0}
					\STATE Compute the (agent and adversary) policy gradient estimates:
					\begin{align*}
					\widehat{\nabla_\theta J \br{\theta , \omega_t}} ~=~&\frac{1-\delta}{N} \sum_{s \in \mathcal{D}}{\nabla_\theta \mu_\theta \br{s} \nabla_{\bar a} Q_{\phi_{1}}\br{s, \bar a} \vert_{\bar a = (1-\delta) \mu_\theta \br{s} + \delta \nu_{\omega_t} \br{s}}} \\
					\widehat{\nabla_{\omega} J \br{\theta_t , \omega}} ~=~& \frac{\delta}{N} \sum_{s \in \mathcal{D}}{\nabla_{\omega} \nu_{\omega} \br{s} \nabla_{\bar a} Q_{\phi_{1}}\br{s, \bar a} \vert_{\bar a = (1-\delta) \mu_{\theta_t} \br{s} + \delta \nu_{\omega} \br{s}}} .
					\end{align*}
		        		\STATE $g \gets \bs{\widehat{\nabla_\theta J \br{\theta, \omega_t}}}_{\theta = \theta_t^{\br{k}}} \,\, ; \,\, m \gets \alpha m + \br{1 - \alpha} g \odot g \,\, ; \,\, C \gets \mathrm{diag}\br{\sqrt{m + \epsilon}}$
		        		\STATE $\theta_{t}^{\br{k+1}} \gets \theta_{t}^{\br{k}} + \eta C^{-1} g + \sqrt{2 \eta} \sigma_t C^{-\frac{1}{2}} \xi$, where $\xi \sim \mathcal{N}\br{0,I}$
				\STATE $g' \gets \bs{\widehat{\nabla_\omega J \br{\theta_t, \omega}}}_{\omega = \omega_t^{\br{k}}} \,\, ; \,\, m' \gets \alpha m' + \br{1 - \alpha} g' \odot g' \,\, ; \,\, D \gets \mathrm{diag}\br{\sqrt{m' + \epsilon}}$
				\STATE $\omega_{t}^{\br{k+1}} \gets \omega_{t}^{\br{k}} - \eta D^{-1} g' + \sqrt{2 \eta} \sigma_t D^{-\frac{1}{2}} \xi'$, where $\xi' \sim \mathcal{N}\br{0,I}$
				\STATE $\bar \omega_t \gets \br{1 - \beta} \bar \omega_t + \beta \omega_{t}^{\br{k+1}} \,\, ; \,\, \bar \theta_t \gets \br{1 - \beta} \bar \theta_t + \beta \theta_{t}^{\br{k+1}}$
				\STATE Update the target networks:
		\begin{align*}
		\phi_\mathrm{targ, i} ~\gets~& \tau \phi_\mathrm{targ, i} + (1-\tau) \phi_{i} \hspace{1cm} \text{for $i = 1, 2$} \\
		\theta_\mathrm{targ} ~\gets~& \tau \theta_\mathrm{targ} + (1 - \tau) \theta_{t}^{\br{k+1}} \\
		\omega_\mathrm{targ} ~\gets~& \tau \omega_\mathrm{targ} + (1 - \tau) \omega_{t}^{\br{k+1}}
		\end{align*}					    
					\ENDIF
					
			\ENDFOR 
			\STATE $\omega_{t+1} \gets \br{1 - \beta} \omega_t + \beta \bar \omega_t \,\, ; \,\, \theta_{t+1} \gets \br{1 - \beta} \theta_t + \beta \bar \theta_t$
					\STATE $t \gets t+1$.
				\ENDFOR
			\ENDIF
		\UNTIL{convergence}
		\STATE \textbf{Output:} $\omega_T$, $\theta_T$.
	\end{algorithmic}
\end{algorithm}



\begin{algorithm}[tb]
	\caption{TD3 with GAD (pre-conditioner = RMSProp) / Extra-Adam}
	\label{alg:nar-td3}
	\begin{algorithmic}
	      \STATE \textbf{Hyperparameters:} see Table~\ref{table:nonlin-fixed-td3}
		\STATE Initialize (randomly) policy parameters $\omega_1 , \theta_1$, and Q-function parameters $\phi_1$, $\phi_2$. 
		\STATE Initialize the target network parameters $\omega_\mathrm{targ} \gets \omega_1$, $\theta_\mathrm{targ} \gets \theta_1$, and $\phi_\mathrm{targ, 1} \gets \phi_1$, $\phi_\mathrm{targ, 2} \gets \phi_2$.
		\STATE Initialize replay buffer $\mathcal{D}$.
		\STATE Initialize $m \gets \mathbf{0} \,\, ; \,\, m' \gets \mathbf{0}$.
		\STATE $t \gets 1$.
		\REPEAT
			\STATE Observe state $s$, and select actions $a =$ clip $(\mu_{\theta_t}(s) + \xi, a_{\mathrm{Low}}, a_{\mathrm{High}})$ ; $a' =$ clip$( \nu_{\omega_t}(s) + \xi', a_{\mathrm{Low}}, a_{\mathrm{High}})$, where $\xi , \xi' \sim \mathcal{N}\br{0,\sigma I}$
			\STATE Execute the action $\bar a = (1-\delta) a + \delta a'$ in the environment.
			\STATE Observe reward $r$, next state $s'$, and done signal $d$ to indicate whether $s'$ is terminal.
			\STATE Store $\br{s, \bar a, r, s',d}$ in replay buffer $\mathcal{D}$.
			\STATE If $s'$ is terminal, reset the environment state. 
			\IF{it's time to update}
				\FOR{however many updates}

					\STATE Sample a random minibatch of $N$ transitions $B=\bc{\br{s, \bar a, r, s',d}}$ from $\mathcal{D}$.
					\STATE Compute target actions
					\[
					    \tilde{a} = \text{clip}((1-\delta) \mu_{\theta_\mathrm{targ}}\br{s'} + \delta \nu_{\omega_\mathrm{targ}}\br{s'} + \epsilon, a_{\mathrm{Low}}, a_{\mathrm{High}}), \text{where} \hspace{0.2cm} \epsilon \sim \text{clip}(\mathcal{N}\br{0,\sigma I}, -c, c)
					\]
					\STATE Compute targets $y\br{r,s',d} = r + \gamma \min_{i=1, 2} \br{1-d} Q_{\phi_\mathrm{targ, i}} \br{s' , \tilde{a}}$. 
					\STATE Update critic by one step of (preconditioned) gradient descent using $\nabla_\phi L\br{\phi}$, where  
					\[
						L\br{\phi} ~=~ \frac{1}{N} \sum_{\br{s, \bar a, r, s',d} \in B} \br{y\br{r,s',d} - Q_{\phi_{i}}\br{s, \bar a}}^2. \hspace{1cm} \text{for $i = 1, 2$}
					\]
					\IF{$t$ mod policy\textunderscore delay = 0}
					\STATE Compute the (agent and adversary) policy gradient estimates:
					\begin{align*}
					\widehat{\nabla_\theta J \br{\theta , \omega_t}} ~=~&\frac{1-\delta}{N} \sum_{s \in \mathcal{D}}{\nabla_\theta \mu_\theta \br{s} \nabla_{\bar a} Q_{\phi_{1}}\br{s, \bar a} \vert_{\bar a = (1-\delta) \mu_\theta \br{s} + \delta \nu_{\omega_t} \br{s}}} \\
					\widehat{\nabla_{\omega} J \br{\theta_t , \omega}} ~=~& \frac{\delta}{N} \sum_{s \in \mathcal{D}}{\nabla_{\omega} \nu_{\omega} \br{s} \nabla_{\bar a} Q_{\phi_{1}}\br{s, \bar a} \vert_{\bar a = (1-\delta) \mu_{\theta_t} \br{s} + \delta \nu_{\omega} \br{s}}} .
					\end{align*}
					\STATE \textbf{GAD (pre-conditioner = RMSProp):}
		       \STATE $g \gets \bs{\widehat{\nabla_\theta J \br{\theta, \omega_t}}}_{\theta = \theta_t} \,\, ; \,\, m \gets \alpha m + \br{1 - \alpha} g \odot g \,\, ; \,\, C \gets \mathrm{diag}\br{\sqrt{m + \epsilon}}$
		        		\STATE $\theta_{t+1} \gets \theta_{t} + \eta C^{-1} g$
				\STATE $g' \gets \bs{\widehat{\nabla_\omega J \br{\theta_t, \omega}}}_{\omega = \omega_t} \,\, ; \,\, m' \gets \alpha m' + \br{1 - \alpha} g' \odot g' \,\, ; \,\, D \gets \mathrm{diag}\br{\sqrt{m' + \epsilon}}$
				\STATE $\omega_{t+1} \gets \omega_{t} - \eta D^{-1} g'$
				\STATE \textbf{Extra-Adam:} use Algorithm~4 from \cite{gidel2018variational}.
				\STATE Update the target networks:
		\begin{align*}
		\phi_\mathrm{targ, i} ~\gets~& \tau \phi_\mathrm{targ, i} + (1-\tau) \phi_{i} \hspace{1cm} \text{for $i = 1, 2$} \\
		\theta_\mathrm{targ} ~\gets~& \tau \theta_\mathrm{targ} + (1 - \tau) \theta_{t+1} \\
		\omega_\mathrm{targ} ~\gets~& \tau \omega_\mathrm{targ} + (1 - \tau) \omega_{t+1}
		\end{align*}					    
					\ENDIF
					
					\STATE $t \gets t+1$.
				\ENDFOR
			\ENDIF
		\UNTIL{convergence}
		\STATE \textbf{Output:} $\omega_T$, $\theta_T$.
	\end{algorithmic}
\end{algorithm}


\begin{algorithm}[tb]
	\caption{VPG with MixedNE-LD (pre-conditioner = RMSProp)}
	\label{alg:vpg-two-sgld}
	\begin{algorithmic}
	        \STATE \textbf{Hyperparameters:} see Table~\ref{table:nonlin-fixed-vpg}
		\STATE Initialize (randomly) policy parameters $\theta_0$, $w_0$
		\FOR{$k=0,1,2,\dots$}
			\STATE $\bar \theta_k , \theta_k^{\br{0}} \gets \theta_k$ ; $\bar w_k , w_k^{\br{0}} \gets w_k$
			\FOR{$n=0,1,\dots,N_k$}
					\STATE Collect set of trajectories $\mathcal{D}_k^{(n)} = \{ (\dots, s_t^{(\tau)}, \bar a_t^{(\tau)}, r_t^{(\tau)}, \dots) \}_\tau$ by running $\pi_{\theta_k^{(n)}}$, and $\pi'_{w_k^{(n)}}$ in $\mathcal{M}$, i.e., $a_t \sim \pi_{\theta_k^{(n)}} (s_t)$, $a'_t \sim \pi'_{w_k^{(n)}} (s_t)$, $\bar a_t = (1 - \delta) a_t + \delta a'_t$, and $s_{t+1} \sim T_\rho (\cdot \mid s_t , \bar a_t)$.
					\STATE Estimate the policy gradient (where $G_t = \sum_{s=0}^T{\gamma^s r_{t+s}}$)
					\begin{align*}
					g ~=~& \frac{1 - \delta}{\abs{\mathcal{D}_k^{(n)}}} \sum_{\tau \in \mathcal{D}_k^{(n)}} \sum_t \gamma^t G_t^{(\tau)} \left[ \nabla_\theta \log \pi_\theta (a_t^{(\tau)} \mid s_t^{(\tau)}) \right]_{\theta = \theta_k^{(n)}} \\
					g' ~=~& \frac{\delta}{\abs{\mathcal{D}_k^{(n)}}} \sum_{\tau \in \mathcal{D}_k^{(n)}} \sum_t \gamma^t G_t^{(\tau)} \left[ \nabla_w \log \pi_w ({a'_t}^{(\tau)} \mid s_t^{(\tau)}) \right]_{w = w_k^{(n)}}
					\end{align*}
		       \STATE $m \gets \alpha m + \br{1 - \alpha} g \odot g \,\, ; \,\, C \gets \mathrm{diag}\br{\sqrt{m + \epsilon}}$
		       \STATE $\theta_{k}^{(n+1)} \gets \theta_{k}^{(n)} + \eta C^{-1} g + \sqrt{2 \eta} \sigma_k C^{-\frac{1}{2}} \xi$, where $\xi \sim \mathcal{N}\br{0,I}$
		       \STATE $\bar \theta_k \gets \br{1 - \beta} \bar \theta_k + \beta \theta_{k}^{\br{n+1}}$
		       \STATE $m' \gets \alpha m' + \br{1 - \alpha} g' \odot g' \,\, ; \,\, D \gets \mathrm{diag}\br{\sqrt{m' + \epsilon}}$
		       \STATE $w_{k}^{(n+1)} \gets w_{k}^{(n)} - \eta D^{-1} g + \sqrt{2 \eta} \sigma_k D^{-\frac{1}{2}} \xi'$, where $\xi' \sim \mathcal{N}\br{0,I}$
		       \STATE $\bar w_k \gets \br{1 - \beta} \bar w_k + \beta w_{k}^{\br{n+1}}$
			\ENDFOR
			\STATE $\theta_{k+1} \gets \br{1 - \beta} \theta_k + \beta \bar \theta_k$
			\STATE $w_{k+1} \gets \br{1 - \beta} w_k + \beta \bar w_k$
		\ENDFOR
	\end{algorithmic}
\end{algorithm}

\begin{algorithm}[tb]
	\caption{VPG with GAD (pre-conditioner = RMSProp) / Extra-Adam}
	\label{alg:vpg-two-baseline}
	\begin{algorithmic}
	        \STATE \textbf{Hyperparameters:} see Table~\ref{table:nonlin-fixed-vpg}
		\STATE Initialize (randomly) policy parameters $\theta_0$, $w_0$
			\FOR{$k=0,1,2,\dots$}
					\STATE Collect set of trajectories $\mathcal{D}_k = \{ (\dots, s_t^{(\tau)}, \bar a_t^{(\tau)}, r_t^{(\tau)}, \dots) \}_\tau$ by running $\pi_{\theta_k}$, and $\pi'_{w_k}$ in $\mathcal{M}$, i.e., $a_t \sim \pi_{\theta_k} (s_t)$, $a'_t \sim \pi'_{w_k} (s_t)$, $\bar a_t = (1 - \delta) a_t + \delta a'_t$, and $s_{t+1} \sim T_\rho (\cdot \mid s_t , \bar a_t)$.
					\STATE Estimate the policy gradient (where $G_t = \sum_{s=0}^T{\gamma^s r_{t+s}}$)
					\begin{align*}
					g ~=~& \frac{1-\delta}{\abs{\mathcal{D}_k}} \sum_{\tau \in \mathcal{D}_k} \sum_t \gamma^t G_t^{(\tau)} \left[ \nabla_\theta \log \pi_\theta (a_t^{(\tau)} \mid s_t^{(\tau)}) \right]_{\theta = \theta_k} \\
					g' ~=~& \frac{\delta}{\abs{\mathcal{D}_k}} \sum_{\tau \in \mathcal{D}_k} \sum_t \gamma^t G_t^{(\tau)} \left[ \nabla_w \log \pi'_w ({a'_t}^{(\tau)} \mid s_t^{(\tau)}) \right]_{w = w_k}
					\end{align*}
		       \STATE \textbf{GAD (pre-conditioner = RMSProp):}
		       \STATE $m \gets \alpha m + \br{1 - \alpha} g \odot g \,\, ; \,\, C \gets \mathrm{diag}\br{\sqrt{m + \epsilon}}$
		       \STATE $\theta_{k+1} \gets \theta_{k} + \eta C^{-1} g$
		        \STATE $m' \gets \alpha m' + \br{1 - \alpha} g' \odot g' \,\, ; \,\, D \gets \mathrm{diag}\br{\sqrt{m' + \epsilon}}$
		       \STATE $w_{k+1} \gets w_{k} - \eta D^{-1} g'$
		       \STATE \textbf{Extra-Adam:} use Algorithm~4 from \cite{gidel2018variational}.
			\ENDFOR
	\end{algorithmic}
\end{algorithm}


\newpage

\begin{table}[ht]
\caption{Common hyperparameters for Algorithm~\ref{alg:nar-sgld-ddpg} and Algorithm~\ref{alg:nar-ddpg}, where most of the values are chosen from \cite{baselines}.}
\centering 
\begin{tabular}{l l} 
\hline\hline \\ [-2ex]
Hyperparameter & Value \\ [0.5ex] 
\hline \\ [-1ex]
critic optimizer & Adam \\
critic learning rate & $10^{-3}$ \\
target update rate $\tau$ & $0.999$ \\
mini-batch size $N$ & $128$ \\
discount factor $\gamma$ & $0.99$ \\ 
damping factor $\beta$ & $0.9$ \\ 
replay buffer size & $10^6$ \\ 
action noise parameter $\sigma$ & $\bc{0,0.01,0.1,0.2,0.3,0.4}$ \\
RMSProp parameter $\alpha$ & $0.999$ \\
RMSProp parameter $\epsilon$ & $10^{-8}$ \\
RMSProp parameter $\eta$ & $10^{-4}$ \\
thermal noise $\sigma_t$ (Algorithm~\ref{alg:nar-sgld-ddpg}) & $\sigma_0 \times (1-5 \times 10^{-5})^t$, where $\sigma_0 \in \bc{10^{-2}, 10^{-3}, 10^{-4}, 10^{-5}}$ \\
warmup steps $K_t$ (Algorithm~\ref{alg:nar-sgld-ddpg}) & $\min \bc{ 15 , \floor{(1+10^{-5})^t}}$ \\ [1ex]
\hline
\end{tabular}
\label{table:nonlin-fixed} 
\end{table}

\begin{table}[ht]
\caption{Exploration-related hyperparameters for Algorithm~\ref{alg:nar-sgld-ddpg} and Algorithm~\ref{alg:nar-ddpg} chosen via grid search (for NR-MDP setting with $\delta = 0.1$).} 
\centering 
\begin{tabular}{l l l l} 
\hline\hline \\ [-2ex]
 & Algorithm~\ref{alg:nar-sgld-ddpg}: $( \sigma_0 , \sigma )$ & Algorithm~\ref{alg:nar-ddpg} (with GAD): $\sigma$ & Algorithm~\ref{alg:nar-ddpg} (with Extra-Adam): $\sigma$ \\ [0.5ex] 
\hline \\ [-1ex]
Walker-v2 & $(10^{-2}, 0.01)$ & $0$ & $0.3$ \\
HalfCheetah-v2 & $(10^{-2}, 0)$ & $0.2$ & $0.01$ \\
Hopper-v2 & $(10^{-3}, 0.2)$ & $0.2$ & $0.3$ \\
Ant-v2 & $(10^{-4}, 0.2)$ & $0.4$ & $0.01$ \\ 
Swimmer-v2 & $(10^{-5}, 0.4)$ & $0.4$ & $0.4$ \\
Reacher-v2 & $(10^{-3}, 0.2)$ & $0.4$ & $0.2$ \\
Humanoid-v2 & $(10^{-4}, 0.01)$ & $0$ & $0.01$ \\
InvertedPendulum-v2 & $(10^{-3}, 0.01)$ & $0.1$ & $0.01$ \\ [1ex]
\hline
\end{tabular}
\label{table:special-two} 
\end{table}

\begin{table}[ht]
\caption{Exploration-related hyperparameters for Algorithm~\ref{alg:nar-sgld-ddpg} and Algorithm~\ref{alg:nar-ddpg} chosen via grid search (for NR-MDP setting with $\delta = 0$).} 
\centering 
\begin{tabular}{l l l l} 
\hline\hline \\ [-2ex]
 & Algorithm~\ref{alg:nar-sgld-ddpg}: $( \sigma_0 , \sigma )$ & Algorithm~\ref{alg:nar-ddpg} (with GAD): $\sigma$ & Algorithm~\ref{alg:nar-ddpg} (with Extra-Adam): $\sigma$ \\ [0.5ex] 
\hline \\ [-1ex]
Walker-v2 & $(10^{-2}, 0.1)$ & $0.01$ & $0.2$ \\
HalfCheetah-v2 & $(10^{-2}, 0.01)$ & $0.4$ & $0.01$ \\
Hopper-v2 & $(10^{-5}, 0.3)$ & $0.4$ & $0.1$ \\
Ant-v2 & $(10^{-2}, 0.4)$ & $0.4$ & $0.01$ \\ 
Swimmer-v2 & $(10^{-2}, 0.2)$ & $0.3$ & $0.3$ \\
Reacher-v2 & $(10^{-3}, 0.2)$ & $0.3$ & $0.2$ \\
Humanoid-v2 & $(10^{-2}, 0.1)$ & $0$ & $0.01$ \\
InvertedPendulum-v2 & $(10^{-3}, 0)$ & $0.01$ & $0.01$ \\ [1ex]
\hline
\end{tabular}
\label{table:special-one} 
\end{table}



\begin{table}[ht]
\caption{Common hyperparameters for Algorithm~\ref{alg:nar-sgld-td3} and Algorithm~\ref{alg:nar-td3}, where most of the values are chosen from \cite{fujimoto2018addressing}.}
\centering 
\begin{tabular}{l l} 
\hline\hline \\ [-2ex]
Hyperparameter & Value \\ [0.5ex] 
\hline \\ [-1ex]
critic optimizer & Adam \\
critic learning rate & $3 \times 10^{-4}$ \\
target update rate $\tau$ & $0.995$ \\
mini-batch size $N$ & $128$ \\
discount factor $\gamma$ & $0.99$ \\ 
damping factor $\beta$ & $0.9$ \\ 
replay buffer size & $10^6$ \\ 
action noise parameter $\sigma$ & $\bc{0.005, 0.01, 0.1}$ \\
RMSProp parameter $\alpha$ & $0.999$ \\
RMSProp parameter $\epsilon$ & $10^{-8}$ \\
RMSProp parameter $\eta$ & $10^{-4}$ \\
thermal noise $\sigma_t$ (Algorithm~\ref{alg:nar-sgld-ddpg}) & $\sigma_0 \times (1-5 \times 10^{-5})^t$, where $\sigma_0 \in \bc{10^{-2}, 10^{-3}, 10^{-4}, 10^{-5}}$ \\
warmup steps $K_t$ (Algorithm~\ref{alg:nar-sgld-ddpg}) & $\min \bc{ 15 , \floor{(1+10^{-5})^t}}$ \\ [1ex]
\hline
\end{tabular}
\label{table:nonlin-fixed-td3} 
\end{table}

\begin{table}[ht]
\caption{Exploration-related hyperparameters for Algorithm~\ref{alg:nar-sgld-td3} and Algorithm~\ref{alg:nar-td3} chosen via grid search (for NR-MDP setting with $\delta = 0.2$).} 
\centering 
\begin{tabular}{l l l l} 
\hline\hline \\ [-2ex]
 & Algorithm~\ref{alg:nar-sgld-td3}: $( \sigma_0 , \sigma )$ & Algorithm~\ref{alg:nar-td3} (with GAD): $\sigma$ & Algorithm~\ref{alg:nar-td3} (with Extra-Adam): $\sigma$ \\ [0.5ex] 
\hline \\ [-1ex]
Walker-v2 & $(10^{-4}, 0.005)$ & $0.005$ & $0.005$ \\
HalfCheetah-v2 & $(10^{-5}, 0.005)$ & $0.005$ & $0.1$ \\
Hopper-v2  & $(10^{-4}, 0.1)$ & $0.01$ & $0.1$ \\
Ant-v2  & $(10^{-4}, 0.005)$ & $0.005$ & $0.1$ \\
Swimmer-v2 & $(10^{-4}, 0.005)$ & $0.01$ & $0.01$ \\
Reacher-v2  & $(10^{-2}, 0.1)$ & $0.005$ & $0.1$ \\
Humanoid-v2  & $(10^{-3}, 0.005)$ & $0.01$ & $0.01$ \\
InvertedPendulum-v2   & $(10^{-4}, 0.01)$ & $0.005$ & $0.01$ \\[1ex]
\hline
\end{tabular}
\label{table:special-two-td3-1} 
\end{table}

\begin{table}[ht]
\caption{Exploration-related hyperparameters for Algorithm~\ref{alg:nar-sgld-td3} and Algorithm~\ref{alg:nar-td3} chosen via grid search (for NR-MDP setting with $\delta = 0.1$).} 
\centering 
\begin{tabular}{l l l l} 
\hline\hline \\ [-2ex]
 & Algorithm~\ref{alg:nar-sgld-td3}: $( \sigma_0 , \sigma )$ & Algorithm~\ref{alg:nar-td3} (with GAD): $\sigma$ & Algorithm~\ref{alg:nar-td3} (with Extra-Adam): $\sigma$ \\ [0.5ex] 
\hline \\ [-1ex]
Walker-v2  & $(10^{-3}, 0.01)$ & $0.01$ & $0.01$  \\
HalfCheetah-v2  & $(10^{-5}, 0.1)$ & $0.01$ & $0.01$ \\
Hopper-v2  & $(10^{-4}, 0.01)$ & $0.01$ & $0.1$ \\
Ant-v2  & $(10^{-3}, 0.01)$ & $0.005$ & $0.005$ \\
Swimmer-v2 & $(10^{-4}, 0.005)$ & $0.1$ & $0.005$ \\
Reacher-v2  & $(10^{-4}, 0.005)$ & $0.005$ & $0.01$ \\
Humanoid-v2  & $(10^{-5}, 0.1)$ & $0.01$ & $0.01$ \\
InvertedPendulum-v2  & $(10^{-3}, 0.01)$ & $0.01$ & $0.01$ \\[1ex]
\hline
\end{tabular}
\label{table:special-two-td3-2} 
\end{table}

\begin{table}[ht]
\caption{Exploration-related hyperparameters for Algorithm~\ref{alg:nar-sgld-td3} and Algorithm~\ref{alg:nar-td3} chosen via grid search (for NR-MDP setting with $\delta = 0$).} 
\centering 
\begin{tabular}{l l l l} 
\hline\hline \\ [-2ex]
 & Algorithm~\ref{alg:nar-sgld-td3}: $( \sigma_0 , \sigma )$ & Algorithm~\ref{alg:nar-td3} (with GAD): $\sigma$ & Algorithm~\ref{alg:nar-td3} (with Extra-Adam): $\sigma$ \\ [0.5ex] 
\hline \\ [-1ex]
Walker-v2  & $(10^{-5}, 0.01)$ & $0.01$ & $0.1$  \\
HalfCheetah-v2 & $(10^{-5}, 0.01)$ & $0.01$ & $0.001$ \\
Hopper-v2  & $(10^{-4}, 0.1)$ & $0.1$ & $0.005$ \\
Ant-v2  & $(10^{-3}, 0.1)$ & $0.1$ & $0.1$ \\
Swimmer-v2 & $(10^{-5}, 0.01)$ & $0.01$ & $0.005$ \\
Reacher-v2  & $(10^{-4}, 0.1)$ & $0.1$ & $0.1$ \\
Humanoid-v2  & $(10^{-4}, 0.1)$ & $0.1$ & $0.005$ \\
InvertedPendulum-v2 & $(10^{-4}, 0.01)$ & $0.01$ & $0.005$ \\[1ex]
\hline
\end{tabular}
\label{table:special-one-td3} 
\end{table}

\begin{table}[ht]
\caption{Common hyperparameters for Algorithm~\ref{alg:vpg-two-sgld} and Algorithm~\ref{alg:vpg-two-baseline}.} 
\centering 
\begin{tabular}{l l} 
\hline\hline \\ [-2ex]
Hyperparameter & Value \\ [0.5ex] 
\hline \\ [-1ex]
discount factor $\gamma$ & $0.99$ \\ 
trajectory length $H$ & $500$ \\ 
number of trajectories per step $\abs{\mathcal{D}_k}$ & $1$ \\ 
RMSProp parameter $\alpha$ & $0.99$ \\
RMSProp parameter $\epsilon$ & $10^{-8}$ \\
learning rate $\eta$ & $\bc{10^{-3}, 10^{-4}, 10^{-5}}$ \\ 
damping factor $\beta$ & $0.9$ \\ [1ex]
\hline
\end{tabular}
\label{table:nonlin-fixed-vpg} 
\end{table}

\begin{table}[ht]
\caption{Additional hyperparameters for Algorithm~\ref{alg:vpg-two-sgld} and Algorithm~\ref{alg:vpg-two-baseline} chosen via grid search (for NR-MDP setting with $\delta = 0.1$)} 
\centering 
\begin{tabular}{l l l l} 
\hline\hline \\ [-2ex]
 & Algorithm~\ref{alg:vpg-two-sgld}: $( \sigma_0 , \eta , N_k)$ & Algorithm~\ref{alg:vpg-two-baseline} (with GAD): $\eta$ & Algorithm~\ref{alg:vpg-two-baseline} (with Extra-Adam): $\eta$ \\ [0.5ex] 
\hline \\ [-1ex]
$\rho = \mathrm{0.2}$ & $(10^{-5}, 10^{-3}, 1)$ & $10^{-4}$ & $10^{-4}$ \\ [1ex]
\hline
\end{tabular}
\label{table:special-three} 
\end{table}

\begin{table}[ht]
\caption{Additional hyperparameters for Algorithm~\ref{alg:vpg-two-sgld} and Algorithm~\ref{alg:vpg-two-baseline} chosen via grid search (for NR-MDP setting with $\delta = 0$)} 
\centering 
\begin{tabular}{l l l l} 
\hline\hline \\ [-2ex]
 & Algorithm~\ref{alg:vpg-two-sgld}: $( \sigma_0 , \eta, N_k)$ & Algorithm~\ref{alg:vpg-two-baseline} (with GAD): $\eta$ & Algorithm~\ref{alg:vpg-two-baseline} (with Extra-Adam): $\eta$ \\ [0.5ex] 
\hline \\ [-1ex]
$\rho = \mathrm{0.2}$ & $(10^{-4}, 10^{-4}, 10)$ & $10^{-4}$ & $10^{-3}$ \\ [1ex]
\hline
\end{tabular}
\label{table:special-four} 
\end{table}


\newpage

\begin{figure*}[ht!]
\centering
\includegraphics[width=1\linewidth,height=1.4\linewidth]{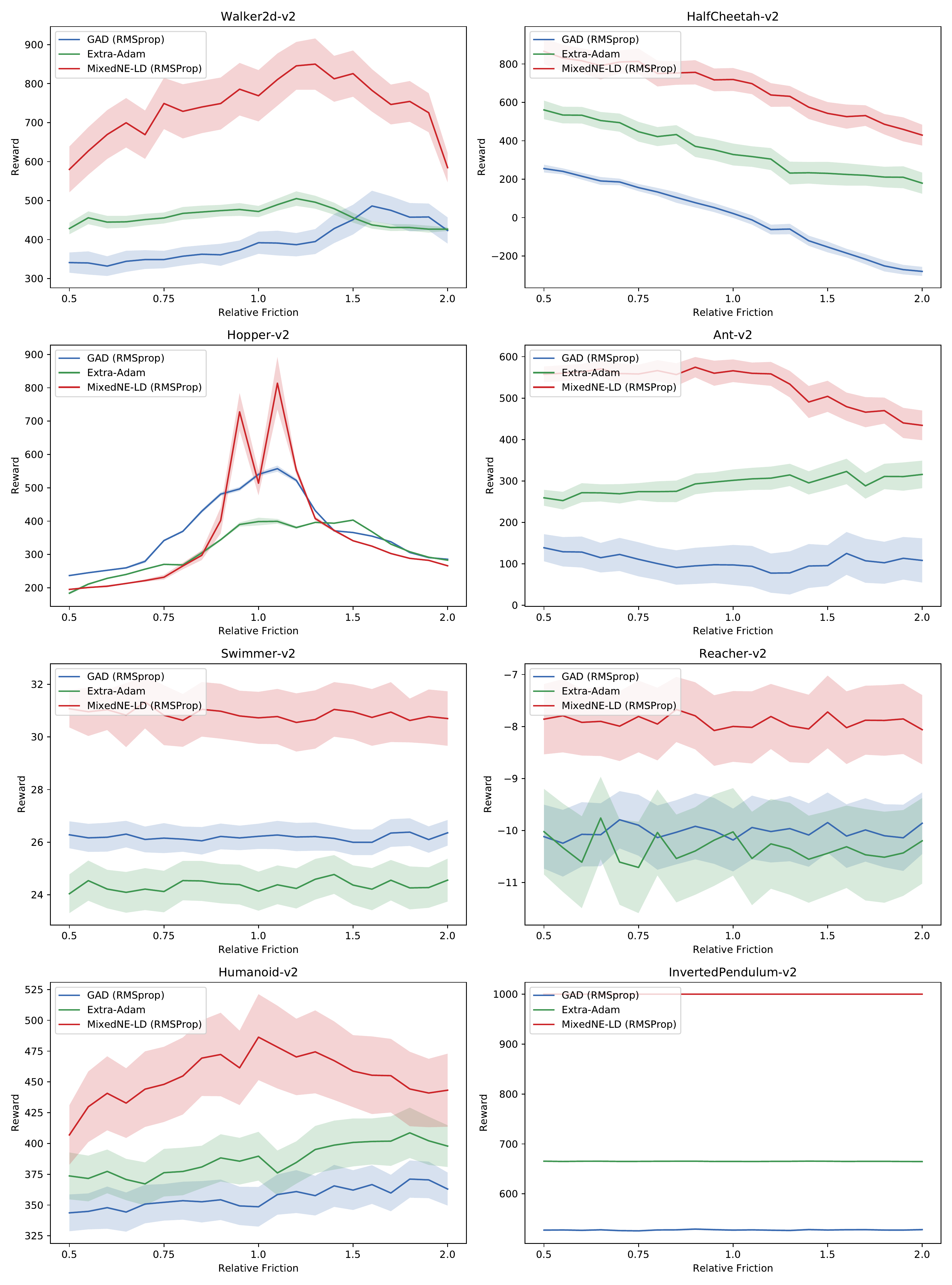}
\caption{Average performance (over 5 seeds) of Algorithm~\ref{alg:nar-sgld-ddpg} (DDPG with MixedNE-LD), and Algorithm~\ref{alg:nar-ddpg} (DDPG with GAD and Extra-Adam), under the NR-MDP setting with $\delta = 0.1$. The evaluation is performed without adversarial perturbations, on a range of friction values not encountered during training.}
\label{fig:Twoplayer_friction_comparison_average}
\end{figure*}

\newpage
 
%% file: 8_appendix-case-study.tex

\section{Algorithms and Omitted Proofs for Section~\ref{sec:case.study}}
\label{app:case-study}

%
\subsection{Algorithms and Hyperparameters}

The pseudocode of the algorithms can be found in \textbf{Algorithm~\ref{algo:cts-bandit}} (the symbol $\Pi$ denotes the projection). The hyperparameter setting for experiments in Section~\ref{sec:case.study} is:
\begin{itemize}
\item Algorithm~\ref{algo:cts-bandit} with GDA, and $\eta_t = 0.1$
\item Algorithm~\ref{algo:cts-bandit} with EG, and $\eta_t = 0.1$
\item Algorithm~\ref{algo:cts-bandit} with MixedNE-LD, $\eta_t = 0.1$, $\epsilon_t = 0.01$, $K_t = 50$, and $\beta = 0.5$.
\end{itemize}

We also note that we focus on the ``last iterate'' convergence for EG \cite{abernethy2019last, daskalakis2019last}, instead of the usual ergodic average in convex optimization literature. This is because, in practice, people almost exclusively use the last iterate.

\begin{algorithm}[tb]
	\caption{Algorithms in Section~\ref{sec:case.study} (MixedNE-LD / GAD / EG)}
	\label{algo:cts-bandit}
	\begin{algorithmic}
		\STATE \textbf{Input:} step-size $\bc{\eta_t}_{t=1}^T$, thermal-noise $\bc{\epsilon_t}_{t=1}^T$, warmup steps $\bc{K_t}_{t=1}^T$, exponential damping factor $\beta$.
		\FOR{$t=1,2,\dots,T-1$}
		\vspace{2mm}
			\STATE {\textbf{MixedNE-LD:}}
		        \STATE $\bar \omega_t , \omega_t^{\br{1}} \gets \omega_t \,\, ; \,\, \bar \theta_t , \theta_t^{\br{1}} \gets \theta_t$
		        \FOR{$k=1,2,\dots,K_t$}
		        		\STATE $\xi,\xi' \sim \mathcal{N}\br{0,I}$
		        		\STATE $\theta_t^{\br{k+1}} ~\gets~ \Pi_{\Theta} \br{\theta_t^{\br{k}} + \eta_t \nabla_\theta f(\theta_t^{\br{k}},\omega_t) + \epsilon_t \sqrt{2 \eta_t} \xi'}$
		        		\STATE $\omega_t^{\br{k+1}} ~\gets~ \Pi_{\Omega} \br{\omega_t^{\br{k}} - \eta_t \nabla_\omega f(\theta_t, \omega_t^{\br{k}}) + \epsilon_t \sqrt{2 \eta_t} \xi}$
				\STATE $\bar \omega_t \gets \br{1 - \beta} \bar \omega_t + \beta \omega_{t}^{\br{k+1}} $
				\STATE $\bar \theta_t \gets \br{1 - \beta} \bar \theta_t + \beta \theta_{t}^{\br{k+1}}$
			\ENDFOR 
			\STATE $\theta_{t+1} \gets \br{1 - \beta} \theta_t + \beta \bar \theta_t$
			\STATE $\omega_{t+1} \gets \br{1 - \beta} \omega_t + \beta \bar \omega_t$
			\vspace{4mm}
			\STATE \textbf{GAD (Gradient Ascent Descent):} \begin{align*}
			\theta_{t+1} ~\gets~& \Pi_{\Theta} \br{\theta_t + \eta_t \nabla_\theta f(\theta_t,\omega_t)}\\
\omega_{t+1} ~\gets~& \Pi_{\Omega} \br{\omega_t - \eta_t \nabla_\omega f(\theta_{t+1},\omega_t)} 
\end{align*}
		        \STATE \textbf{EG (Extra-Gradient):} \begin{align*}
\theta_{t+\frac{1}{2}} ~\gets~& \Pi_{\Theta} \br{\theta_t + \eta_t \nabla_\theta f(\theta_t,\omega_t))} \\
\omega_{t+\frac{1}{2}} ~\gets~& \Pi_{\Omega} \br{\omega_t - \eta_t \nabla_\omega f(\theta_t,\omega_t)} \\
\theta_{t+1} ~\gets~& \Pi_{\Theta} \br{\theta_t + \eta_t \nabla_\theta f(\theta_{t+\frac{1}{2}},\omega_{t+\frac{1}{2}})} \\
\omega_{t+1} ~\gets~& \Pi_{\Omega} \br{\omega_t - \eta_t \nabla_\theta f(\theta_{t+\frac{1}{2}},\omega_{t+\frac{1}{2}})} 
\end{align*}
		\ENDFOR
		\STATE \textbf{Output:} $\omega_T$, $\theta_T$.
	\end{algorithmic}
\end{algorithm}

\def\nn{{\nonumber}}

\subsection{Proof of Theorem~\ref{thm:gad_eg_trapped}}
We will focus on the case $f(\theta,\omega) = \theta^2\omega^2 - \theta\omega$. Without loss of generality, we may also assume that $\omega(0) > \theta(0) >0$; the proof of the other cases follows the same argument.

Let $(\theta(t),\omega(t))$ follow the dynamics \eqref{eq:gad.dynamics} with $\theta(0) \cdot \omega(0) > 0.5$. Assume, for the moment, that both $\theta$ and $\omega$ are without constraint. Then we have
\begin{align}
\frac{1}{2} \frac{\drm}{\drm t} \left(  \theta(t)^2 + \omega(t)^2  \right) &=  \theta \frac{\drm \theta}{\drm t} + \omega \frac{\drm \omega}{\drm t} \nn \\
&= 2\theta^2\omega^2 - \theta \omega +  ( - 2\theta^2 \omega^2 + \theta\omega  ) \nn \\
&= 0 \nn
\end{align}implying that $ \theta^2(t) + \omega^2(t) =  \theta^2(0) + \omega^2(0)$ for all $t$. Therefore $\left(r \cos \left(t + \phi_1 \right), r \sin \left(t + \phi_2 \right)  \right)$, where $(r \cos\phi_1, r \sin \phi_2) = (\theta(0), \omega(0))$, is a solution for dynamics for small enough $t$. 

On the other hand, we have
\begin{align}
\frac{\drm}{\drm t} \left(  \theta(t)  \omega(t)  \right) &= \frac{\drm \theta}{\drm t}(t) \cdot \omega(t) +  \theta(t)\cdot \frac{\drm \omega}{\drm t}(t) \nn \\
&= 2\theta(t) \omega^3(t) - \omega^2(t) + \left( - 2 \theta^3(t) \omega(t) + \theta^2(t)  \right) \nn \\
&=  \left(  \theta^2(t)  -  \omega^2(t) \right) \left(1- 2\theta(t) \omega(t) \right)  \nn \\
&=  \left(  \theta^2(t)  -  \omega^2(t) \right) \left(1- 2r^2 \cos \left(t + \phi_1 \right)\sin \left(t + \phi_2 \right)  \right). \nn
\end{align}When $t=0$, we have $1- 2r^2 \cos \left(t + \phi_1 \right)\sin \left(t + \phi_2 \right) = 1- 2 \theta(0) \omega(0) < 0$.  When $t= \frac{\pi}{t}$, we have
\begin{align*}
2r^2 \cos \left(t + \phi_1 \right)\sin \left(t + \phi_2 \right) &= 2r^2 \left(  \frac{\sqrt{2}}{2} \cos \phi_1  -  \frac{\sqrt{2}}{2} \sin\phi_1  \right) \left(  \frac{\sqrt{2}}{2} \cos \phi_2  + \frac{\sqrt{2}}{2} \sin\phi_2  \right) \\
&= \left(  \theta(0) - \sqrt{r^2 - \theta(0)^2} \right)  \left(   \sqrt{r^2 - \omega(0)^2}+\omega(0) \right)  \\
&= (\theta^2(0) - \omega^2(0)  ) <0
\end{align*}whence $1- 2r^2 \cos \left(t + \phi_1 \right)\sin \left(t + \phi_2 \right)  > 0$. The intermediate value theorem then implies that there exists a $\tilde{t}$ such that $1- 2\theta(\tilde{t}) \omega( \tilde{t})=0$. But since $\{(\theta, \omega)~\mid~2\theta\omega~=~1\}$ are the stationary points of the dynamics \eqref{eq:gad.dynamics}, we conclude that $\frac{\drm}{\drm t} \left(  \theta(t)  \omega(t)  \right)  = 0$ whenever $t \geq \tilde{t}$; that is, $(\theta(t),\omega(t))$ gets trapped at the stationary point $(\theta(\tilde{t}),\omega(\tilde{t}))$. The concludes the first part the theorem when there is no boundary. 

If the boundary is present, the dynamics \eqref{eq:gad.dynamics} should be modified to the \emph{projected dynamics} \cite{bubeck2018k} and the proof remains the same, except that when $(\theta(t),\omega(t))$ hits the boundary, the curve needs to traverse along the boundary to decrease the norm.

We now turn to the statement for MixedNE-LD. Let $(\theta_1,\omega_1)$ be initialized at any stationary point: $\theta_1\omega_1=0.5$. Consider the two-step evolution of MixedNE-LD:
\begin{align*}
\theta_2 &= \theta_1 + \sqrt{2\eta} \xi, \\
\omega_2 &= \omega_1 + \sqrt{2\eta}\xi',\\
\theta_3 &= \theta_2 + \eta \left(  2\theta_2\omega_2^2 - \omega_2  \right) + \sqrt{2\eta} \xi'', \\
\omega_3 &=  \omega_2 - \eta \left( 2\theta_2^2\omega_2 - \theta_2 \right) + \sqrt{2\eta} \xi'''
\end{align*}where $\xi,\xi',\xi'',$ and $\xi'''$ are independent standard Gaussian. Since we initialize at a stationary point $\theta_1 \omega_1 = 0.5$, we have 
\begin{align}
2\theta_2\omega_2 -1 &= 2 \theta_1\omega_1 +  \sqrt{2\eta}\omega_1\xi + \sqrt{2\eta}\theta_1  \xi' + 2\eta\xi\xi' -1 \nn \\
&= \sqrt{2\eta}\omega_1\xi + \sqrt{2\eta}\theta_1  \xi' + 2\eta\xi\xi'. \label{eq:proof_1_hold}
\end{align}

Using the towering property of the expectation, \eqref{eq:proof_1_hold}, and the fact that $\xi,\xi',\xi'',$ and $\xi'''$ are independent standard Gaussian, we compute
\begin{align*}
\EE \theta_3 \omega_3 &= \EE\left[ \EE \left[ \theta_3 \omega_3\ |\ \theta_2, \omega_2 \right] \right] \\
&= \EE\left[ \EE \left[  \left( \theta_2 + \eta \left(  2\theta_2\omega_2^2 - \omega_2  \right) + \sqrt{2\eta} \xi'' \right) \left(    \omega_2 - \eta \left( 2\theta_2^2\omega_2 - \theta_2 \right) + \sqrt{2\eta} \xi'''\right)   \ |\ \theta_2, \omega_2 \right] \right] \\
&=\EE\left[ \EE \left[  \left( \theta_2 + \eta \left(  2\theta_2\omega_2^2 - \omega_2  \right) \right) \left(    \omega_2 - \eta \left( 2\theta_2^2\omega_2 - \theta_2 \right) \right)   \ |\ \theta_2, \omega_2 \right] \right] \\
&= \EE\left[ \left( \theta_2 + \eta \omega_2 \left(  2\theta_2\omega_2 - 1  \right)  \right) \left(    \omega_2 - \eta \theta_2\left( 2\theta_2\omega_2 - 1 \right) \right)  \right] \\
&= \EE \left[    \theta_2 \omega_2  - \eta \theta_2^2\left( 2\theta_2\omega_2-1 \right) + \eta \omega_2^2 \left( 2\theta_2\omega_2-1 \right) - \eta^2 \theta_2\omega_2\left( 2\theta_2\omega_2-1 \right)^2   \right] \\
&= \EE \Big[     \theta_1\omega_1  - \eta   \left(   \theta_1^2 + 2\eta\xi^2 + 2\sqrt{2\eta}\theta_1\xi -\omega_1^2 - 2\eta\xi'^2 - 2\sqrt{2\eta}\omega_1 \xi'     \right)  \left(    \sqrt{2\eta}\omega_1\xi + \sqrt{2\eta}\theta_1  \xi' + 2\eta\xi\xi' \right) \\
&\hspace{10mm}   -4\eta^2 \left(    \sqrt{2\eta}\omega_1\xi + \sqrt{2\eta}\theta_1  \xi' + 2\eta\xi\xi'  \right) \\
&\hspace{20mm} \left(   2\eta\omega_1^2\xi^2 + 2\eta \theta_1^2\xi'^2 + 4\eta^2\xi^2\xi'^2 + 2\eta\xi\xi' + 4\sqrt{2}\eta^{\frac{3}{2}} \theta_1 \xi\xi'^2 + 4\sqrt{2}\eta^{\frac{3}{2}} \omega_2\xi^2\xi'   \right)      \Big]  \\
&= \theta_1\omega_1  - 0 - 4\eta^2 \left(   \eta\omega_2^2 + \eta \theta_1^2 + 2\eta^2 + 4\eta^2 + 4\eta^2 + 4 \eta^2 \right) \\
&= \theta_1\omega_1 - 4\eta^2 \left(   \eta \left(\theta_1^2+\omega_1^2\right) + 14\eta^2 \right)
\end{align*}which is \eqref{eq:LD_decrease}.

\subsection{Proof of Theorem~\ref{thm:second.order.doesnt.help}}

Spelling out the Newton dynamics \eqref{eq:newton.dynamics}, we get
\begin{align*}
\frac{\drm \theta}{\drm t} (t) &= \frac{1}{2\omega^2(t)} \left(  2\theta(t) \omega^2(t) - \omega(t) \right) \\
&= \theta(t) - \frac{1}{2\omega(t)}
\end{align*}and similarly $\frac{\drm \omega}{\drm t} (t) = -\omega(t) + \frac{1}{2\theta(t)}$. As a result, we have
\begin{align*}
\frac{\drm}{\drm t} \left(  \theta(t)  \omega(t)  \right) &= \frac{\drm \theta}{\drm t}(t) \cdot \omega(t) +  \theta(t)\cdot \frac{\drm \omega}{\drm t}(t) \\
&=  \theta(t)  \omega(t)  -\frac{1}{2} -  \theta(t) \omega(t)  + \frac{1}{2} \\
&=0
\end{align*}which concludes the proof.